\documentclass{article} 
\usepackage{iclr2022_conference,times}

\usepackage{amsmath,amsfonts,bm}

\def\eqref#1{equation~(\ref{#1})}
\def\Eqref#1{Eq.~(\ref{#1})}

\def\1{\bm{1}}
\newcommand{\train}{\mathcal{D}}

\DeclareMathAlphabet{\mathsfit}{\encodingdefault}{\sfdefault}{m}{sl}
\SetMathAlphabet{\mathsfit}{bold}{\encodingdefault}{\sfdefault}{bx}{n}

\usepackage{amssymb}
 
\usepackage{hyperref}
\usepackage{url}
\usepackage{graphicx}

\usepackage{booktabs}
\usepackage[ruled,vlined]{algorithm2e}
\usepackage{multirow, makecell}
\usepackage{caption}
\usepackage{subcaption}
\usepackage{enumitem}

\title{\textsc{Igeood}: An Information Geometry Approach to Out-of-Distribution Detection}

\author{%
  Eduardo D. C. Gomes, Florence Alberge \& Pierre Duhamel    \\
    Laboratoire des signaux et systèmes (L2S) \\
   Université Paris-Saclay CNRS CentraleSupélec\\
   91190, Gif-sur-Yvette, France. \\
   \texttt{\small\{eduardo.dadalto,florence.alberge,pierre.duhamel\}@centralesupelec.fr} \\
   \AND
Pablo Piantanida \\
International Laboratory on Learning Systems (ILLS) \\
McGill  ETS  MILA  CNRS  Université Paris-Saclay  CentraleSupélec \\
H3C 1K3 Quebec, Canada\\
{\tt\small piantani@mila.quebec}
\\
}

\iclrfinalcopy 
\begin{document}

\maketitle

\begin{abstract}
Reliable out-of-distribution (OOD) detection is fundamental to implementing safer modern machine learning (ML)  systems. In this paper, we introduce \textsc{Igeood}, an effective method for detecting OOD samples. \textcolor{black}{\textsc{Igeood} applies to any pre-trained neural network, works under various degrees of access to the ML model, does not require OOD samples or assumptions on the OOD data but can also benefit (if available) from OOD samples}. By building on the geodesic (Fisher-Rao) distance between the underlying data distributions, our discriminator can combine confidence scores from the logits outputs and the learned features of a deep neural network. Empirically, we show that \textsc{Igeood} outperforms competing state-of-the-art methods on a variety of network architectures and datasets. 
\end{abstract}

\section{Introduction}
Deep neural networks (DNNs) reach the state-of-the-art in several classification tasks as they are known to generalize well on data with a  distribution close to the training set. Whereas, in many practical applications, the training set does not reflect well enough the real-life environment~\citep{dataset_shift_book} which is often non-stationary and sometimes with unpredictable events. Therefore, matching the training scenario to reality can be impossible or too complex. The inability of machine learning (ML) models to adapt to non-stationary distributions could limit their adoption in mission-critical systems (e.g., autonomous devices, healthcare applications).

\textcolor{black}{Out-of-Distribution (OOD) or novelty detection is one of the main objectives in conceiving reliable ML systems~\citep{problems-in-ai-safety}. A typical application is monitoring ML-based online services for periodically shifting distributions. However, tracking changes in the underlying data distribution is challenging as they contain unusual (irregular or unexpected) events and have large dimensions. For instance, relying on the intrinsic properties of  ML models and their statistical behavior in the presence of in-distribution data is essential to identify OOD samples. Classic approaches to OOD detection consist of deriving metrics for detecting those abnormalities from the lens of ML models (e.g., softmax output, latent representations across layers), provided that often only a single test example is available. Furthermore, these metrics are subject to potential limitations inherent in practical scenarios depending on the level of access to information in the ML model, e.g.,  having access only to the last layer or to all intermediate layers.}

The baseline approach for OOD detection relies on the predictive uncertainty of DNNs. \citet{baseline} demonstrated that OOD samples, in general, induce DNN classifiers to output less confident softmax scores, while existing state-of-the-art methods on classification problems still output high accuracy even under dataset shift. For instance, \citet{ovadia2019trust} show that as the accuracy of the underlying DNN increases, the supervisors' outlier detection accuracy also improves. Unfortunately, also the variance increases. \citet{perforrmance2021ood} observed that small changes in model parameters that marginally impact the accuracy could have a degrading impact on the performance of the OOD discriminator. This challenge is not exclusive to discriminative models. Deep generative models also fall short in discerning OOD from in-distribution samples. \citet{nalisnick2019deep} raise awareness of the fact that deep generative models also may output a higher likelihood to OOD samples. They show that, even though the samples from the in-distribution CIFAR-10 \citep{cifar10} dataset (e.g., cats, dogs, airplanes, ships) are conceptually and visually different from house numbers from  SVHN \citep{svhn} dataset, DNN-based classifiers may still assign a high likelihood to SVHN samples.

In this paper, we propose \textsc{Igeood}, a new unified and effective method to perform OOD detection by rigorously exploring the information-geometric properties of the feature space on various depths of a DNN. \textsc{Igeood} provides a flexible framework that applies to any pre-trained softmax neural classifier. A key ingredient of \textsc{Igeood} is the Fisher-Rao distance. This distance is used as an effective differential geometry tool for clustering and as a distance in the context of multivariate Gaussian pdfs \citep{fisher_rao_tutorial, fisher_rao_clustering}. In our context, we measure the dissimilarity between probability distributions (in and out), as the length of the shortest path within the manifold induced by the underlying class of distributions (i.e., the softmax probabilities of the neural classifier or the densities modeling the learned representations across the layers). By doing so, we can explore statistical invariances of the geometric properties of the learned features~\citep{bronstein2021geometric}. Our method adapts to the various scenarios depending on the level of information access of the DNN and uses only in-distribution samples but can also benefit (if available) from OOD samples.  


\subsection{Contributions}
Our work investigates the problem of OOD detection and advances state-of-the-art in different ways.

{\bf{i}} To the best of our knowledge, this is the first work studying \textit{information geometry} tools to devise a unified metric for OOD  detection. We derive an explicit characterization of the Fisher-Rao distance based on the information-geometric properties of the softmax probabilities of the neural classifier and the class of multivariate Gaussian pdfs. In general terms, our Fisher-Rao-based metric measures the mismatch--in the geometry space--between the probability density functions of the pre-trained DNN classifier conditioned on test and in-distribution samples. Section~\ref{sec:fisher_rao} details \textsc{Igeood}.

{\bf{ii}} Experiments on \emph{\textsc{Black-Box}} and \emph{\textsc{Grey-Box}} setups using various datasets, architectures, and classification tasks show that \textcolor{black}{\textsc{Igeood} is competitive with state-of-the-art methods}. In the \textit{\textsc{Black-Box}} setup, we assume that only the outputs, i.e., the logits of the DNN, are available. In the \textit{\textsc{Grey-Box}} setup, we allow access to all parameters of the network; however, the detection must be performed using only the output softmax probabilities. The latter permits input pre-processing which introduces a small (additive) noise in the direction of the gradients w.r.t the test sample. This pre-processing allows for further discrimination between in- and out-of-distribution samples. Our benchmark contains two DNN architectures, three in-distribution datasets, and nine OOD datasets.

{\bf{iii}} In a \emph{\textsc{White-Box}} setting, we combine the logits with the low-level features of the DNN to leverage further useful statistical information of the encoded in-distribution data. We model the pre-trained latent representations as a mixture of Gaussian pdfs with a diagonal covariance matrix. Under this assumption, we derive a confidence score based on the Fisher-Rao distance between conditional pdfs corresponding to the test and the closest in-distribution samples. \textcolor{black}{Experiments based on various datasets, architectures, and classification tasks clearly show consistent improvement of \textsc{Igeood}, achieving new state-of-the-art performance on a couple of benchmarks. In particular, we increased the average TNR at 95\% TPR by 11.2\% with tuning on OOD data and by 2.5\% with tuning on adversarial data compared to \citet{mahalanobis}.}


\subsection{Related works}
OOD discriminators consist of a binary classifier to distinguish between in- and out-of-distribution samples. A few works \citep{shalev2019outofdistribution, hendrycks2019OE_ood, bitterwolf2021certifiably, Mohseni_Pitale_Yadawa_Wang_2020, contrastive_training_ood, vyas2018outofdistribution, hein2019relu} propose retraining the base (or an auxiliary) model with synthetic or ground truth OOD samples to serve as a classifier and as an OOD discriminator. Disposing of both OOD and in-distribution samples during training enables the latent representations to learn the decision boundaries to facilitate OOD detection. These methods will not be compared to ours in this work, as they entail retraining or modifying the base neural network by using OOD data to further train parameters. \citet{nagarajan2021understanding} studies failure modes of OOD detection methods to better understand how to improve them, especially how spurious features like the background can vastly degrade detection performance. \citet{lee2021removing} leverage OOD data as a regularization technique to improve the generalization and robustness of current neural networks. References \citep{anoGAN, kirichenko2020normalizing, choi2019waic, ood_via_generation, xiao2020likelihood,ren2019likelihoodratios, zhang2021outofdistribution, mahmood2021multiscale, Zhang2020HybridMF, Zisselman2020DeepRF} study OOD detection in the context of generative models for density estimation. Open set recognition \citep{open_set_deep_2015}, outlier or anomaly detection \citep{anomaly_survey_pimentel}, concept drift detection \citep{dataset_shift_book}, and adversarial attacks detection \citep{goodfellow2015explaining, madry2019deep} are related topics.

\textbf{\textsc{\textsc{Black-Box}} and \textsc{\textsc{Grey-Box}} scenarios.} It is often the case on ML as a service \citep{mlaas} that the model's parameters knowledge and access are not allowed to the end-user, granting access only to the computation of the forward and the logits or softmax outputs. The baseline work \citep{baseline} for \textsc{\textsc{Black-Box}} techniques simply consider  the unscaled maximum value of the softmax (MSP) as OOD score. In some cases, this confidence score is enough to distinguish between in-distribution and out-of-distribution examples, but it also may assign overconfident values to OOD examples \citep{hein2019relu}. ODIN's \citep{odin} method has two variations. The \textsc{\textsc{Black-Box}} variation consists of temperature scaling the softmax outputs. While the \textsc{Grey-Box} variation also uses an input pre-processing technique that calculates the gradient of the model parameters and adds to the input in an adversarial manner for a more effective OOD detection. \citet{Hsu2020GeneralizedOD} proposes a variation of ODIN that does not need access to OOD data for validation. \citet{liu2020energybased} proposes an energy-based OOD score. They substitute the softmax confidence score with the free energy function with a temperature parameter without retraining. They also propose a \textsc{Grey-Box} variation with posterior processing for improved results. Fine-tuning is done differently across the literature and should be considered when comparing methodologies.

\textbf{\textsc{White-Box} scenario.} This class of OOD detectors has access to all intermediate layer outputs. Naturally, discriminators have access to more information than the \textsc{Black-Box} or \textsc{Grey-Box} setups, warranting greater detection capacity. Batch-normalization statistics between layers are used \citep{quintanilha2019detecting} to fit a logistic regression that serves as an OOD detection score. \citet{gram_matrice} proposes high order Gram matrices to perform OOD detection by computing class-conditional pairwise feature correlations between the test sample and the training set across the hidden layers of the network. \citet{mahalanobis} assume that latent features of DNN models trained under the softmax score follow a class-conditional Gaussian mixture distribution with tied covariance matrix and different class-conditional mean vectors. They calculate the Mahalanobis distance between a test sample as a single estimator of the mean of a class-conditional Gaussian distribution with a tied covariance matrix estimated on the training set. The importance of each low-level component and hyperparameters are tuned using validation data. \citet{ren2021simple} modifies this method to improve detection of near-OOD data. They fit the layer-wise background distribution with a Gaussian distribution fit from the training set. They subtract the Mahalanobis distance between the test example and this distribution from the score proposed in \citet{mahalanobis}, reducing the importance of features shared by in- and out-of-distribution data.


\section{Background}\label{sec:problem_statement}
Let $\mathcal{X}\subseteq \mathbb{R}^d$ be the feature space (continuous) and $\mathcal{Y}$ a label space. Moreover, let $p_{{X}Y}$ be the underlying unknown probability density function (pdf) over $\mathcal{X}\times\mathcal{Y}$. We define the \textit{in-distribution training dataset} as $
\mathcal{D}_{N} \triangleq \big\{( \boldsymbol{x}_i, y_i ) \big\}_{i=1}^{N} \sim p_{{X}Y}$, where $\boldsymbol{x}_i \in \mathcal{X}$ is the input feature data, $y_i \in \mathcal{Y}\triangleq\{1,\dots,C\}$ is the output class among $C$ possible classes and $N$ denotes the number of training samples. The training dataset is characterized by the joint pdf $p_{{X}Y}$ with \textit{in-distribution marginals} ${X} \sim p_{{X}}$ and $Y \sim P_{Y}$. The predictor denoted by  ${f}_{\mathcal{D}_{N} }: \mathcal{X}\rightarrow\mathcal{Y}$  is based on the inferred model $P_{\widehat{Y}|{X}}$, i.e., 
$
f_{\mathcal{D}_{N} }(\boldsymbol{x}) \equiv  f_{{n}}(\boldsymbol{x};\mathcal{D}_{N} ) \triangleq \arg\max_{y\in\mathcal{Y}}~P_{\widehat{Y}|{X}}(y|\boldsymbol{x};\mathcal{D}_{N} ).
$
In order to model the underlying problem, we introduce an artificial binary random variable $Z\in \{0,1\}$ indicating with  $z=1$ that the test sample $\boldsymbol{x}$ is OOD and otherwise, it is in-distribution. The open-world data can then be modeled as a \textit{mixture} distribution $p_{X|Z}$ defined by $p_{X \mid Z}(\boldsymbol{x}| z=0) \triangleq p_{X}(\boldsymbol{x}), \text{ and } p_{X \mid Z}(\boldsymbol{x}| z=1) \triangleq q_{X}(\boldsymbol{x})$. The intrinsic difficulty arises from the fact that very little can be assumed about the unknown distributions $p_{X}$ and $q_{X}$, in particular for out-of-distribution.


\section{\textsc{Igeood}: OOD Detection using the Fisher-Rao Distance}\label{sec:fisher_rao}

This section introduces \textsc{Igeood}, a flexible framework for OOD detection. \textsc{Igeood} is implemented in two ways: at the level of the logits using temperature scaling (Section~\ref{sec:confidence_score_logits}), which  mitigates the high-confidence scores assigned to OOD examples, and layer-wise  level (Section~\ref{sec:confidence_score_layers}).  The key ingredient of \textsc{Igeood} is the Fisher-Rao distance that allows for effective differentiation between in-distribution and out-of-distribution samples. This distance measures the dissimilarity between two probability models within a class of probability distributions by calculating the geodesic distance between two points on the learned manifold. This measure connects information geometry and differential geometry through the R. Fisher information matrix~\citep{Fisher22:MathFound}. Closed-form expressions of this distance are known to multivariate normal distributions under certain assumptions, among others distributions~\citep{fisher_rao_tutorial}.

\subsection{Motivation for the use of the Fisher-Rao distance for OOD detection}

We introduce a simple example to demonstrate conceptually how Fisher-Rao distance is instrumental to OOD detection. It should be noted that this example is limited to one dimension. However, we expect similar behavior with more complex data under the Gaussianity assumptions. 

\begin{figure}
     \centering
     \begin{subfigure}[b]{0.32\textwidth}
         \centering
         \includegraphics[width=\textwidth]{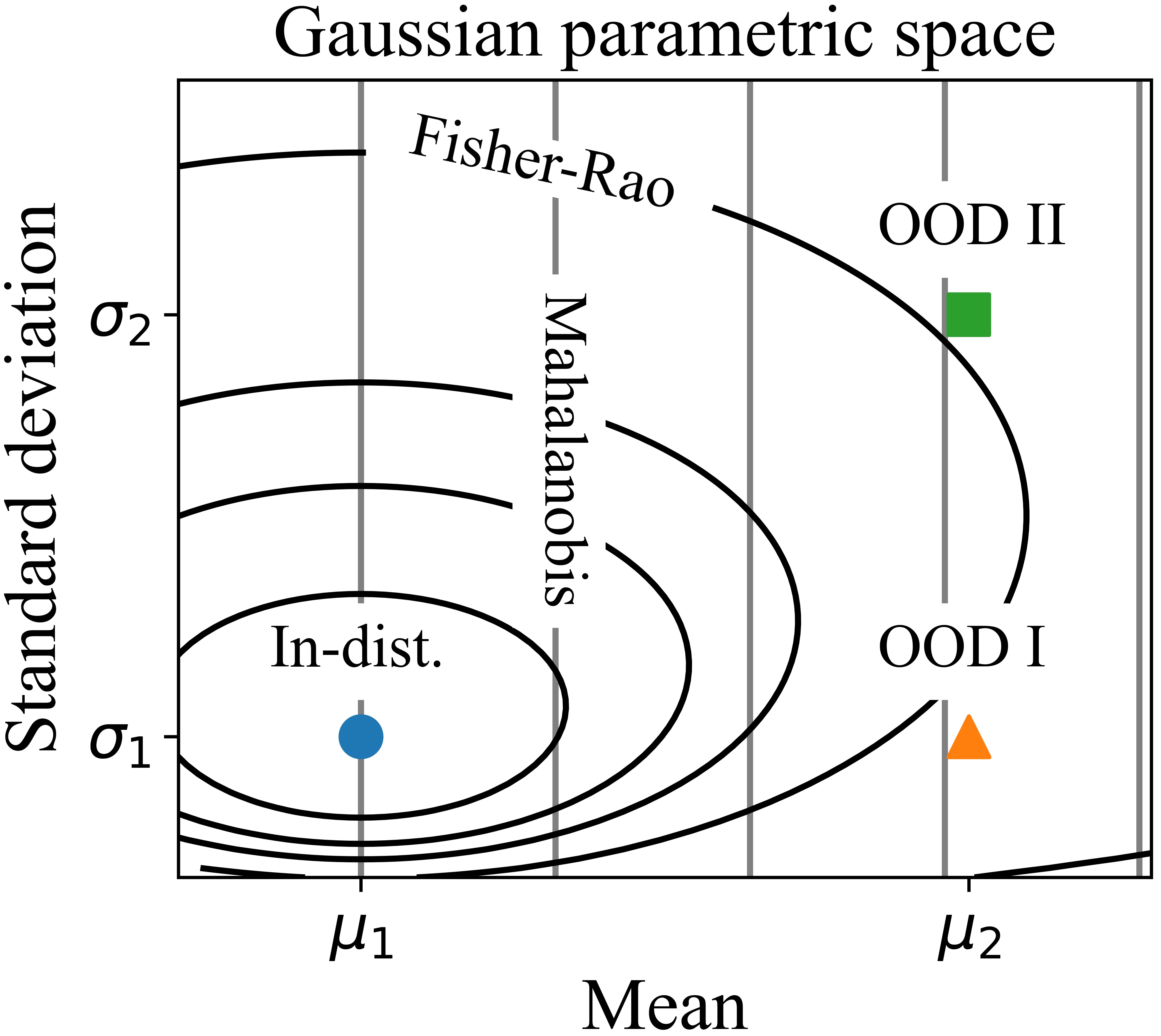}
         \caption{Contour lines.}
         \label{fig:toy_contour}
     \end{subfigure}
     \hfill
     \begin{subfigure}[b]{0.32\textwidth}
         \centering
         \includegraphics[width=\textwidth]{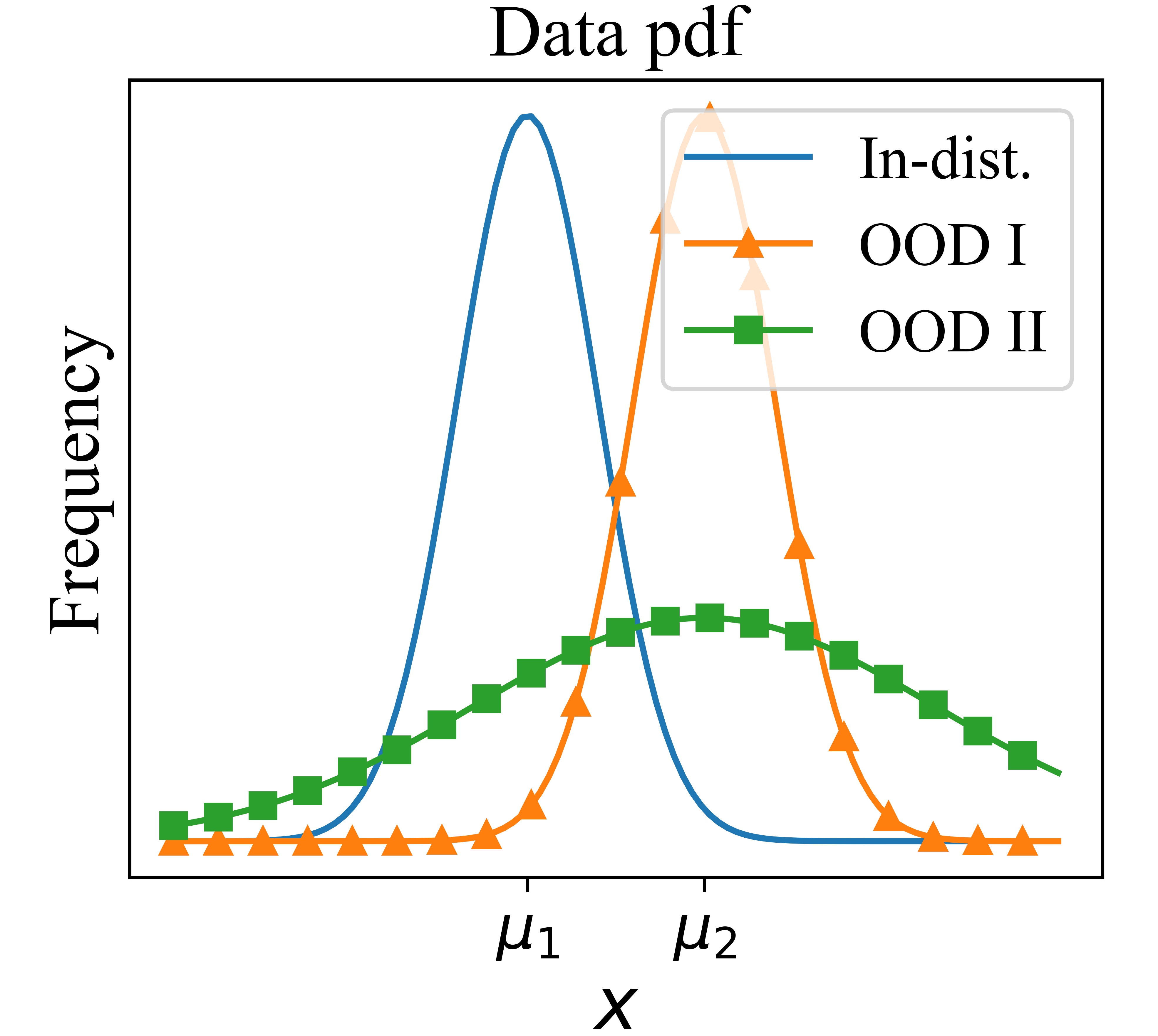}
         \caption{Synthetic data distribution.}
         \label{fig:toy_dist}
     \end{subfigure}
     \hfill
     \begin{subfigure}[b]{0.32\textwidth}
         \centering
         \includegraphics[width=\textwidth]{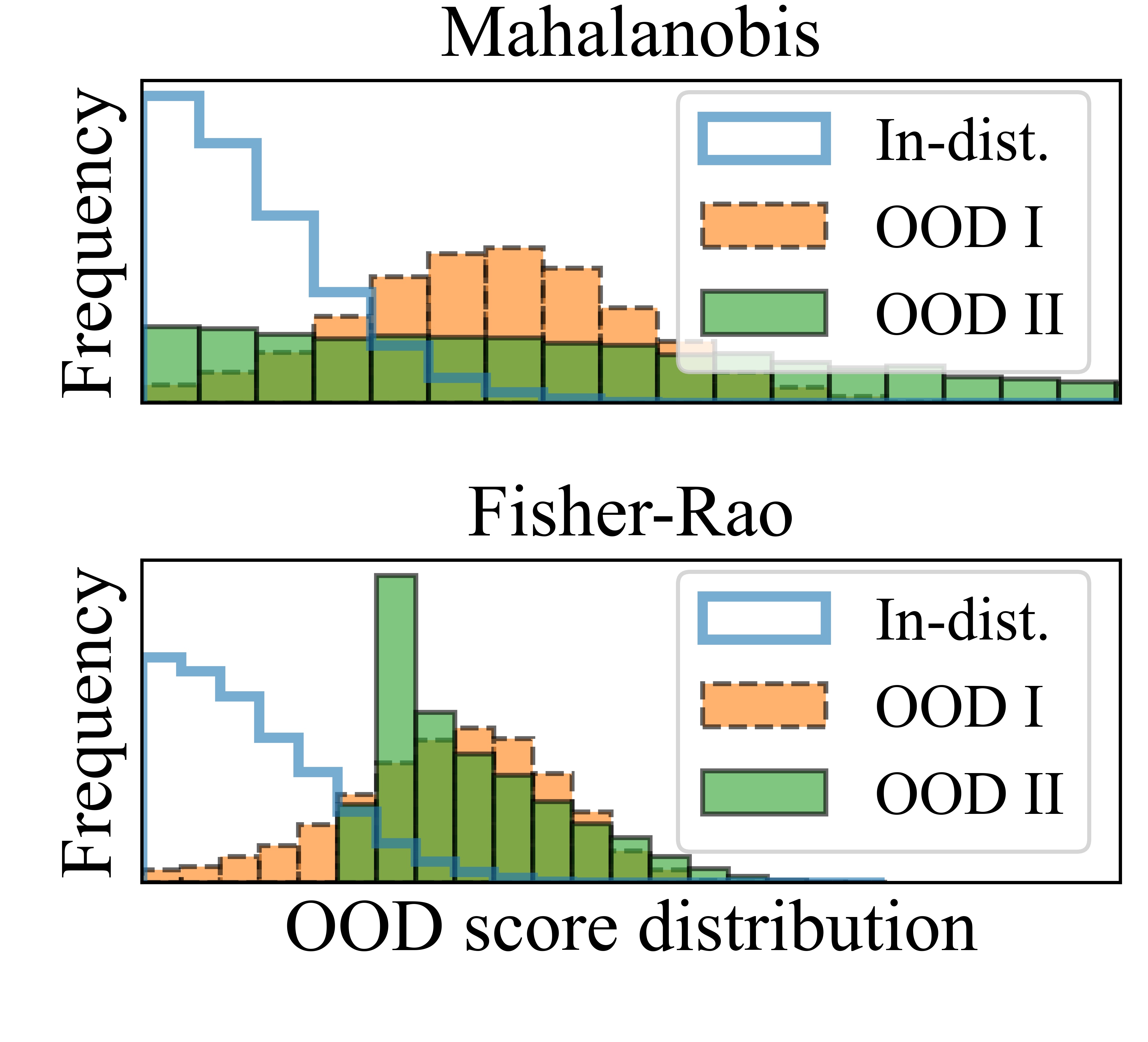}
         \caption{OOD detection score histogram.}
         \label{fig:toy_scores}
     \end{subfigure}
       \caption{Example comparing Fisher-Rao with Mahalanobis distances to  distinguish between 1D Gaussian distributions, showcasing the motivation to use of Fisher-Rao metric for OOD detection.}
    \label{fig:toy_example}
\end{figure}

Consider the case where we try to distinguish between samples from distinct Gaussian distributions on 1D.  Assume that the in-distribution data follows a Gaussian $\mathcal{N}(\mu_1, \sigma_1)$ while OOD data is drawn according to either  $\mathcal{N}(\mu_2, \sigma_1)$ or $\mathcal{N}(\mu_2, \sigma_2)$. These distributions are illustrated in Figures \ref{fig:toy_contour} and \ref{fig:toy_dist}. In this setup, distance-based approaches which are invariant to the variance of the distributions would have the performance limited to the information given by the difference between the means of the underlying distributions. For instance, in the case of the Mahalanobis distance, we would rely our discrimination on the difference between the sample and the in-distribution mean, rescaled by the in-distribution standard deviation only, but nothing further could be obtained. However, if we can estimate OOD standard deviations from actual or pseudo OOD data, we expect the Fisher-Rao distance between Gaussian distributions to be more effective in distinguishing between distributions. Figure \ref{fig:toy_scores} shows that the Fisher-Rao distance distinguishes better between ``In-dist." and ``OOD II" samples, while the other distances fail.

\subsection{\textsc{Igeood} score using the softmax probability}\label{sec:confidence_score_logits}

The Fisher-Rao distance \citep{rao_distance_measure} takes as input two probability distributions. For the classification problem, we can take the temperature $T$ scaled softmax function (\Eqref{eq:softmax}) as an approximation of a class-conditional probability distribution:
\begin{equation}\label{eq:softmax}
q_{\boldsymbol{\theta}}\left(y|f(\boldsymbol{x});T\right) \triangleq \frac{\exp{\left(f_y(\boldsymbol{x})/T\right)}}{\sum_{y^\prime\in\mathcal{Y}} \exp{\left(f_{y^\prime}(\boldsymbol{x})/T\right)}}, 
\end{equation}
where $f:\mathcal{X}\rightarrow \mathbb{R}^C$ is a vectorial function with $f \triangleq \big(f_1, f_2, \dots, f_C\big)$ and $f_y(\cdot)$ denotes the $y$-th logits output value of the DNN classifier. The Fisher-Rao distance $d_{\rm FR-Logits}$ between two  distributions resulting from the softmax probability evaluated at two data points is  (see Appendix~\ref{ap:appendix_a}): 
\begin{equation}\label{eq:fr_dist_softmax}
     d_{\rm FR-Logits}\big(  q_{\boldsymbol{\theta}}(\cdot |f(\boldsymbol{x})), q_{\boldsymbol{\theta}}(\cdot |f(\boldsymbol{x}^\prime) ) \big) \triangleq 2\text{ } \text{arccos}\left(\sum_{y\in\mathcal{Y}}\sqrt{q_{\boldsymbol{\theta}}\big(y|f(\boldsymbol{x})\big) q_{\boldsymbol{\theta}}\big(y|f(\boldsymbol{x}^\prime)\big)}\right).
\end{equation}
\textbf{Class conditional centroid estimation.} We model the training dataset class-conditional posterior distribution by calculating the centroid of the logits representations of this set. Precisely, we compute the \emph{empirical centroid} for the logits of each class $y \in \mathcal{Y}=\{1,\dots, C\}$ of the in-distribution training dataset $\mathcal{D}_{N}$ corresponding to the Fisher-Rao distance, i.e.,  
\begin{equation}\label{eq:centroid}
    \boldsymbol{\mu}_y \triangleq \min_{\boldsymbol{\mu} \in \mathbb{R}^C }\frac{1}{N_y} \sum_{\forall\,i\,:\,y_i=y} d_{\rm FR-Logits}\big(q_{\boldsymbol{\theta}}(\cdot |f(\boldsymbol{x}_i)), q_{\boldsymbol{\theta}}(\cdot |\boldsymbol{\mu})\big),
\end{equation}
where $N_y$ is the amount of training examples with label $y$. We optimize this expression offline using SGD algorithm, where the parameter to be tuned  is $\boldsymbol{\mu}$ in the logits space. This is equivalent to finding the centroid of a cluster using the Fisher-Rao distance, after each example has been assigned to a cluster. Please refer to the appendix (see Section \ref{sec:sup_algorithms}) for further details on this optimization.

\textbf{OOD and confidence score.}
Using the softmax probability, we can define a confidence score to be the minimum of the Fisher-Rao distance between  $f(\boldsymbol{x})$ and the class-conditional centroids. \textcolor{black}{As a sanity check, we show empirically in the appendix (see Section~\ref{sec:sup_experimental_setup}) that this confidence score does not degrade the in-distribution test classification accuracy.} Thus, the estimated class $\widehat{y}_{\rm FR}$ follows as: 
\begin{equation}\label{eq:confidence_score}
    \widehat{y}_{\rm FR}(\boldsymbol{x})\triangleq \arg \min_{y\in\mathcal{Y}} d_{\rm FR-Logits}\big(q_{\boldsymbol{\theta}}(\cdot |f(\boldsymbol{x})), q_{\boldsymbol{\theta}}(\cdot |\boldsymbol{\mu}_y)\big).
\end{equation}
\textcolor{black}{However, we obtained slightly better OOD detection performance by using \Eqref{eq:fr_score_logits} instead of the minimal value. A likely explanation would be that this metric uses extra information from the other logits dimensions. We provide an empirical study comparing both methods in the appendix (see Section~\ref{sec:fr_vs_kl}).} 
Thus, we propose the Fisher-Rao distance-based OOD detection score $\textrm{FR}_0(\boldsymbol{x})$ for the logits to be the sum of the distances between $f(\boldsymbol{x})$ and each individual class conditional centroid $\boldsymbol{\mu}_y$ given by \Eqref{eq:centroid}. By taking the sum instead of the minimal distance, we leverage useful information related to the example's confidence score for each class $y$. We denote it by  
\begin{equation}\label{eq:fr_score_logits}
    \textrm{FR}_0(\boldsymbol{x}) \triangleq  \sum_{y\in \mathcal{Y}} d_{\rm FR-Logits}\big(q_{\boldsymbol{\theta}}(\cdot |f(\boldsymbol{x})), q_{\boldsymbol{\theta}}(\cdot |\boldsymbol{\mu}_y)\big). 
\end{equation}
\textbf{Input pre-processing.} In consonance with the literature \citep{odin, liu2020energybased, mahalanobis}, we also perform input pre-processing to enhance the detection between in-distribution and OOD samples and potentially improve OOD detection performance for the \textsc{Grey-Box} discriminator. We add small magnitude perturbations $\varepsilon$ in a Fast Gradient-Sign Method-style (FGSM) \citep{goodfellow2015explaining} to each test sample $\boldsymbol{x}$ to increase the proposed metric, that is:  
\begin{equation}\label{eq:input_pre_processing}
\widetilde{\boldsymbol{x}}=\boldsymbol{x}+\varepsilon\odot  \operatorname{sign}\big[\nabla_{\boldsymbol{x}} \textrm{FR}_0(\boldsymbol{x})\big].
\end{equation}

\textbf{The OOD detector.} The detector consists of a threshold-based function for discriminating between in-distribution and OOD data. This threshold $\delta$ and parameters are set so that the true positive rate, i.e., the in-distribution samples correctly classified as in-distribution, becomes 95\%. Mathematically, the \textsc{Black-Box} OOD detector $g_{\rm BB}$ and the \textsc{Grey-Box} OOD detector $g_{\rm GB}$ writes:
\begin{equation}
g_{\rm BB}(\boldsymbol{x} ; \delta, T)=\left\{\begin{array}{ll}
1 & \text {if } \textrm{FR}_0\left(\boldsymbol{x}\right) \leq \delta \\
0 & \text {if } \textrm{FR}_0\left(\boldsymbol{x}\right) >\delta
\end{array}\right. \ \text{and } \ \
g_{\rm GB}(\widetilde{\boldsymbol{x}} ; \delta, T, \varepsilon)=\left\{\begin{array}{ll}
1 & \text {if } \textrm{FR}_0\left(\widetilde{\boldsymbol{x}}\right) \leq \delta \\
0 & \text {if } \textrm{FR}_0\left(\widetilde{\boldsymbol{x}}\right) >\delta
\end{array}\right. .
\end{equation}

\subsection{\textsc{Igeood} score leveraging latent features}\label{sec:confidence_score_layers}

For each layer, we define a set of class-conditional Gaussian distributions with diagonal standard deviation matrix $\boldsymbol{\sigma}^{(\ell)}$ and class-conditional mean $\boldsymbol{\mu}_{y}^{(\ell)}$, where $y\in\{1,\dots,C\}$ and $\ell$ is the index of the latent feature. We compute the empirical estimates of these parameters according to 
\begin{equation}\label{eq:mean_cov}
\boldsymbol{\mu}_{y}^{(\ell)}=\frac{1}{N_{y}} \sum_{\forall i\,: \,y_{i}=y} f^{(\ell)}\left(\boldsymbol{x}_{i}\right),\ \ \text{ and } \ \  \boldsymbol{\sigma}^{(\ell)}=\text{diag}\left(\sqrt{\frac{1}{N} \sum_{y\in\mathcal{Y}} \sum_{\forall i\,:\, y_{i}=y}\left(f^{(\ell)}_j\left(\boldsymbol{x}_{i}\right)-\mu_{y, j}^{(\ell)}\right)^2}\right),
\end{equation}
where $j \in \{1,\dots,k\}$, $k$ is the size of feature $\ell$, and $f^{(\ell)}(\cdot)$ is the output of the network for feature $\ell$. The Fisher-Rao distance $\rho_{\rm FR}$ between two arbitrary \textit{univariate} Gaussian pdfs $\mathcal{N}(\mu_1, \sigma_1^2)$ and $\mathcal{N}(\mu_2 , \sigma_2^2)$ is given by (See Section \ref{ap:appendix_a})
\begin{equation}\label{eq:fr_univariate}
\rho_{\rm FR}\left(\left(\mu_{1}, \sigma_{1}\right),\left(\mu_{2}, \sigma_{2}\right)\right)=
\sqrt{2} \log \frac{\left|\left(\frac{\mu_{1}}{\sqrt{2}}, \sigma_{1}\right)-\left(\frac{\mu_{2}}{\sqrt{2}},-\sigma_{2}\right)\right|+\left|\left(\frac{\mu_{1}}{\sqrt{2}}, \sigma_{1}\right)-\left(\frac{\mu_{2}}{\sqrt{2}}, \sigma_{2}\right)\right|}{\left|\left(\frac{\mu_{1}}{\sqrt{2}}, \sigma_{1}\right)-\left(\frac{\mu_{2}}{\sqrt{2}},-\sigma_{2}\right)\right|-\left|\left(\frac{\mu_{1}}{\sqrt{2}}, \sigma_{1}\right)-\left(\frac{\mu_{2}}{\sqrt{2}}, \sigma_{2}\right)\right|}.
\end{equation}
Similarly, the Fisher-Rao distance $d_{\rm FR-Gauss}$ between two \textit{multivariate} Gaussian pdfs with diagonal standard deviation matrix is derived from the univariate case and is given by 
\begin{equation}\label{eq:fr_multivariate_gaussian}
d_{\rm FR-Gauss}\big((\boldsymbol{\mu}, \boldsymbol{\sigma}), (\boldsymbol{\mu}^\prime, \boldsymbol{\sigma}^\prime)\big)=\sqrt{\sum_{i=1}^{k} \rho_{\rm FR}\left(\left(\mu_{i}, \sigma_{ i,i}\right),\left(\mu_{i}^\prime, \sigma_{ i,i}^\prime\right)\right)^{2}},
\end{equation}
where $k$ is the cardinality of the distributions $\mathcal{N}(\boldsymbol{\mu}, \boldsymbol{\sigma})$ and $\mathcal{N}(\boldsymbol{\mu}^\prime, \boldsymbol{\sigma}^\prime)$, $\mu_{i}$ is the $i$-th component of the vector $\boldsymbol{\mu}$, and $\sigma_{i,i}$ is the entry with index $(i,i)$ of the standard deviation matrix $\boldsymbol{\sigma}$. 

\textbf{Experimental support for a diagonal Gaussian mixture model.}
It is known that intermediate features of a DNN can be valuable for detecting abnormal samples as demonstrated by \citet{mahalanobis}. Nonetheless, we observed that the latent features covariance matrices are often \textit{ill-conditioned} and are diagonal dominant. In other words, the condition number of the covariance matrix often diverges, and the magnitude of the diagonal entry in a row is greater than or equal to the sum of all the other entries in that row for most rows. Thus, a diagonal covariance matrix will be a favorable compromise for OOD detection. \textcolor{black}{See Appendix, Section \ref{sec:sup_gauss_test} for further details.}

\textbf{Fisher-Rao distance-based feature-wise confidence score.}
We derive a confidence score by applying the Fisher-Rao distance between the test sample $\boldsymbol{x}$ and the closest class-conditional diagonal Gaussian distribution. Contrarily to the logits, taking the sum did not improve results, so we kept the minimal distance. We can consider two scenarios:  \textbf{(i)} We do not have access to any validation OOD data whatsoever. In this case, the natural choice is to model the test samples as Gaussian distribution with the same diagonal standard deviation as the learned representation, i.e.,
 \begin{equation}\label{eq:fr_score_layer}
    \textrm{FR}_{\ell}(\boldsymbol{x})= \min_{y\in\mathcal{Y}} d_{\rm FR-Gauss}\big((\boldsymbol{x}, \boldsymbol{\sigma}^{(\ell)}),(\boldsymbol{\mu}_{y}^{(\ell)}, \boldsymbol{\sigma}^{(\ell)})\big);
\end{equation}
\textcolor{black}{and \textbf{(ii)} we dispose of a validation OOD dataset on which the features' diagonal standard deviation matrices $\boldsymbol{\sigma}^{\prime(\ell)}$ and the means $\boldsymbol{\mu}^{\prime(\ell)}$ can be estimated, as well as the quantity:}
\textcolor{black}{
\begin{equation}\label{eq:fr_score_layer_2}
    \textrm{FR}^\prime_{\ell}(\boldsymbol{x})=\min_{y\in\mathcal{Y}}  d_{\rm FR-Gauss}\big((\boldsymbol{x}, \boldsymbol{\sigma}^{(\ell)}), (\boldsymbol{\mu}^{\prime(\ell)}, \boldsymbol{\sigma}^{\prime(\ell)})\big).
\end{equation}}
This validation dataset could be obtained from a synthetic dataset, a dataset different from the testing one, or even by adversarially creating OOD data by attacking the classifier model on the training dataset. In the  appendix (Section \ref{sec:sup_algorithms}), we include pseudo-codes for calculating the \textsc{Igeood} score for the \textsc{Black-Box}, \textsc{Grey-Box}, and \textsc{White-Box} settings.

\textbf{Feature ensemble.}\label{sec:feature_ensemble}
To further improve performance, we combine the confidence scores of the logits and the ones from the low-level features through a linear combination. Similarly to the strategy in \citet{mahalanobis}, we choose the weights $\alpha_0$, $\alpha_{\ell}$ and $\alpha^\prime_{\ell}\in \mathbb{R}$ by training a logistic regression detector using validation samples. Thus, we ensure that the metric emphasizes features that demonstrate a greater capacity for detecting abnormal samples. \textsc{Igeood} score for the \textsc{White-Box} setting is: 
\begin{equation}\label{eq:wb_igeood_score}
    \textrm{FR}(\boldsymbol{x}) \triangleq  \alpha_0\textrm{FR}_0(\boldsymbol{x}) + \sum_{\ell} \alpha_{\ell} \cdot  \textrm{FR}_{\ell}(\boldsymbol{x}) + \alpha^\prime_{\ell} \cdot  \textrm{FR}_{\ell}^\prime(\boldsymbol{x}),
\end{equation}
where $\textrm{FR}_0$ is given by \eqref{eq:fr_score_logits}, $\textrm{FR}_{\ell}$ is given by \eqref{eq:fr_score_layer} and $\textrm{FR}^\prime$ considers a different validation diagonal covariance matrix for the test samples (\eqref{eq:fr_score_layer_2}). We also apply input pre-processing similarly to the \textsc{Grey-Box} setting (\eqref{eq:input_pre_processing}), obtaining $\textrm{FR}(\widetilde{\boldsymbol{x}})$ as final score.

\begin{figure}[ht]
    \centering
    \includegraphics[width=1.0\textwidth]{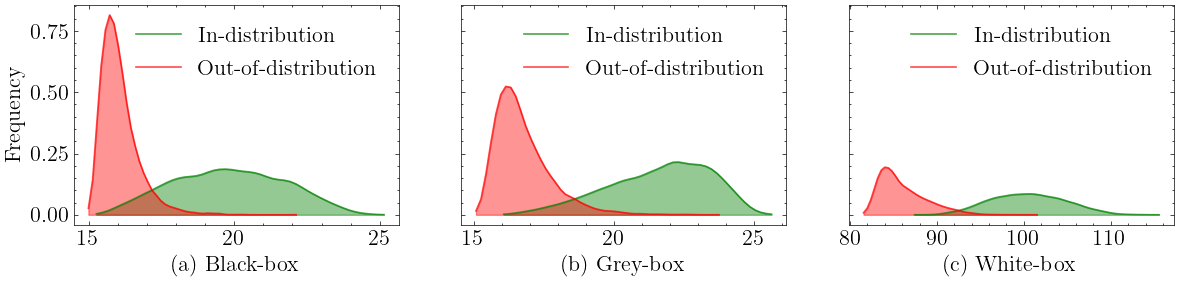}
    \caption{Probability distributions of the \textsc{Igeood} score under three different settings for a pre-trained DenseNet on CIFAR-10 for in-distribution and OOD data (TinyImageNet downsampled).}
    \label{fig:histograms}
\end{figure}

\textbf{Unified metric.} For the three settings, the metric is the same but has different formulations given the family of the distributions. For the DNN outputs, we use the softmax posterior probability distribution formulation. For the intermediate layers, it is under the model of diagonal Gaussian pdfs.  \emph{Therefore, we have derived a unified OOD detection framework that combines a single distance for both the softmax outputs and the latent features of a neural network.} Figure \ref{fig:histograms} illustrates how each of the presented techniques contributes towards separating in-distribution and OOD samples. \textcolor{black}{Additional histograms of the detection scores are relegated to the appendix (see Section \ref{sec:histograms}).}

\section{Experimental Results}\label{sec:results}
We show the effectiveness of \textsc{Igeood} comparing to state-of-the-art methods. Details about the experimental setup \footnote{Our code is publicly available at \url{https://github.com/edadaltocg/Igeood}.} and additional results are given in appendices (see~Sections \ref{sec:sup_experimental_setup}, \ref{sec:sup_benchmark_methods}, and \ref{sec:sup_extended_ood_results}).

\subsection{Setup}
The experimental setup follows the setting established by \citet{baseline}, \citet{odin} and \citet{mahalanobis}. We use two \textit{pre-trained} deep neural networks architectures for image classification tasks: a Dense Convolutional Network (DenseNet-BC-100) \citep{densenet} and a Residual Neural Network (ResNet-34) \citep{resnet}. We take as \textit{in-distribution data} images from CIFAR-10 \citep{cifar10}, CIFAR-100 and SVHN \citep{svhn} datasets.

For \textit{out-of-distribution data}, we use natural image examples from the datasets: Tiny-ImageNet \citep{tiny-imagenet}, LSUN \citep{lsun}, Describable Textures Dataset \citep{cimpoi14describing}, Chars74K \citep{deCampos09}, Places365 \citep{zhou2017places}, iSUN \citep{isun} and a synthetic dataset generated from Gaussian noise. For models pre-trained on CIFAR-10, data from CIFAR-100 and SVHN are also considered  OOD; for models pre-trained on CIFAR-100, data from CIFAR-10 and SVHN are considered OOD, and for models pre-trained on SVHN, CIFAR-10 and CIFAR-100 datasets are considered OOD. We resize the images to dimension $32 \times 32$ by downsampling and applying center crop when needed. We only use test data for evaluation. Even though we ran experiments with image data, \textsc{Igeood} could be applied to any neural-based classification task. 

We measure the effectiveness of the OOD detectors with three standard\textit{ evaluation metrics}: (i) The true negative rate at 95\% true positive rate (TNR at TPR-95\%); (ii) the area under the receiving operating curve (AUROC); and (iii) the area under the precision-recall curve (AUPR). We use the scores over the test set of in-distribution and OOD datasets to calculate them. For the \textsc{Black-Box} and \textsc{Grey-Box} experimental settings, we \textit{tune hyperparameters} for all of the OOD detectors only based on the DNN classifier architecture, the in-distribution dataset, and a validation dataset. The iSUN \citep{isun} dataset is chosen as a source of OOD validation data, independently from OOD test data. We choose the parameters that maximize the TNR at TPR-95\% on the validation OOD dataset. For the \textsc{White-Box} framework, we allow both the benchmark and our method to tune either on adversarially generated data from in-distribution training samples or a separate validation dataset containing $1,000$ images from the OOD test dataset with feature ensemble described in Section \ref{sec:feature_ensemble}. In this case, we evaluate performance on the remaining test samples.


\subsection{Results for the \textsc{Black-Box} and the \textsc{Grey-Box} setups}

For comparing \textsc{Igeood} under the hypothesis of a {\textsc{Black-Box}} scenario, we consider the Baseline \citep{baseline} method, ODIN \citep{odin} with temperature scaling only, and the free-energy-based metric \citep{liu2020energybased} with temperature scaling only. The results for the \textsc{Black-Box} setting are available in Table \ref{tab:black_box_avg}, where we show the average and one standard deviation OOD detection performance for each of the eight OOD detection method in six different image classification contexts (couple DNN model and in-distribution dataset). The extended results for each OOD dataset can be found in Table \ref{tab:black_box_extended}. For comparison under the {\textsc{Grey-Box}} assumption, we consider ODIN and the free-energy-based methods, both with input pre-processing. The results for the {\textsc{Grey-Box}} setup are provided in the appendix (see Section \ref{sec:sup_extended_ood_results} and Table~\ref{tab:grey_box_avg}). \textcolor{black}{For the \textsc{Black-Box} setting, \textsc{Igeood} slight improves the benchmark by less than 1\% in TNR at TPR-95\%. While for the \textsc{Grey-Box} setting, results show \textsc{Igeood} is outperformed by $<$1\% in a few benchmarks by ODIN, which is greatly improved by input pre-processing techniques.}

\begin{table}[ht]
    \begin{center}
      \caption{\textcolor{black}{Average and standard deviation} OOD detection performance across eight OOD datasets for each model and in-distribution dataset in a \textsc{Black-Box} setting. \textsc{Igeood} is compared to Baseline \citep{baseline}, ODIN \citep{odin}, and Energy \citep{liu2020energybased} methods. The extended results can be found in Table \ref{tab:black_box_extended} in the appendix.}
      \label{tab:black_box_avg}
  \begin{tabular}{>{\color{black}}c>{\color{black}}c>{\color{black}}c>{\color{black}}c}
  \toprule
    & & TNR at TPR-95\% & AUROC \\
    Model & In-dist. & \multicolumn{2}{>{\color{black}}c}{Baseline / ODIN  / Energy / \textsc{Igeood} (ours)} \\
      \hline
  \multirowcell{3}{DenseNet} 
& C-10  &  
52.5\textcolor{black}{{\scriptsize$\pm$16}}/\textbf{66.8}\textcolor{black}{{\scriptsize$\pm$20}}/65.3\textcolor{black}{{\scriptsize$\pm$23}}/65.6\textcolor{black}{{\scriptsize$\pm$23}}   &   91.8\textcolor{black}{{\scriptsize$\pm$3.2}}/\textbf{92.8}\textcolor{black}{{\scriptsize$\pm$4.6}}/92.1\textcolor{black}{{\scriptsize$\pm$5.3}}/92.3\textcolor{black}{{\scriptsize$\pm$5.1}}\\
& C-100 & 
15.9\textcolor{black}{{\scriptsize$\pm$6.8}}/20.5\textcolor{black}{{\scriptsize$\pm$9.5}}/20.3\textcolor{black}{{\scriptsize$\pm$9.6}}/\textbf{20.7}\textcolor{black}{{\scriptsize$\pm$9.8}}   &   69.1\textcolor{black}{{\scriptsize$\pm$15}}/71.6\textcolor{black}{{\scriptsize$\pm$20}}/71.6\textcolor{black}{{\scriptsize$\pm$20}}/\textbf{73.2}\textcolor{black}{{\scriptsize$\pm$17}}   \\  
& SVHN        & 
68.4\textcolor{black}{{\scriptsize$\pm$14}}/68.8\textcolor{black}{{\scriptsize$\pm$20}}/70.2\textcolor{black}{{\scriptsize$\pm$17}}/\textbf{72.1}\textcolor{black}{{\scriptsize$\pm$15}}   &   \textbf{92.3}\textcolor{black}{{\scriptsize$\pm$4.0}}/87.3\textcolor{black}{{\scriptsize$\pm$14}}/90.1\textcolor{black}{{\scriptsize$\pm$5.9}}/90.9\textcolor{black}{{\scriptsize$\pm$5.3}}  \\
\hline
\multirowcell{3}{ResNet} 
& C-10      & 
41.7\textcolor{black}{{\scriptsize$\pm$16}}/51.9\textcolor{black}{{\scriptsize$\pm$15}}/56.3\textcolor{black}{{\scriptsize$\pm$13}}/\textbf{56.7}\textcolor{black}{{\scriptsize$\pm$13}}   &   89.6\textcolor{black}{{\scriptsize$\pm$3.1}}/90.4\textcolor{black}{{\scriptsize$\pm$3.1}}/90.4\textcolor{black}{{\scriptsize$\pm$3.0}}/\textbf{90.5}\textcolor{black}{{\scriptsize$\pm$3.0}}  \\
& C-100     & 
15.0\textcolor{black}{{\scriptsize$\pm$5.5}}/16.0\textcolor{black}{{\scriptsize$\pm$6.3}}/16.3\textcolor{black}{{\scriptsize$\pm$7.1}}/\textbf{16.4}\textcolor{black}{{\scriptsize$\pm$6.8}}   &   74.0\textcolor{black}{{\scriptsize$\pm$1.9}}/75.2\textcolor{black}{{\scriptsize$\pm$1.7}}/\textbf{75.5}\textcolor{black}{{\scriptsize$\pm$1.9}}/\textbf{75.5}\textcolor{black}{{\scriptsize$\pm$1.7}} \\ 
& SVHN          & 76.2\textcolor{black}{{\scriptsize$\pm$7.8}}/77.7\textcolor{black}{{\scriptsize$\pm$7.9}}/78.0\textcolor{black}{{\scriptsize$\pm$7.9}}/\textbf{78.3}\textcolor{black}{{\scriptsize$\pm$8.0}}   &   \textbf{92.2}\textcolor{black}{{\scriptsize$\pm$2.9}}/91.4\textcolor{black}{{\scriptsize$\pm$3.2}}/91.4\textcolor{black}{{\scriptsize$\pm$3.2}}/91.7\textcolor{black}{{\scriptsize$\pm$3.2}}  \\
\hline
\multicolumn{2}{>{\color{black}}c}{Average and Std.}                    & 44.9\textcolor{black}{{\scriptsize$\pm$24}}/50.3\textcolor{black}{{\scriptsize$\pm$24}}/51.1\textcolor{black}{{\scriptsize$\pm$24}}/\textbf{51.6}\textcolor{black}{{\scriptsize$\pm$24}}   &   84.8\textcolor{black}{{\scriptsize$\pm$9.5}}/84.8\textcolor{black}{{\scriptsize$\pm$8.3}}/85.2\textcolor{black}{{\scriptsize$\pm$8.4}}/\textbf{85.7}\textcolor{black}{{\scriptsize$\pm$8.0}}  \\

  \bottomrule
  \end{tabular}
    \end{center}
\end{table}

\textbf{Temperature scaling and input pre-processing.}
We observed that low values of temperature and moderate noise magnitude yield better detection performance for \textsc{Igeood} on the logits. For most models and datasets, we obtained better results for temperatures between 1 and 6 and noise magnitudes below 0.002. Detailed results and the best hyperparameters found for each configuration, as well as figures of their impact on performance, are delegated to the appendix (see Section~\ref{sec:sup_extended_ood_results}).

\textbf{How the choice of validation dataset impacts performance.}
We include in the appendix (see Section~\ref{sec:sup_extended_ood_results}) the average OOD detection performance for each method when we change the validation set among the nine available ones. We show that the average TNR at TPR-95\% for \textsc{Igeood} ranges between 63\% and 72\% on a \textsc{Black-Box} scenario and between 65\% and 74\% on a \textsc{Grey-Box} scenario. \textcolor{black}{The performances among the compared methods are consistent across validation datasets.}

\subsection{Results for the \textsc{White-Box} setting}
For benchmarking \textsc{Igeood} on the \textsc{White-Box} setting, we compare results to the Mahalanobis \citep{mahalanobis} method with input pre-processing and feature ensemble. For both of them, we extract features from every output of the dense (or residual) block of the DenseNet (or ResNet) model and the first convolutional layer. The size of each feature is reduced by average pooling in the spatial dimensions. Thus, the initial dimension $\mathcal{F}_{\ell}\times\mathcal{W}_{\ell}\times\mathcal{H}_{\ell}$ is reduced to $\mathcal{F}_{\ell}$, where $\mathcal{F}_{\ell}$ is the number of channels in block $\ell$. For DenseNet, this reduction translates to features of sizes $\boldsymbol{\mathcal{F}}_1=\{24, 108, 150, 342\}$; and for ResNet, to features of sizes $\boldsymbol{\mathcal{F}}_2=\{64, 64, 128, 256, 512\}$. 

We consider two scenarios for tuning hyperparameters for both Mahalanobis and \textsc{Igeood}: one with adversarially generated (FGSM) and in-distribution data and another one with 1,000 OOD samples and in-distribution data. We derive two methods: \textsc{Igeood+}, which is given by \eqref{eq:wb_igeood_score} and considers that we can calculate the statistics from OOD data as additional information; and \textsc{Igeood}, which doesn't consider any prior on OOD data, i.e., set $\alpha_\ell^\prime=0$ on \eqref{eq:wb_igeood_score}.

\textbf{\textcolor{black}{Comparison with current literature.}}
For each DNN model and in-distribution dataset pair, we report the average and one standard deviation OOD detection performance for Mahalanobis \citep{mahalanobis}, \textsc{Igeood} and \textsc{Igeood+}. Table~\ref{tab:white_box_avg} validates the contributions of our techniques. We observe substantial performance improvement in all experiments for the left-hand side of the table, where we outperform Mahalanobis on average for all test cases. \textsc{Igeood+} show improvements of at least 2.1\% up to 23\% on TNR at TPR-95\%. Since the results are usually above 90\%, these improvements are significant.
To assess the consistency of \textsc{Igeood} to the choice of validation data, we measured the detection performance when all hyperparameters are tuned only using in-distribution and generated adversarial data, as observed in the right-hand side of Table~\ref{tab:white_box_avg}. \textsc{Igeood} record improvements up to 10.5\%, and improves by 2.5\% the average TNR at TPR-95\% across all datasets and models. \textcolor{black}{We provide an extra benchmark against other \textsc{White-Box} methods \citep{gram_matrice, Hsu2020GeneralizedOD, Zisselman2020DeepRF} (see Table~\ref{tab:other_comparisons} in the appendix).}

\begin{table}[ht]
\begin{center}
    \caption{\textcolor{black}{Average and standard deviation} OOD detection performance for the \textsc{White-Box} settings. The abbreviation TNR-95\%, C-10 and C-100 stands for TNR at TPR-95\%, CIFAR-10 and CIFAR-100, respectively. The extended results can be found in Tables \ref{tab:wb_ood_extended} and \ref{tab:wb_adv_extended} in the appendix.}
    \label{tab:white_box_avg}
\resizebox{\textwidth}{!}{%
\begin{tabular}{>{\color{black}}c>{\color{black}}c>{\color{black}}c>{\color{black}}c|>{\color{black}}c>{\color{black}}c}
\toprule
 & & \multicolumn{2}{>{\color{black}}c|}{Validation on OOD data}  & \multicolumn{2}{>{\color{black}}c}{Validation on adversarial data}\\
& & TNR-95\% & AUROC & TNR-95\% & AUROC \\
Model &In-dist. & \multicolumn{2}{>{\color{black}}c|}{Mahalanobis  / \textsc{Igeood+} (ours)}  & \multicolumn{2}{>{\color{black}}c}{Mahalanobis / \textsc{Igeood} (ours)}\\
    \hline
\multirowcell{3}{DenseNet} 
& C-10    &  76.6\textcolor{black}{\scriptsize{$\pm$31}}/\textbf{92.6}\textcolor{black}{\scriptsize{$\pm$14}}  &  92.1\textcolor{black}{\scriptsize{$\pm$12}}/\textbf{98.4}\textcolor{black}{\scriptsize{$\pm$3.0}} 
&  75.9\textcolor{black}{\scriptsize{$\pm$30}}/\textbf{77.9}\textcolor{black}{\scriptsize{$\pm$29}}  &  91.7\textcolor{black}{\scriptsize{$\pm$12}}/\textbf{94.0}\textcolor{black}{\scriptsize{$\pm$9.0}} \\
& C-100   &  67.2\textcolor{black}{\scriptsize{$\pm$28}}/\textbf{90.2}\textcolor{black}{\scriptsize{$\pm$21}}  &  90.2\textcolor{black}{\scriptsize{$\pm$13}}/\textbf{97.7}\textcolor{black}{\scriptsize{$\pm$5.0}} 
&  60.4\textcolor{black}{\scriptsize{$\pm$34}}/\textbf{70.9}\textcolor{black}{\scriptsize{$\pm$35}}  &  85.3\textcolor{black}{\scriptsize{$\pm$19}}/\textbf{90.8}\textcolor{black}{\scriptsize{$\pm$13}} \\
& SVHN    &  93.3\textcolor{black}{\scriptsize{$\pm$8.0}}/\textbf{98.0}\textcolor{black}{\scriptsize{$\pm$2.0}}  &  98.6\textcolor{black}{\scriptsize{$\pm$1.0}}/\textbf{99.6}\textcolor{black}{\scriptsize{$\pm$0.1}} 
& \textbf{93.7}\textcolor{black}{\scriptsize{$\pm$10}}/92.2\textcolor{black}{\scriptsize{$\pm$9.0}}  &  \textbf{98.6}\textcolor{black}{\scriptsize{$\pm$2.0}}/98.4\textcolor{black}{\scriptsize{$\pm$1.0}} \\
\hline
\multirowcell{3}{ResNet} 
& C-10      &  82.5\textcolor{black}{\scriptsize{$\pm$23}}/\textbf{91.6}\textcolor{black}{\scriptsize{$\pm$16}}  &  96.5\textcolor{black}{\scriptsize{$\pm$4.0}}/\textbf{98.4}\textcolor{black}{\scriptsize{$\pm$3.0}} 
&  \textbf{78.6}\textcolor{black}{\scriptsize{$\pm$24}}/77.3\textcolor{black}{\scriptsize{$\pm$32}}  &  \textbf{95.3}\textcolor{black}{\scriptsize{$\pm$6.0}}/90.0\textcolor{black}{\scriptsize{$\pm$15}} \\
& C-100     &  70.4\textcolor{black}{\scriptsize{$\pm$30}}/\textbf{86.4}\textcolor{black}{\scriptsize{$\pm$23}}  &  91.9\textcolor{black}{\scriptsize{$\pm$10}}/\textbf{97.1}\textcolor{black}{\scriptsize{$\pm$5.0}} 
&  57.4\textcolor{black}{\scriptsize{$\pm$36}}/\textbf{65.1}\textcolor{black}{\scriptsize{$\pm$33}}  &  86.9\textcolor{black}{\scriptsize{$\pm$13}}/\textbf{88.6}\textcolor{black}{\scriptsize{$\pm$15}} \\
& SVHN      &  96.8\textcolor{black}{\scriptsize{$\pm$6.0}}/\textbf{98.9}\textcolor{black}{\scriptsize{$\pm$2.0}}  &  99.2\textcolor{black}{\scriptsize{$\pm$1.0}}/\textbf{99.7}\textcolor{black}{\scriptsize{$\pm$0.1}} 
&  \textbf{96.3}\textcolor{black}{\scriptsize{$\pm$8.0}}/93.6\textcolor{black}{\scriptsize{$\pm$14}}  &  \textbf{99.1}\textcolor{black}{\scriptsize{$\pm$1.0}}/98.4\textcolor{black}{\scriptsize{$\pm$3.0}} \\
\hline
 \multicolumn{2}{>{\color{black}}c}{Average and Std.} & 81.1\textcolor{black}{\scriptsize{$\pm$11}}/\textbf{92.9}\textcolor{black}{\scriptsize{$\pm$4.0}} & 94.8\textcolor{black}{\scriptsize{$\pm$4.0}}/\textbf{98.5}\textcolor{black}{\scriptsize{$\pm$1.0}}  
 & 77.0\textcolor{black}{\scriptsize{$\pm$15}}/\textbf{79.5}\textcolor{black}{\scriptsize{$\pm$10}} & 92.8\textcolor{black}{\scriptsize{$\pm$5.4}}/\textbf{93.4}\textcolor{black}{\scriptsize{$\pm$3.9}}\\
\bottomrule
\end{tabular}
}
\end{center}
\end{table}


\textbf{Ablation study.}
\textsc{Igeood} has three components, $\textrm{FR}_0$, $\textrm{FR}_\ell$, and $\textrm{FR}_\ell^\prime$, that together compose the final metric of \eqref{eq:wb_igeood_score}. The outputs of the network provide limited OOD detection capacity as observed in Table~\ref{tab:black_box_avg}. When available, the intermediate features, i.e., $\textrm{FR}_{\ell}$, are a valuable resource for OOD detection. Moreover, when few reliable OOD data are available, calculating $\textrm{FR}^\prime_{\ell}$ can further improve the detection performance (left-hand side column of Table~\ref{tab:white_box_avg}). Also, data from a source other than in-distribution, e.g., adversarial samples, is enough for tuning hyperparameters and combining features (right-hand side column of Table~\ref{tab:white_box_avg}). \textcolor{black}{The detection capacity of each hidden layer before any tuning is studied  in Appendix~\ref{sec:sup_feature_importance}. Experiments show that the Fisher-Rao metric effectively separates in- and out-of-distribution data for each of the features individually as well.}

\section{Summary and Concluding Remarks}
This paper introduces \textsc{Igeood}, an effective and flexible method for OOD detection that applies to any pre-trained neural network. The main feature of \textsc{Igeood} relies on the geodesic distance of the probabilistic manifold of the learned latent representations that induces an effective measure for OOD detection. First, in a (\textsc{Grey-}) \textsc{Black-Box} setup, we calculate the sum of the Fisher-Rao distance between the softmax output, corresponding to the test (pre-processed) sample, and a reference probability, corresponding to the conditional-class of softmax probabilities.  Similarly, in a \textsc{White-Box} setup, we model the low-level features of a DNN as a diagonal Gaussian mixture. The Fisher-Rao distance between the pdf of the latent feature, corresponding to the test sample, and a reference pdf, corresponding to the conditional-class of pdfs, provides an effective confidence score. We considered diverse testing environments where prior knowledge of OOD data may or may not be available,  reflecting diverse application scenarios. \textcolor{black}{It is observed that \textsc{Igeood} significantly and consistently improves the accuracy of OOD detection on several DNN architectures across various datasets for a \textsc{White-Box} setting.} Some perspectives for future work include studying causal factors, explainable components for OOD detection, and extensions to textual data.  

\subsubsection*{Acknowledgments}
This work has been supported by the project PSPC AIDA: 2019-PSPC-09 funded by BPI-France.

\bibliography{iclr2022_conference}
\bibliographystyle{iclr2022_conference}

\newpage
\appendix

\section{Review of Fisher-Rao Distance (FRD)}\label{ap:appendix_a}

In this section, we review some results from  references \citet{rao_distance_measure, fisher_rao_tutorial}. We intend to clarify some basic concepts surrounding the Fisher-Rao distance while motivating this measure in the context of OOD detection. 

In few words, the Fisher-Rao's distance is given by the geodesic distance, i.e., the shortest path between points in a Riemannian space induced by a parametric family. Consider the family $\mathcal{C}$ of probability distributions over the class of discrete concepts or labels: $\mathcal{Y}=\{1,\dots,C\}$, denoted by \mbox{$\mathcal{C} \triangleq  \big\{ q_{\boldsymbol{\theta}}(\cdot|\boldsymbol{x}) : \boldsymbol{x} \in \mathcal{X} \subseteq \mathbb{R}^C \big\}$}. 

We are interested in measuring the distance between probability distributions  $q_{\boldsymbol{\theta}}(\cdot|\boldsymbol{x})$ with respect to the testing    input $\boldsymbol{x}$ and a population of inputs drawn accordingly to the in-distribution data set. To this end, we first need to characterize the Fisher-Rao distance for two inputs or for two probability distributions  $q_{\boldsymbol{\theta}},q_{\boldsymbol{\theta}}^\prime\in \mathcal{C} $. 

Assume that the following regularity conditions hold \citep{rao_distance_measure}:
\begin{enumerate}[label=(\roman*)]
	\item $\nabla_{\boldsymbol{x}} \, q_{\boldsymbol{\theta}}(y|\boldsymbol{x})$ exists for all $\boldsymbol{x},y$ and $\boldsymbol{\theta}\in\Theta $;
	\item $\sum\limits_{y \in \mathcal{Y}} \nabla_{\boldsymbol{x}} \, q_{\boldsymbol{\theta}}(y|\boldsymbol{x}) = 0$ for all $\boldsymbol{x}$ and $\boldsymbol{\theta}\in\Theta $;
	\item $\boldsymbol{G}(\boldsymbol{x}) = \mathbb{E}_{Y \sim q_{\boldsymbol{\theta}}(\cdot|\boldsymbol{x})} \big[ \nabla_{\boldsymbol{x}} \, \log q_{\boldsymbol{\theta}}(Y|\boldsymbol{x}) \nabla_{\boldsymbol{x}}^\top \, \log q_{\boldsymbol{\theta}}(Y|\boldsymbol{x}) \big]$ is positive definite for any $\boldsymbol{x}$ and $\boldsymbol{\theta}\in\Theta $.
\end{enumerate}
Notice that if (i) holds, (ii) also holds immediately for discrete distributions over finite spaces (assuming that $\sum_{y \in \mathcal{Y}}$ and $\nabla_{\boldsymbol{x}}$ are interchangeable operations) as in our case. When (i)-(iii) are met, the variance of the differential form $\nabla_{\boldsymbol{x}}^\top \log q_{\boldsymbol{\theta}}(Y|\boldsymbol{x}) d\boldsymbol{x}$ can be interpreted as the square of a differential arc length $ds^2$ in the space $\mathcal{C}$, which yields 
\begin{equation} \label{eq:ds2} 
d s^2 = \langle d\boldsymbol{x}, d\boldsymbol{x} \rangle_{\boldsymbol{G}(\boldsymbol{x})} = d\boldsymbol{x}^\top \boldsymbol{G}(\boldsymbol{x}) d\boldsymbol{x}. 
\end{equation}

Thus, $\boldsymbol{G}$, which is the Fisher Information Matrix (FIM), can be adopted as a metric tensor. We now consider a curve $\boldsymbol{\gamma}:[0,1] \rightarrow \mathcal{X}$ connecting a pair of arbitrary points  $\boldsymbol{x}$, $\boldsymbol{x}^\prime$ in the input space $\mathcal{X}$, i.e., $\boldsymbol{\gamma}(0)=\boldsymbol{x}$ and $\boldsymbol{\gamma}(1)=\boldsymbol{x}^\prime$. Notice that any curve $\boldsymbol{\gamma}$ induces a curve $q_{\boldsymbol{\theta}}(\cdot | \boldsymbol{\gamma}(t))$ for $t\in[0,1]$ in the space $\mathcal{C}$. The Fisher-Rao distance between the distributions $q_{\boldsymbol{\theta}}=q_{\boldsymbol{\theta}}(\cdot | \boldsymbol{x})$ and $q_{\boldsymbol{\theta}}^\prime=q_{\boldsymbol{\theta}}(\cdot | \boldsymbol{x}^\prime)$ will be denoted as $d_{R,\mathcal{C}}(q_{\boldsymbol{\theta}},q_{\boldsymbol{\theta}}^\prime)$ and is formally defined by the expression:
\begin{equation} \label{eq:rao_def} 
d_{R,\mathcal{C}}(q_{\boldsymbol{\theta}},q_{\boldsymbol{\theta}}^\prime) \triangleq  \underset{\boldsymbol{\gamma}}{\text{inf}} \int_{0}^{1} \sqrt{\frac{d \boldsymbol{\gamma}^\top(t)}{dt} \boldsymbol{G}(\boldsymbol{\gamma}(t)) \frac{d \boldsymbol{\gamma}(t)}{dt}}, 
\end{equation}
where the infimum is taken over all piecewise smooth curves. This means that the FRD is the length of the \emph{geodesic} between points $\boldsymbol{x}$ and $\boldsymbol{x}^\prime$ using the FIM as the metric tensor. In general, the minimization of the functional in \eqref{eq:rao_def} is a problem that can be solved using the well-known Euler-Lagrange differential equation. 

\subsection{Derivation of Fisher-Rao distance for the class of Softmax probability distributions}

The direct computation of the FIM of the family $\mathcal{C}$ with $q_{\boldsymbol{\theta}}(y|\boldsymbol{x})$ in the form of the softmax probability distribution function given by \eqref{eq:softmax} can be shown to be singular, i.e., $\text{rank}(\boldsymbol{G}(\boldsymbol{x})) \leq C - 1$, where $C-1$ is the number of degrees of freedom of the manifold $\mathcal{C}$. To  overcome this issue, we introduce the probability simplex $\mathcal{P}$ defined by 
\begin{equation}
\mathcal{P}=\left\{q: \mathcal{Y} \rightarrow[0,1]^{C}: \sum_{y \in \mathcal{Y}} q(y)=1\right\}.
\end{equation}
Next, we consider the following parametrization for any distribution $q\in \mathcal{P}$:
\begin{equation}
q(y | \boldsymbol{z})=\frac{z_{y}^{2}}{4}, \quad y \in\{1, \ldots, C\}.
\end{equation}
From this expression, we consider the statistical manifold $
\mathcal{D}~=\left\{q(\cdot | \boldsymbol{z}):\|\boldsymbol{z}\|^{2}=4, z_{y} \geq 0, \forall y \in \mathcal{Y}\right\}$. Note that the parameter vector $\boldsymbol{z}$ belongs to the positive portion of a sphere of radius 2 and centered at the origin in $\mathbb{R}^C$. The computation of the FIM for $\boldsymbol{z}$ on $\mathcal{D}$ yields:
\begin{equation}\label{eq:fim_softmax}
\begin{aligned}
\boldsymbol{G}(\boldsymbol{z}) &=\mathbb{E}_{q(y | \boldsymbol{z})}\left[\nabla_{\boldsymbol{z}} \log q(y | \boldsymbol{z}) \nabla_{\boldsymbol{z}}^{\top} \log q(y | \boldsymbol{z})\right] \\
&=\sum_{y \in \mathcal{Y}} \frac{z_{y}^{2}}{4}\left(\frac{2}{z_{y}} \boldsymbol{e}_{y}\right)\left(\frac{2}{z_{y}} \boldsymbol{e}_{y}^{\boldsymbol{\top}}\right) \\
&=\sum_{y \in \mathcal{Y}} \boldsymbol{e}_{y} \boldsymbol{e}_{y}^{\top} \\
&=\boldsymbol{I},
\end{aligned}
\end{equation}
where $\left\{\boldsymbol{e}_{y}\right\}$ are the canonical basis vectors in $\mathbb{R}^{C}$ and $\boldsymbol{I}$ is the identity matrix. From \eqref{eq:fim_softmax} we can conclude that the Fisher-Rao metric in this parametric space is equal to the Euclidean metric. Also, since the parameter vector lies on a sphere, the FRD between the distributions $q=q(\cdot | \boldsymbol{z})$ and $q^{\prime}=q\left(\cdot | \boldsymbol{z}^{\prime}\right)$ can be written as the radius of the sphere times the angle between the vectors $\boldsymbol{z}$ and $\boldsymbol{z}^{\prime}$. Which leads to expression:
\begin{equation}
d_{R, \mathcal{D}}\left(q, q^{\prime}\right)=2 \arccos \left(\frac{\boldsymbol{z}^{\boldsymbol{\top}} \boldsymbol{z}^{\prime}}{4}\right)=2 \arccos \left(\sum_{y \in \mathcal{Y}} \sqrt{q(y | \boldsymbol{z}) q\left(y | \boldsymbol{z}^{\prime}\right)}\right).
\end{equation}
Finally, we can compute the FRD for softmax distributions in $\mathcal{C}$ as
\begin{equation}
d_{\rm FR-Logits}\left(q_{\boldsymbol{\theta}}, q_{\boldsymbol{\theta}}^{\prime}\right)=2 \arccos \left(\sum_{y \in \mathcal{Y}} \sqrt{q_{\boldsymbol{\theta}}(y | \boldsymbol{x}) q_{\boldsymbol{\theta}}\left(y | \boldsymbol{x}^{\prime}\right)}\right),
\end{equation}
obtaining the same form of \eqref{eq:fr_dist_softmax}. Notice that $0 \leq d_{\rm FR-Logits}\left(q_{\boldsymbol{\theta}}, q_{\boldsymbol{\theta}}^{\prime}\right) \leq \pi$ for all $\boldsymbol{x}, \boldsymbol{x}^{\prime} \in \mathcal{X}\subseteq \mathbb{R}^C$, being zero when $q_{\boldsymbol{\theta}}(\cdot | \boldsymbol{x})=q_{\boldsymbol{\theta}}\left(\cdot | \boldsymbol{x}^{\prime}\right)$ and maximum when the vectors
$\big(q_{\boldsymbol{\theta}}(1 | \boldsymbol{x}), \ldots, q_{\boldsymbol{\theta}}(C | \boldsymbol{x})\big)$ and $\big(q_{\boldsymbol{\theta}}\left(1 | \boldsymbol{x}^{\prime}\right), \ldots, q_{\boldsymbol{\theta}}\left(C | \boldsymbol{x}^{\prime}\right)\big)$ are orthogonal.

\subsection{Derivation of Fisher-Rao distance for multivariate Gaussian distributions}

Consider a broader statistical manifold  $\mathcal{S}\triangleq \{p_{\boldsymbol{\theta}}=p(\boldsymbol{x} ; \boldsymbol{\theta}) : \boldsymbol{\theta}=\left(\theta_{1}, \theta_{2}, \dots, \theta_{m}\right) \in \Theta\}$ of multivariate differential probability density functions. The Fisher information matrix $\boldsymbol{G}(\boldsymbol{\theta})=\left[g_{i j}(\boldsymbol{\theta})\right]$ in this parametric space is provided by:
\begin{equation}\label{eq:fim_gaussian_1}
\begin{aligned}
g_{i j}(\boldsymbol{\theta}) &=\mathbb{E}_{\boldsymbol{\theta}}\left(\frac{\partial}{\partial \theta_{i}} \log p(\boldsymbol{x} ; \boldsymbol{\theta}) \frac{\partial}{\partial \theta_{j}} \log p(\boldsymbol{x} ; \boldsymbol{\theta})\right) \\
&=\int \frac{\partial}{\partial \theta_{i}} \log p(\boldsymbol{x} ; \boldsymbol{\theta}) \frac{\partial}{\partial \theta_{j}} \log p(\boldsymbol{x} ; \boldsymbol{\theta}) p(\boldsymbol{x} ; \boldsymbol{\theta}) dx.
\end{aligned}
\end{equation}
Next, consider a multivariate Gaussian distribution:
\begin{equation}\label{eq:multivariate_gaussian}
p(\boldsymbol{x} ; \boldsymbol{\mu}, \Sigma)=\frac{(2 \pi)^{-\left(\frac{n}{2}\right)}}{\sqrt{\operatorname{Det}(\Sigma)}} \exp \left(-\frac{(\boldsymbol{x}-\boldsymbol{\mu})^\top \Sigma^{-1}(\boldsymbol{x}-\boldsymbol{\mu})}{2}\right),
\end{equation}
where $\boldsymbol{x} \in \mathbb{R}^{k}$ is the variable vector, $\boldsymbol{\mu} \in \mathbb{R}^{k}$ is the mean vector, $\Sigma \in P_k(\mathbb{R})$ is the covariance matrix, and $P_k(\mathbb{R})$ is the space of $k$ positive definite symmetric matrices. We can define the statistical manifold composed by these distributions as $\mathcal{M}=\{p_{\boldsymbol{\theta}} ; \boldsymbol{\theta} = (\boldsymbol{\mu}, \Sigma) \in \mathbb{R}^k \times P_k(\mathbb{R}) \}$.
By substituting \eqref{eq:multivariate_gaussian} in \eqref{eq:fim_gaussian_1}, we can derive the Fisher information matrix for this parametrization, obtaining:
\begin{equation}
g_{i j}(\boldsymbol{\theta})=\frac{\partial \boldsymbol{\mu}^{\top}}{\partial \theta_{i}} \Sigma^{-1} \frac{\partial \boldsymbol{\mu}}{\partial \theta_{j}}+\frac{1}{2} \operatorname{tr}\left(\Sigma^{-1} \frac{\partial \Sigma}{\partial \theta_{i}} \Sigma^{-1} \frac{\partial \Sigma}{\partial \theta_{i}}\right),
\end{equation}
which induces the following square differential arc length in $\mathcal{M}$:
\begin{equation}\label{eq:gaussian_rao_diff}
d s^{2}=d \boldsymbol{\mu}^{\top} \Sigma^{-1} d \boldsymbol{\mu}+\frac{1}{2} \operatorname{tr}\left[\left(\Sigma^{-1} d \Sigma\right)^{2}\right].
\end{equation}
Here, $d\boldsymbol{\mu}=(d\mu_1,\dots,d\mu_n)\in\mathbb{R}^k$ and $d\Sigma=\left[d\sigma_{ij}\right]\in P_k(\mathbb{R})$. We observe that this metric is invariant to affine transformations \citep{fisher_rao_tutorial}, i.e., for any $(\boldsymbol{c}, Q)\in \mathbb{R}^k \times GL_k(\mathbb{R})$, with $GL_k(\mathbb{R})$ the space of non-singular order $k$ matrices, the map $(\boldsymbol{\mu}, \Sigma) \mapsto\left(Q \boldsymbol{\mu}+\boldsymbol{c}, Q \Sigma Q^{\top}\right)$ is an isometry in $\mathcal{M}$. Thus, the Fisher-Rao distance between two multivariate normal distributions with parameters $\boldsymbol{\theta}_1=(\boldsymbol{\mu}_1, \Sigma_1)$ and $\boldsymbol{\theta}_2=(\boldsymbol{\mu}_2, \Sigma_2)$ in $\mathcal{M}$ satisfies:
\begin{equation}\label{eq:fr_normal_isometry}
d_{R,\mathcal{M}}\left(\boldsymbol{\theta}_{1}, \boldsymbol{\theta}_{2}\right)=d_{R,\mathcal{M}}\left(\left(Q \boldsymbol{\mu}_{1}+\boldsymbol{c}, Q \Sigma_{1} Q^{\top}\right),\left(Q \boldsymbol{\mu}_{2}+\boldsymbol{c}, Q \Sigma_{2} Q^{\top}\right)\right).
\end{equation}
Unfortunately, a closed-form solution for the Fisher-Rao distance remains unknown. This is still an open problem for an arbitrary covariance matrix $\Sigma$ and mean vector $\mu$. Fortunately, the FRD is known for the univariate case and hence, for the submanifold where $\Sigma$ is diagonal. Notice that in this case \eqref{eq:gaussian_rao_diff} admits an additive form.

From \citet{fisher_rao_tutorial}, we obtain the analytical expression of the Fisher-Rao in the 2-dimensional submanifold of univariate Gaussian probability distributions $\mathcal{M}_2=\{p_\mathcal{\boldsymbol{\theta}} : \boldsymbol{\theta}=(\mu, \sigma^2)\in \mathbb{R} \times (0,+\infty)\}$:
\begin{equation}
\rho_{\rm FR}\left(\left(\mu_{1}, \sigma_{1}^2\right),\left(\mu_{2}, \sigma_{2}^2\right)\right)=
\sqrt{2} \log \frac{\left|\left(\frac{\mu_{1}}{\sqrt{2}}, \sigma_{1}\right)-\left(\frac{\mu_{2}}{\sqrt{2}},-\sigma_{2}\right)\right|+\left|\left(\frac{\mu_{1}}{\sqrt{2}}, \sigma_{1}\right)-\left(\frac{\mu_{2}}{\sqrt{2}}, \sigma_{2}\right)\right|}{\left|\left(\frac{\mu_{1}}{\sqrt{2}}, \sigma_{1}\right)-\left(\frac{\mu_{2}}{\sqrt{2}},-\sigma_{2}\right)\right|-\left|\left(\frac{\mu_{1}}{\sqrt{2}}, \sigma_{1}\right)-\left(\frac{\mu_{2}}{\sqrt{2}}, \sigma_{2}\right)\right|},
\end{equation}
where $|\cdot|$ is the Euclidian norm in $\mathbb{R}^2$ and $\sigma$ denotes the standard deviation. Consequently, the FRD for Gaussian distributions with diagonal covariance matrix $\Sigma=\operatorname{diag}\left(\sigma_{1}^{2},\sigma_{2}^{2}, \dots, \sigma_{k}^{2}\right)$ in the $2k$-dimensional statistical submanifold
$
\mathcal{M}_{D}=\big\{p_{\boldsymbol{\theta}} : \boldsymbol{\theta} = (\boldsymbol{\mu}, \Sigma), \Sigma=\operatorname{diag}\left(\sigma_{1}^{2}, \sigma_{2}^{2}, \dots, \sigma_{k}^{2}\right), \sigma_{i}>0, i=1, \dots, k\big\}
$
is 
\begin{equation}
d_{\rm FR-Gauss}\left(\boldsymbol{\theta}_{1}, \boldsymbol{\theta}_{2}\right)=\sqrt{\sum_{i=1}^{k} d_{R,\mathcal{M}_2}\Big(\left(\mu_{1 i}, \sigma_{1 i}\right),\left(\mu_{2 i}, \sigma_{2 i}\right)\Big)^{2}}.
\end{equation}

\subsection{Fisher-Rao vs. Mahalanobis distance}

There is an intricate relationship between the FRD for multivariate Gaussian distributions and the Mahalanobis distance. We borrow the result from \citet{fisher_rao_tutorial}, which states that in the $k$-dimensional submanifold $\mathcal{M}_{\Sigma}$ of $\mathcal{M}$ where $\Sigma$ is constant, i.e., $\mathcal{M}_{\Sigma}=\{p_{\boldsymbol{\theta}} :  \boldsymbol{\theta}=(\boldsymbol{\mu}, \Sigma), \Sigma=\Sigma_{0} \in P_{k}(\mathbb{R}) \}$, the Fisher-Rao distance $d_{R,\mathcal{M}_{\Sigma}}$ between two distributions 
is given by the Mahalanobis distance \citep{mahalanobis1936generalized}:
\begin{equation}
d_{R,\mathcal{M}_{\Sigma}}\big(\mathcal{N}(\boldsymbol{\mu}_1, \Sigma),\mathcal{N}(\boldsymbol{\mu}_2,\Sigma)\big) = \sqrt{(\boldsymbol{\mu}_1 - \boldsymbol{\mu}_2)^T \Sigma^{-1}(\boldsymbol{\mu}_1-\boldsymbol{\mu}_2)}.
\end{equation}
The Mahalanobis distance is also used for OOD detection \citep{mahalanobis} and its performance is compared to the FRD through several experiments in Section \ref{sec:results}. Since the covariance matrix for the hidden layers' outputs is often not full rank, the pseudo-inverse is calculated instead of the inverse.

\section{\textsc{Igeood} algorithms and computation details}\label{sec:sup_algorithms}

In this section, we provide pseudo-code for calculating the \textsc{Igeood} score from the logits (Algorithm \ref{alg:pbb_fisher_rao}) and from the latent features (Algorithm \ref{alg:wb_fisher_rao}). The \textsc{Black-Box} \textsc{Igeood} score is obtained with Algorithm \ref{alg:pbb_fisher_rao} by setting $\varepsilon~=0$, while the \textsc{Grey-Box} \textsc{Igeood} score is obtained with $\varepsilon>0$. We calculated the centroid of the logits for the in-distribution training set by optimizing the objective function given by \eqref{eq:centroid} through a gradient descent algorithm for each DNN. We used a constant learning rate of $0.01$ and a batch size of $128$ for $100$ epochs. Finally, the \textsc{White-Box} \textsc{Igeood} score is obtained by combining the outputs of Algorithms \ref{alg:pbb_fisher_rao} and \ref{alg:wb_fisher_rao} through fitting the multiplicative weights $\alpha$ through a logistic function classifier on a labeled mixture dataset composed from in- and out-of-distribution data according to a validation dataset, which leads to expression \eqref{eq:wb_igeood_score}.

\begin{algorithm}[ht]
 \caption{Evaluating \textsc{Igeood} score based on the logits.}
\SetAlgoLined\label{alg:pbb_fisher_rao}
\SetKwInOut{Input}{Input}
\SetKwInOut{Output}{Output}
\Input{Test sample $\boldsymbol{x}$, temperature $T$ and noise magnitude $\varepsilon$ parameters, and training set $\mathcal{D}_N=\left\{\left(\boldsymbol{x}_i, y_i\right)\right\}_{i=1}^{N}$.}
\Output{$\rm FR_0$: \textsc{Igeood} score in the logits level.}
\medskip
    \texttt{// Offline computation} \\
    \vskip1mm
  Calculate the logits centroids from the training data:     $\boldsymbol{\mu}_y \triangleq \min_{\boldsymbol{\mu} \in \mathbb{R}^C }\frac{1}{N_y} \sum_{\forall\,i\,:\,y_i=y}  2\text{ } \text{arccos}\left(\sum_{y^\prime\in\mathcal{Y}}\sqrt{q_{\boldsymbol{\theta}}\big(y^\prime|f(\boldsymbol{x}_i)\big) q_{\boldsymbol{\theta}}\big(y^\prime|\boldsymbol{\mu}\big)}\right)$\\
      \vskip1mm
    \texttt{// Online computation} \\
    \vskip1mm
  Add small perturbation to $\boldsymbol{x}$: $\widetilde{\boldsymbol{x}}\gets\boldsymbol{x} + \varepsilon \odot  \text{sign}\left[  \nabla_{\boldsymbol{x}}\sum_y 2\text{ } \text{arccos}\left(\sum_{y^\prime\in\mathcal{Y}}\sqrt{q_{\boldsymbol{\theta}}(y^\prime|f(\boldsymbol{x})) q_{\boldsymbol{\theta}}(y^\prime|\boldsymbol{\mu}_{y}})\right)\right]$\\
  \textbf{return } $ \textrm{FR}_0(\widetilde{\boldsymbol{x}})\gets\sum_y 2\text{ } \text{arccos}\left(\sum_{y^\prime\in\mathcal{Y}}\sqrt{q_{\boldsymbol{\theta}}(y^\prime|f(\widetilde{\boldsymbol{x}})) q_{\boldsymbol{\theta}}(y^\prime|\boldsymbol{\mu}_{y}})\right)$\\
\end{algorithm}

\begin{algorithm}[ht]
\caption{Evaluating feature-wise \textsc{Igeood} score.}
\SetAlgoLined\label{alg:wb_fisher_rao}
\SetKwInOut{Input}{Input}
\SetKwInOut{Output}{Output}
\Input{ Test sample $\boldsymbol{x}$ and training set $\mathcal{D}_N=\left\{\left(\boldsymbol{x}_i, y_i\right)\right\}_{i=1}^{N}$.}
\Output{ $\rm FR_{\ell}$: feature-wise \textsc{Igeood} scores.}
\medskip
\For{each feature $\ell \in \{1,\dots,L\}$}{
    \vskip1mm
    \texttt{// Offline computation} \\
    \vskip1mm
    Calculate the means: $\boldsymbol{\mu}_{y}^{(\ell)} \gets \frac{1}{N_{y}} \sum_{i: y_{i}=y} f^{(\ell)}\left(\boldsymbol{x}_{i}\right)$\\
    Calculate the diagonal standard deviation matrix: $\sigma_{jj}^{(\ell)} \gets \sqrt{\frac{1}{N} \sum_{y\in\mathcal{Y}} \sum_{\forall i\,:\, y_{i}=y}\left(f^{(\ell)}_j\left(\boldsymbol{x}_{i}\right)-\mu_{y, j}^{(\ell)}\right)^2}$\\
    \vskip1mm
    \texttt{// Online computation} \\
    \vskip1mm
    Compute the OOD score for $\ell$: $\textrm{FR}_{\ell}(\boldsymbol{x}) \gets \min_y \sqrt{\sum_{j=1}^{k} \rho_{\rm FR}\left(\left(\mu_{y,j}^{(\ell)}, \sigma_{jj}^{(\ell)}\right),\left(f^{(\ell)}_j(\boldsymbol{x}), \sigma_{jj}^{(\ell)}\right)\right)^{2}} $\\
}
\textbf{return }$\big(\textrm{FR}_{1}(\boldsymbol{x}), \dots, \textrm{FR}_{L}(\boldsymbol{x})\big)$
\end{algorithm}

\begin{algorithm}[ht]
\caption{Evaluating feature-wise \textsc{Igeood+} score.}
\SetAlgoLined\label{alg:wb_fisher_rao_plus}
\SetKwInOut{Input}{Input}
\SetKwInOut{Output}{Output}
\Input{ Test sample $\boldsymbol{x}$, training set $\mathcal{D}_N=\left\{\left(\boldsymbol{x}_i, y_i\right)\right\}_{i=1}^{N}$ and $M$ OOD samples $\mathcal{O}_M=\left\{\boldsymbol{x}_i^\prime\right\}_{i=1}^{M}$.}
\vskip1mm
\Output{ $\rm FR_{\ell}$ and $\rm FR_{\ell}^\prime$: feature-wise \textsc{Igeood+} scores.}
\medskip
\For{each feature $\ell \in \{1,\dots,L\}$}{
    \vskip1mm
    \texttt{// Offline computation} \\
    
    Calculate class conditional means: $\boldsymbol{\mu}_{y}^{(\ell)} \gets \frac{1}{N_{y}} \sum_{i: y_{i}=y} f^{(\ell)}\left(\boldsymbol{x}_{i}\right)$\\
    \vskip1mm
    Calculate OOD samples mean: $\boldsymbol{\mu}^{(\ell)\prime} \gets \frac{1}{M} \sum_{i=1}^{M} f^{(\ell)}\left(\boldsymbol{x}^\prime_{i}\right)$\\
    \vskip1mm
    Calculate the diagonal standard deviation matrix from training data: \\  
    \vskip1mm
    \ $\sigma_{jj}^{(\ell)} \gets \sqrt{\frac{1}{N} \sum_{y\in\mathcal{Y}} \sum_{\forall i\,:\, y_{i}=y}\left(f^{(\ell)}_j\left(\boldsymbol{x}_{i}\right)-\mu_{y, j}^{(\ell)}\right)^2}$\\
    \vskip1mm
    Calculate the diagonal standard deviation matrix from OOD data: \\ 
    \vskip1mm
    \ $\sigma_{jj}^{(\ell)\prime} \gets \sqrt{\frac{1}{M} \sum_{i=i}^M\left(f^{(\ell)}_j\left(\boldsymbol{x}_{i}^\prime\right)-\mu_{j}^{(\ell)\prime}\right)^2}$\\
    \vskip2mm
    \texttt{// Online computation}\\
    \vskip1mm
    Compute the OOD scores for $\ell$:\\ \ $\textrm{FR}_{\ell}(\boldsymbol{x}) \gets \min_y \sqrt{\sum_{j=1}^{k} \rho_{\rm FR}\left(\left(\mu_{y,j}^{(\ell)}, \sigma_{jj}^{(\ell)}\right),\left(f^{(\ell)}_j(\boldsymbol{x}), \sigma_{jj}^{(\ell)}\right)\right)^{2}} $\\
    \vskip1mm
    \ $\textrm{FR}^\prime_{\ell}(\boldsymbol{x}) \gets \min_y \sqrt{\sum_{j=1}^{k} \rho_{\rm FR}\left(\left(\mu_{j}^{(\ell)\prime}, \sigma_{jj}^{(\ell)\prime}\right),\left(f^{(\ell)}_j(\boldsymbol{x}), \sigma_{jj}^{(\ell)}\right)\right)^{2}} $\\
}
\textbf{return }$\big(\textrm{FR}_{1}(\boldsymbol{x}),\textrm{FR}^\prime_{1}(\boldsymbol{x}) \dots, \textrm{FR}_{L}(\boldsymbol{x}),\textrm{FR}^\prime_{L}(\boldsymbol{x})\big)$
\end{algorithm}

Note that the calculation of the training logits centroids $\boldsymbol{\mu}_y$, as well as the latent representations' mean vectors $\boldsymbol{\mu}_{y}^{(\ell)}$ and standard covariance matrices $\boldsymbol{\sigma}^{(\ell)}$ is performed beforehand, prior to inference. In this way, we retrieve the objects from memory at inference time. Also, we define $k$ as the cardinality of feature $\ell$, or $|f^{(\ell)}|$ and $\rho_{\rm FR}$ as the Fisher-Rao distance between univariate Gaussian distribution given by expression \eqref{eq:fr_univariate}.

\subsection{Logits centroids estimation details}

In order to obtain the logits centroids given the Fisher-Rao distance in the space of softmax probability distributions, we designed a simple optimization problem. This problem aims to minimize the average distance between the class conditional training samples and the centroids as given by \eqref{eq:centroid}. We initialized the $C$ centroids, where $C$ is the number of classes of a given model, with the identity matrix of size $C\times C$. Note that the initial centroid for class $i$ is given by the matrix's line number $i$. We minimized the expression in \eqref{eq:centroid} with a gradient descent optimizer for 100 epochs with a fixed learning rate equal to 0.1 for every DNN model and in-distribution dataset.

\textcolor{black}{The computation of the logits centroid is done offline, and the loss of the centroid estimation converges fast. We show in Table~\ref{tab:time_analysis} the execution time for some operations in the OOD detection pipeline accelerated by one GPU. The left-hand column shows the offline computations needed to run our setup. They are as follow: 
\begin{itemize}
\item Save train set logits: We first do a forward pass through all the training sets and save in memory the resulting logits for a given network, which takes on average 83s for CIFAR-10 and CIFAR-100;
\item Centroid estimation: We load the training logits from memory and run the  Gradient Descent algorithm, which takes on average 1.2s for CIFAR-10 and 11s for CIFAR-100;
\end{itemize}
The right-hand side of Table~\ref{tab:time_analysis} shows the average online computation time for one test sample in a \textsc{Black-Box} setting. 
\begin{itemize}
\item Model inference: The average time needed to complete one forward pass for a DenseNet-BC-100 model is 28 ms and 19 ms for CIFAR-10 and CIFAR-100, respectively.
\item MSP and \textsc{Black-Box} \textsc{Igeood} computations. Computing the  OOD detection scores from the calculated softmax output is roughly 100 to 1000 times faster than the inference time taken by the model. 
\end{itemize}
Hence, computing the Fisher-Rao distance between a test sample and the class-conditional centroids does not account for a considerable overhead in execution time.
}
\begin{table}[ht]
    \begin{center}
    \caption{\textcolor{black}{Execution time analysis for an experimental set accelerated by a single GPU for a DenseNet-BC-100 architecture pre-trained on CIFAR-10 and CIFAR-100. We show the average value for 5 runs.}}
    \label{tab:time_analysis}
    \begin{tabular}{>{\color{black}}l >{\color{black}}c>{\color{black}}c|>{\color{black}}c>{\color{black}}c>{\color{black}}c}
    \toprule
    & \multicolumn{2}{>{\color{black}}c|}{Offline computation} & \multicolumn{3}{>{\color{black}}c}{Online computation}\\
        \cmidrule{2-6}
        \multirowcell{2}{In-dist.\\Dataset} &
        \multirowcell{2}{Save\\train set logits} &
        \multirowcell{2}{Centroid\\estimation} & 
        \multirowcell{2}{Model\\inference} &
        \multirowcell{2}{MSP\\computation} &
        \multirowcell{2}{\textsc{Black-Box}\\\textsc{Igeood} computation}\\
        & & & & &\\
        \midrule
        CIFAR-10 & 83 s & 1.2 s & 28 ms & 63 $\mu$s & 66 $\mu$s \\
        CIFAR-100 & 83 s & 11 s & 19 ms & 34 $\mu$s & 171 $\mu$s  \\
    \bottomrule
    \end{tabular}
    \end{center}
\end{table}

\subsection{Covariance matrix estimation details}

We model the latent output probability distributions as Gaussian distributions with diagonal covariance matrix calculated with \eqref{eq:mean_cov}. We chose this model motivated by a closed form for the FRD and by observing that the standard covariance matrix for the latent features is often ill-conditioned and diagonal dominant. The condition number of a matrix correlates to its numerical stability, i.e., a small rounding error in its estimation may cause a large difference in its values. So, a matrix with a low condition number is said to be well-conditioned, while a matrix with a high condition number is said to be ill-conditioned. We calculate the condition number of the covariance matrices with the formula $\kappa(\Sigma)=\left\|\Sigma^{-1}\right\|_{\infty}\|\Sigma\|_{\infty}$, where $\|\cdot\|_{\infty}$ is the infinity norm. For each of the four dense blocks outputs of a DenseNet trained on CIFAR-10, we obtained the condition numbers $\kappa_{\Sigma}=\{2.8\mathrm{e}10, 3.5\mathrm{e}6, 3.1\mathrm{e}5, 3.5\mathrm{e}21\}$. While for the \textit{diagonal} covariance matrix, we obtained smaller values of condition numbers: $\kappa_{\Sigma_D}=\{1.0\mathrm{e}3, 3.0\mathrm{e}1, 1.4\mathrm{e}1, 7.6\mathrm{e}20\} $. We associate the high value for the last feature mainly because the last feature is high dimensional and coarse, i.e., most of the values in the diagonal are close to zero. 

\subsection{\textcolor{black}{Gaussianity test of the hidden layers' outputs}}\label{sec:sup_gauss_test}
\begin{figure}[ht]
        \centering
     \begin{subfigure}[b]{1.0\textwidth}
         \centering
         \includegraphics[width=\textwidth]{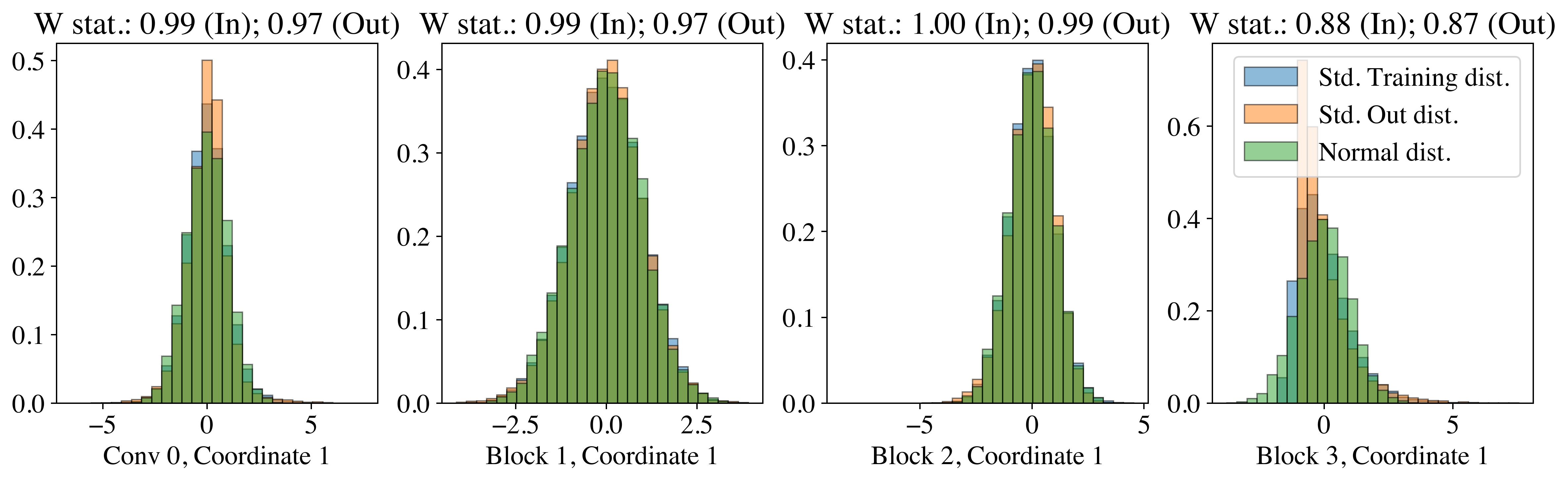}
         \caption{DenseNet pre-trained on CIFAR-10.}
         \label{fig:w_c10}
     \end{subfigure}
     \hfill
     \begin{subfigure}[b]{1.0\textwidth}
         \centering
         \includegraphics[width=\textwidth]{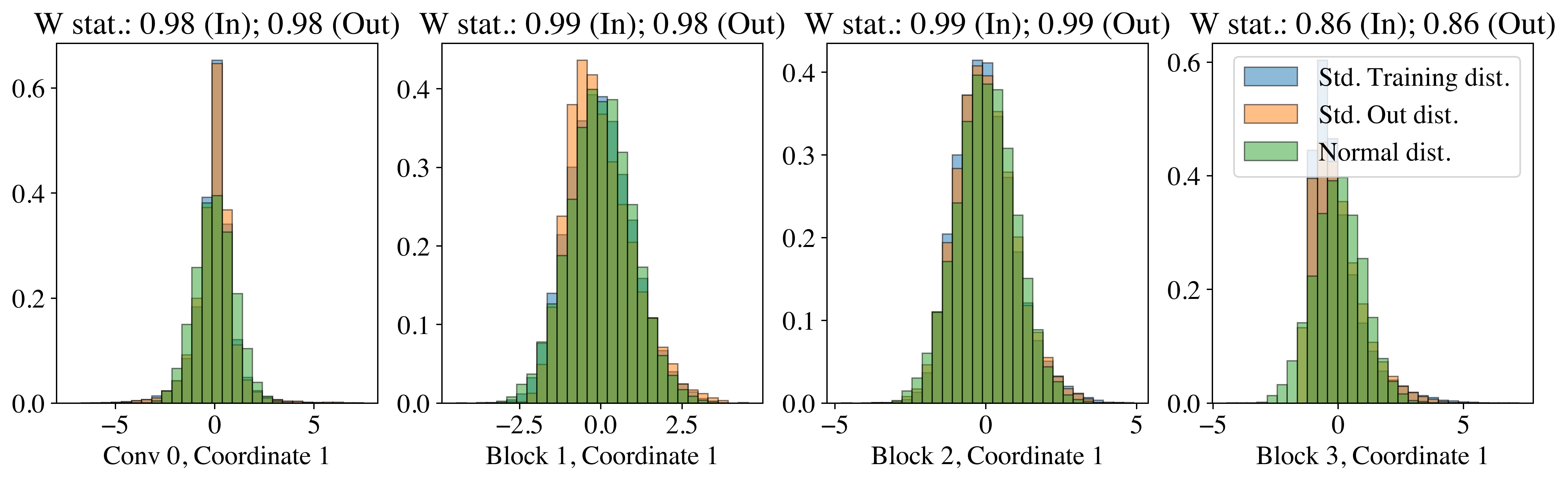}
         \caption{DenseNet pre-trained on CIFAR-100.}
         \label{fig:w_c100}
     \end{subfigure}
     \hfill
     \begin{subfigure}[b]{1.0\textwidth}
         \centering
         \includegraphics[width=\textwidth]{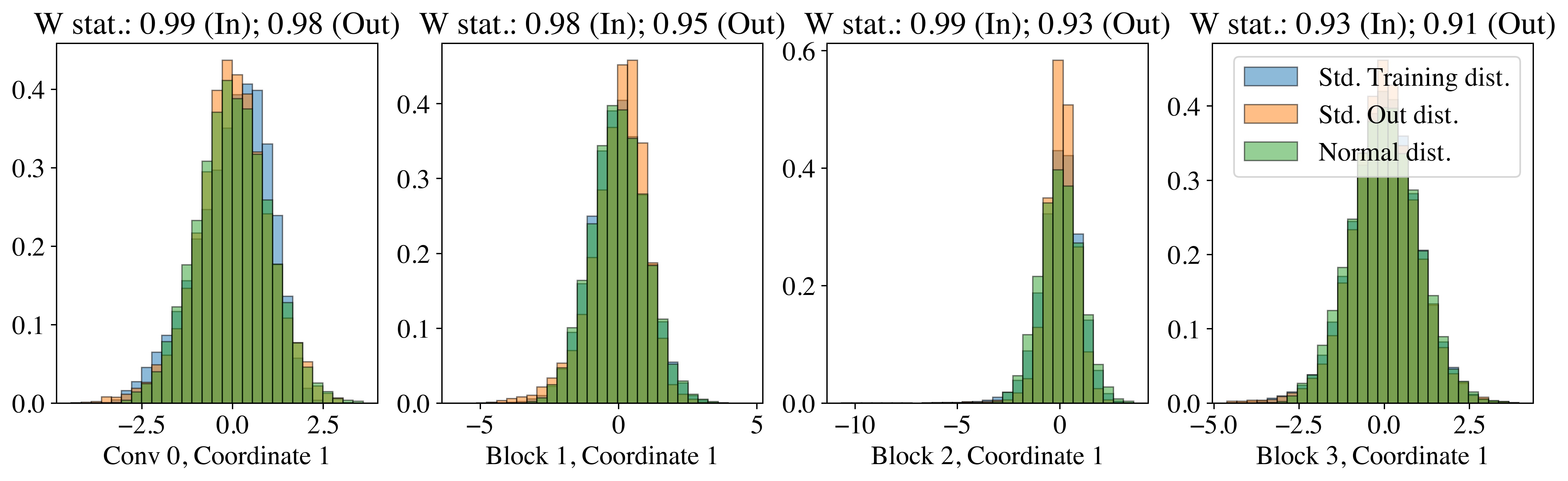}
         \caption{DenseNet pre-trained on SVHN.}
         \label{fig:w_svhn}
     \end{subfigure}
    \caption{\textcolor{black}{Histograms of the standardized first coordinate output of each hidden feature of a DenseNet model for in-distribution and out-of-distribution (TinyImageNet) compared to a 1-D Normal distribution. The Average Shapiro-Wilk test's W statistics is close to one for Conv 0, Block 1 and Block 2, which indicates that the coordinates, and potentially the feature vector, are provably Gaussian. The penultimate layer (outputs of Block 3) has a lower test statistic for the given experiments.}}
    \label{fig:g_test}
\end{figure}

\textcolor{black}{
In order to test if the Gaussian assumption is valid for the outputs of the hidden feature, we conduct a \citet{shapiro_wilk} normality test for each coordinate of the features for the training data of a DenseNet model. We calculated the test's $W$ statistic for each coordinate and class and averaged them. We chose a univariate normality test because they are often powerful and the problem is high dimensional, which would be unfavorable for a multivariate statistic test. Thus, this study should be considered with caution, given the considered hypothesis.  In Figure \ref{fig:g_test}, we also show the standardized histograms for the first coordinate of each layer. 
}
\textcolor{black}{
Note that, apart from the penultimate layer, if we consider the coordinates of the hidden features independently, the Gaussianity assumption holds, as we obtain a $W$ statistics close to 1. However, for the last block, this assumption sometimes does not hold. Hence, modeling the penultimate layer with a more powerful density estimator, and using a metric that considers this more complex distribution, may be favorable for OOD detection. 
}

\subsection{Feature importance regression details}\label{sec:sup_feature_importance}

For both Mahalanobis and \textsc{Igeood} methods, we fitted a logistic regression model with cross-validation using 1,000 OOD and 1,000 in-distribution data samples. Each regression parameter multiplies the layer scores outputs with the objective function of maximizing the TNR at TPR-95\%. We set the maximum number of iterations to 100.

\begin{figure}[ht]
        \centering
     \begin{subfigure}[b]{0.32\textwidth}
         \centering
         \includegraphics[width=\textwidth]{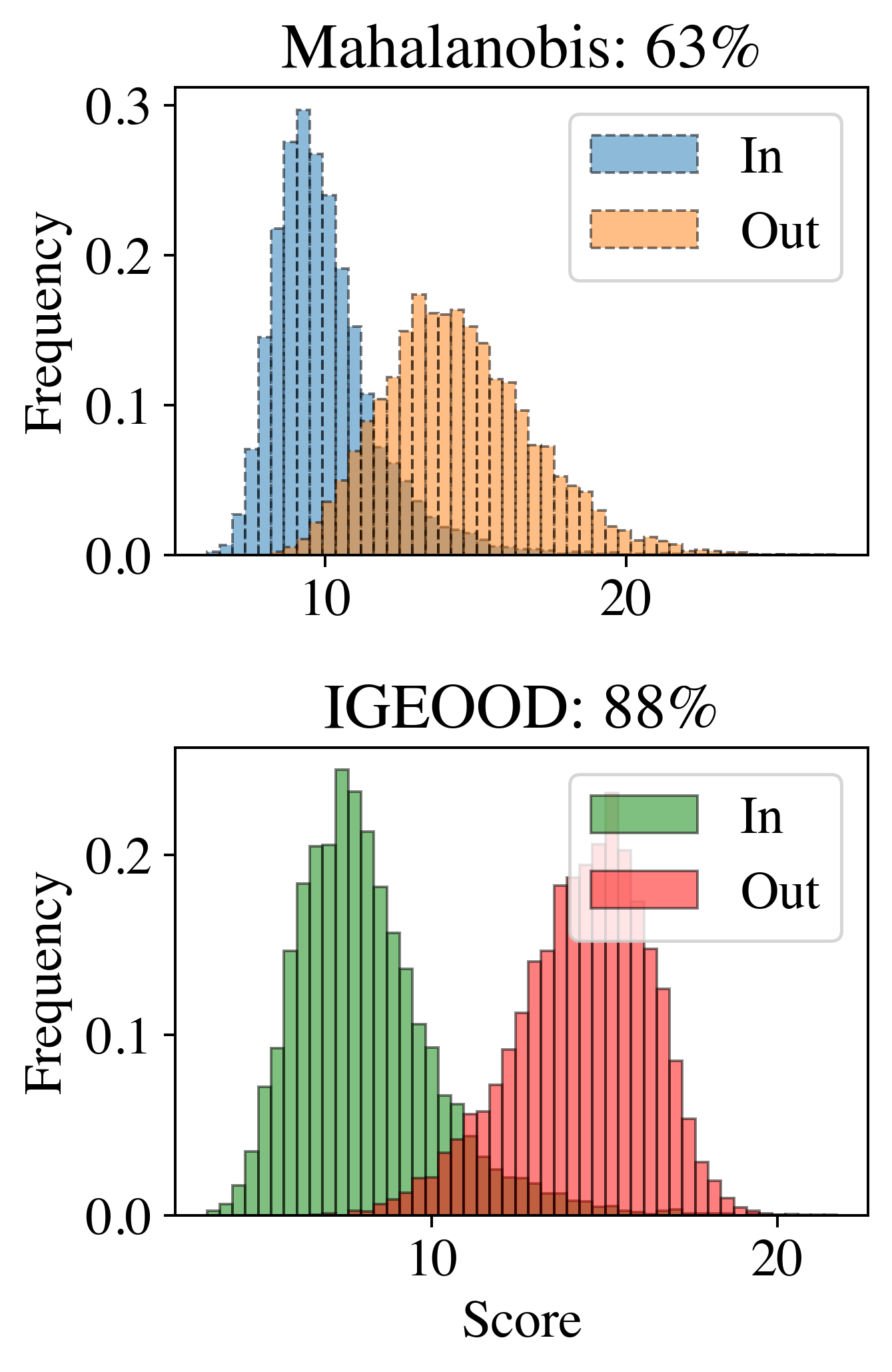}
         \caption{Block 1.}
         \label{fig:hist_block_1}
     \end{subfigure}
     \hfill
     \begin{subfigure}[b]{0.32\textwidth}
         \centering
         \includegraphics[width=\textwidth]{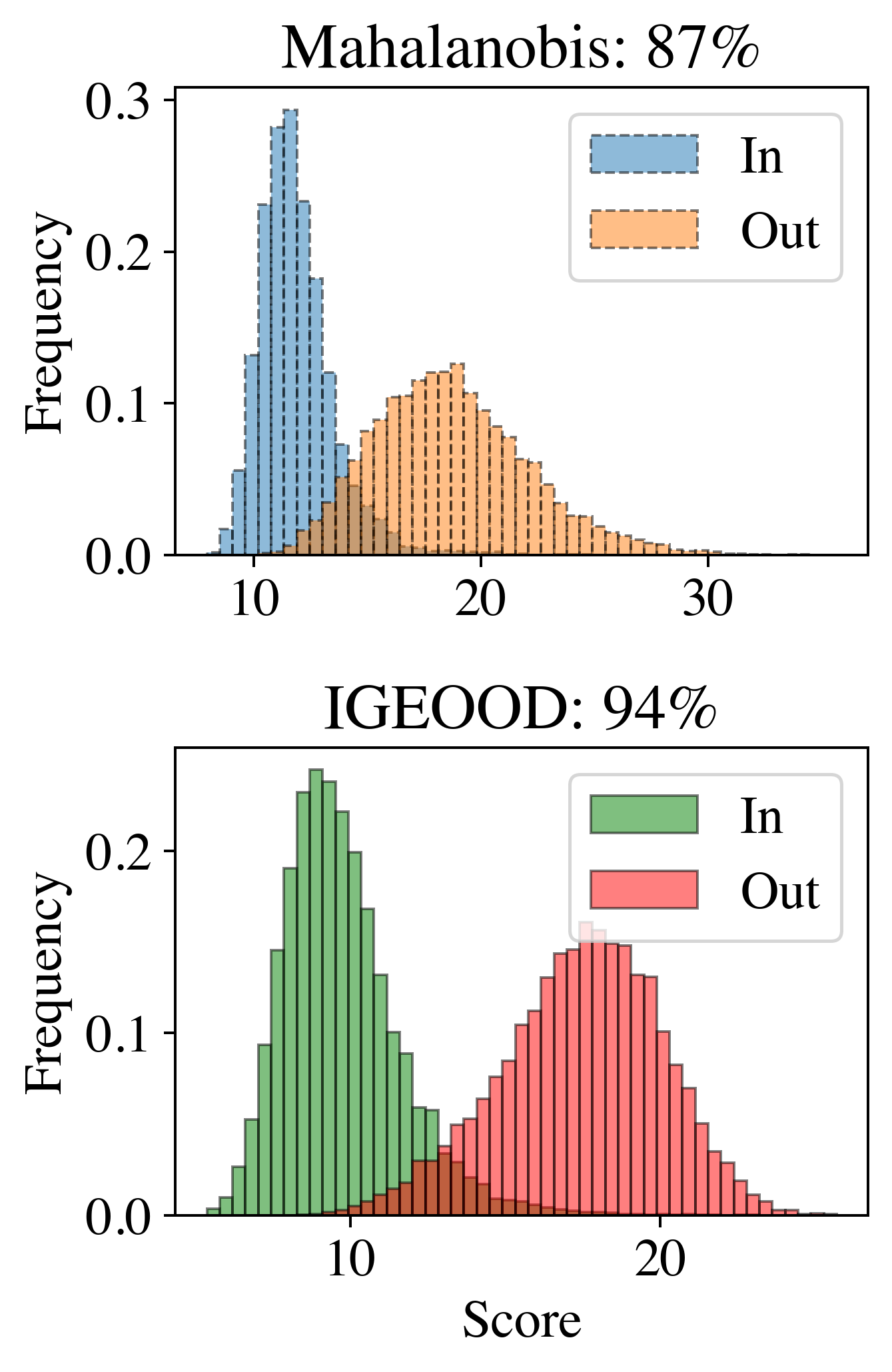}
         \caption{Block 2.}
         \label{fig:hist_block_2}
     \end{subfigure}
     \hfill
     \begin{subfigure}[b]{0.32\textwidth}
         \centering
         \includegraphics[width=\textwidth]{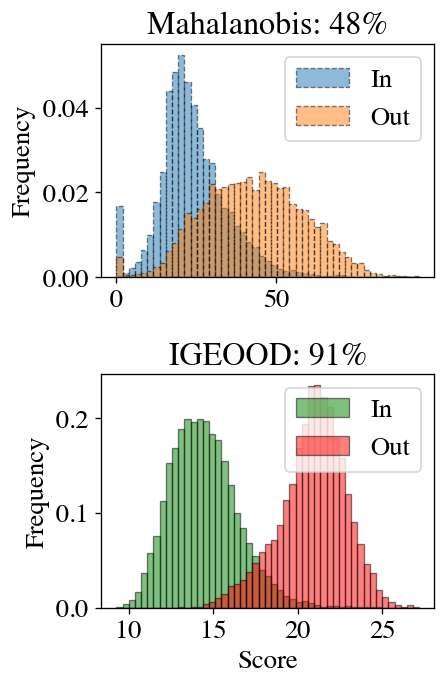}
         \caption{Block 3.}
         \label{fig:hist_block_3}
     \end{subfigure}
     \begin{subfigure}[b]{0.4\textwidth}
         \centering
         \includegraphics[width=\textwidth]{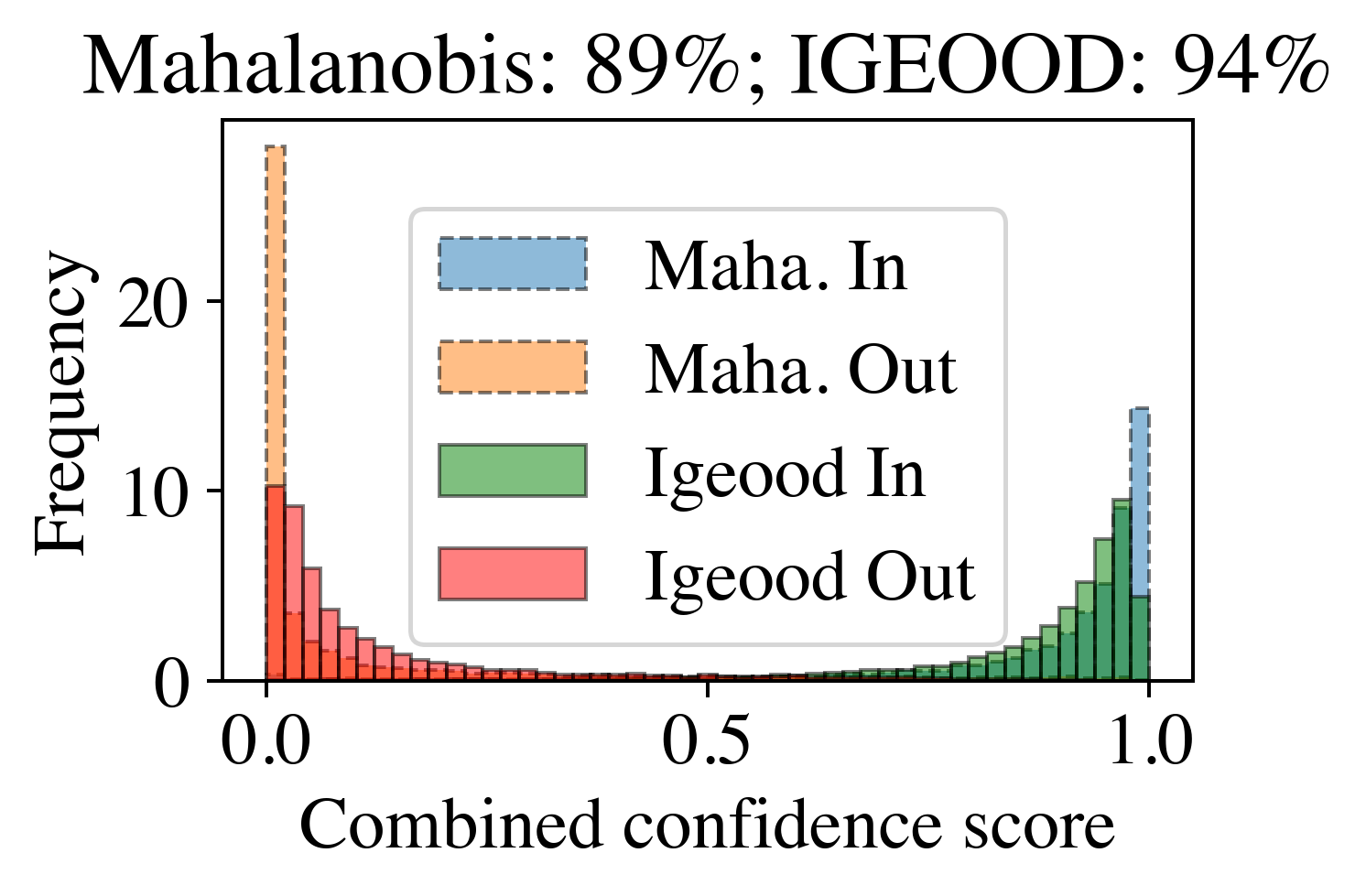}
         \caption{Logistic regression result.}
         \label{fig:hist_combined}
     \end{subfigure}
     \caption{\textcolor{black}{Histograms of the Mahalanobis and \textsc{Igeood} scores for the output of each hidden block of a DenseNet model for CIFAR-10 (in-dstribution) and SVHN (out-of-distribution). The title shows the TNR at TPR-95\% considering only the scores of the outputs of the given layer. The logistic regression found as coefficients: $\boldsymbol{\alpha}=\left(1.0, -3.6, -0.13\right)$ for Mahalanobis and $\boldsymbol{\alpha}=\left(1.0, 1.3, 1.2\right)$ for \textsc{Igeood}.}}
    \label{fig:hist_maha_fr}
\end{figure}

\textcolor{black}{In order to investigate which hidden feature assists the most in OOD detection, we calculate the TNR at TPR-95\% for the scores in the outputs of Blocks 1, 2 and 3 of a DenseNet pre-trained on CIFAR-10. We took as OOD data the SVHN dataset. Figure \ref{fig:hist_maha_fr} shows the histogram and detection performance for each layer as well as the results from the logistic regression. Note that for the \textsc{Igeood} score in this study, we did not consider the logits.}

\section{Detailed Experimental Setup}\label{sec:sup_experimental_setup}

\subsection{DNN models and training details}

We describe the DNN models used in the experiments:
\begin{itemize}
    \item \textbf{DenseNet.} Densely Connected Convolutional Networks \citep{densenet}, or DenseNet for short,  are compositions of dense blocks, which are composed of multiple layers directly connected to every other layer in a feed-forward fashion. In this work, we use the DenseNet-BC-100 architecture. The BC stands for a model with 1x1 convolutional bottleneck (B) layers and channel number compression (C) of 0.5. The models have depth $L=100$ and growth rate $k=12$. We consider the outputs of each dense block after the transition layer (3 in total) and the first convolutional layer output as the latent features. After an averaging pooling, the latent features have dimensions $\boldsymbol{\mathcal{F}}_1=\{24, 108, 150, 342\}$.
    
    \item \textbf{ResNet.} Residual Networks \citep{resnet}, or ResNet, are deep neural networks composed of residual blocks. Each residual block is composed of layers connected in a feed-forward manner plus a skip connection. We use the ResNet with 34 layers pre-trained on CIFAR-10, CIFAR-100, and SVHN datasets. We take the output of every residual block (4 in total) and the first convolutional layer for calculating the score on the \textsc{White-Box} setting. After an averaging pooling, the latent features have dimensions $\boldsymbol{\mathcal{F}}_2=\{64, 64, 128, 256, 512\}$.
\end{itemize}

We train each model by minimizing the cross-entropy loss using SGD with Nesterov momentum equal to 0.9, weight decay equal to 0.0001, and a multi-step learning rate schedule starting at 0.1 for 300 epochs. The pre-trained models is available at \footnote{\url{https://github.com/edadaltocg/Igeood}}. We report their test set accuracy in Table \ref{tab:model_accuracy} with the softmax function and by replacing it with the Fisher-Rao distance between the training class-conditional centroids and the test sample outputs. Also, it is worth noting that one high-end GPU is sufficient for running every experiment presented in this work.

\begin{table}[ht]
  \caption{Test set accuracy in percentage for ResNet and DenseNet architectures pre-trained on CIFAR-10, CIFAR-100 and SVHN.}
  \label{tab:model_accuracy}
    \begin{center}
\begin{tabular}{lcc|cc}
\toprule
    & \multicolumn{2}{c|}{ResNet-34}    & \multicolumn{2}{c}{DenseNet-BC-100} \\
    In-Dataset     & Softmax & Fisher-Rao  & Softmax & Fisher-Rao\\
    \hline
    CIFAR-10  & 93.52 & \textbf{93.53} & 95.20 & 95.20\\
    CIFAR-100 & \textbf{77.11} & 77.09 & 77.62 & \textbf{77.63}\\
    SVHN & 96.61 & 96.61 & 95.16 & 95.16\\
    \bottomrule
  \end{tabular}
    \end{center}
\end{table}

\subsection{Evaluation metrics}
We introduce below standard binary classification performance metrics used to evaluate the OOD discriminators.

\begin{itemize}
    \item \textbf{True Negative Rate at 95\% True Positive Rate (TNR at TPR-95\% (\%)).}  This metric measures the true negative rate (TNR) at a specific true positive rate (TPR). The operating point is chosen such that the TPR of the in-distribution test set is fixed to some value, 95\% in this case. Mathematically, let TP, TN, FP, and FN denote true positive, true negative, false positive and false negative, respectively. We measure $\text{TNR} = \text{TN} / (\text{FP}+\text{TN})$, when $\text{TPR}=\text{TP}/(\text{TP}+ \text{FN})$ is 95\%.

    \item \textbf{Area Under the Receiver Operating Characteristic curve (AUROC (\%)).} The ROC curve is constructed by plotting the true positive rate (TPR) against the false positive rate ($= \text{FP} / (\text{FP}+\text{TN})$) at various threshold values. The area under this curve tells how much the OOD discriminator can distinguish in-distribution and OOD data in a threshold-independent manner.

    \item \textbf{Area Under the Precision-Recall curve (AUPR (\%)).} The PR curve plots the precision ($= \text{TP} / (\text{TP}+\text{FP})$) against the recall ($= \text{TP} / (\text{TP}+\text{FN})$) by varying a threshold. For the experiments, in-distribution data are specified as positives while OOD data as negative.
\end{itemize}

Note that the TNR at TPR-95\% is significant because we want to identify OOD data and preserve a sufficiently good performance on identifying in-distribution data, which is not the case for the other metrics.

\subsection{Datasets}
We use natural image examples from the following image classification and synthetic datasets in our experiments. We normalize the test samples with the in-distribution dataset statistics.

\begin{itemize}
    \item \textbf{CIFAR-10.} The CIFAR-10 \citep{cifar10} dataset is composed of $32 \times 32$ natural images of 10 different classes,  e.g., airplane, ship, bird, etc. The training set comprises 50,000 images, and the test set is composed of 10,000 images. The classes are approximately equally distributed (5,000 examples each label). The CIFAR-10 dataset is under the MIT license.  
    
    \item \textbf{CIFAR-100.} The CIFAR-100 \citep{cifar10} dataset contains similar natural images to the CIFAR-10 dataset, but with 90 additional categories. Its set repartition is 50,000 for training and 10,000 for the test set. We expect around 500 samples for each class of the training set. It is also under the MIT license.
    
    \item \textbf{SVHN. } The SVHN \citep{svhn} dataset collects street house numbers for digit classification. It contains 73,257 training and 26,032 test RGB images of size $32\times32$ of printed digits (from 0 to 9). We take only the first 10,000 examples of the test set for evaluating the methods to have a balanced dataset of in-distribution and out-of-distribution data. This dataset is subject to a non-commercial license.
    
    \item \textbf{Tiny-ImageNet.} The Tiny-ImageNet \citep{tiny-imagenet} dataset is a subset of the large-scale natural image dataset ImageNet \citep{imagenet}. It contains 200 different classes and 10,000 test examples. We downsize the images from their original resolution to images of dimension $32 \times 32 \times 3$.
    
    \item \textbf{LSUN.} The LSUN \citep{lsun} dataset, which has equally 10,000 test examples, is used for the large-scale scene classification of different scene categories (e.g.,  bedroom, bridge, kitchen, etc.). Similarly, we resize the images following the same procedure for the Tiny-ImageNet dataset. LSUN is under the Apache 2.0 license.
    
    \item \textbf{iSUN.} The iSUN \citep{isun} dataset consists of selected natural scene images from the SUN \citep{sun} dataset. The test set has 8925 images, which we downsample to $32\times32\times 3$. We use this dataset as a source of OOD for validation purposes as an independent dataset from the test OOD data.
    
    \item \textbf{Textures.} The Describable Textures Dataset (DTD) \citep{cimpoi14describing} is a collection of textural pattern images observed in nature. It contains 47 categories totaling 5640 images of various sizes, which are resized and center cropped to fit into the input size of $32\times32$.
    
    \item \textbf{Chars74K.} The Chars74K dataset \citep{deCampos09} contains 74,000 samples of 62 classes of characters found in natural images, handwritten text, and synthesized from computer fonts. We used as OOD data only the \textit{EnglishImg} dataset split, which contains 7705 characters from natural scenes. We resized and center-cropped the images.
    \item \textbf{Places365.} The Places365 dataset \citep{zhou2017places} contains images of 365 natural scenes categories. We used the small images validation split as OOD data in our experiments. It contains 36,500 RGB images  which were downsampled from $256\times256$ to $32\times32$.
    \item \textbf{Gaussian.} For the Gaussian dataset, we generated 10,000 synthetic RGB images from 2D Gaussian noise, where each RGB pixel is sampled from an i.i.d Gaussian distribution with mean 0.5 and variance 1.0. The pixel values are clipped to $\left[0,1\right]$ interval. This synthetic data was introduced in previous work as an easy benchmark \citep{baseline}.
\end{itemize}

\subsection{Adversarial data generation}

We generate adversarial samples from the in-distribution dataset using the fast gradient sign method (FGSM). This method works by exploiting the gradients of the neural network to create a non-targeted adversarial attack. For an input image $\boldsymbol{x}_i$, the method computes the sign of the gradients of the loss function $J$ with respect to the input image to create a new image $\boldsymbol{x}_i^{\rm adv}$ that maximizes the loss as given by \eqref{eq:fgsm}. This fabricated image is called an adversarial image, which we use for tuning the hyperparameters of the OOD detection methods in the \textsc{White-Box} case. Mathematically,
\begin{equation}\label{eq:fgsm}
    \boldsymbol{x}_i^{\rm adv} = \boldsymbol{x}_i + \varepsilon^{\rm adv}\odot \text{sign}(\nabla_{\boldsymbol{x}_i}J(\boldsymbol{\theta}, \boldsymbol{x}_i, y_i)),
\end{equation}
where $\varepsilon^{\rm adv}>0$ is the additive noise magnitude parameter. Table \ref{tab:adversarial_table} shows the resulting $L_\infty$ mean perturbation and classification accuracy on adversarial samples.

\begin{table}[ht]
    \caption{The $L_\infty$ mean perturbation used to generate adversarial data with FGSM algorithm and classification accuracy on adversarial samples for the DNN models and in-distribution datasets.}
    \label{tab:adversarial_table}
    \begin{center}
    \begin{tabular}{l cc | cc | cc}
    \toprule
        & \multicolumn{2}{c|}{CIFAR-10} & \multicolumn{2}{c|}{CIFAR-100} & \multicolumn{2}{c}{SVHN} \\
        & $L_\infty$ & Acc. & $L_\infty$ & Acc. & $L_\infty$ & Acc. \\ 
        \hline
        DenseNet-BC-100 & 0.21 & 19.5\%& 0.20 & 4.45\% & 0.32 & 54.7\% \\
        ResNet-34 & 0.21 & 23.7\% & 0.20 & 12.49\% & 0.25 & 50.0\% \\
    \bottomrule
    \end{tabular}
    \end{center}
\end{table}

\section{Benchmark methods}\label{sec:sup_benchmark_methods}

This section briefly introduces the benchmark OOD detection methods with a standardized notation.

\subsection{Baseline}
DNNs tend to assign lower confidence for OOD samples. So, calculating the Maximum Softmax Probability (MSP) \citep{baseline} is a natural baseline for OOD detection. In other words, provided an input data $\boldsymbol{x}$, a pre-trained neural network $f(\cdot)$,  and a confidence threshold $\delta$, the OOD score, and the discriminator are given by 
\begin{equation}
s(\boldsymbol{x}) = \underset{y\in\mathcal{Y}}{\max} \frac{e^{f_{y}(\boldsymbol{x})}}{\sum_{y^\prime \in \mathcal{Y}} e^{f_{y^\prime}(\boldsymbol{x})}} \ \ \text{ and } \ \
S(\boldsymbol{x} ; \delta)=\left\{
\begin{array}{ll}
1 & \text { if } s(\boldsymbol{x}) \leq \delta \\
0 & \text { if } s(\boldsymbol{x})>\delta
\end{array}\right.,
\end{equation}
respectively. Here, $f_y(\boldsymbol{x})$ indicates the $y$-th logits output. A limitation of this method is that unscaled softmax posterior distributions are usually spiky, i.e., softmax trained deep neural models are incorrectly calibrated, which does not favor OOD detection \citep{predictive_uncertainty}.

\subsection{ODIN: OOD detector for Neural networks}
In summary, ODIN \citep{odin} explores the weaknesses of the MSP criterion by recalibrating the output's confidence to the task of OOD detection. They improve the MSP baseline by using the temperature scaled softmax function (\eqref{eq:softmax}) instead. Also, ODIN adds small adversarial noise perturbation to the inputs, i.e.,
\begin{equation}
\widetilde{\boldsymbol{x}}=\boldsymbol{x}-\varepsilon \odot \operatorname{sign}\left( -\nabla_{\boldsymbol{x}} \log q_{\boldsymbol{\theta}}(y | f(\boldsymbol{x}) ;T )\right),
\end{equation}
where $\varepsilon$ is the perturbation magnitude. Hyperparameters $T$ and $\varepsilon$ are tuned on a validation dataset without requiring prior knowledge of test OOD data. They calculate the confidence score by taking the maximum of the perturbed input temperature scaled softmax outputs. 

\subsection{Energy-based OOD detector}

An energy-based OOD discriminator is proposed by \citet{liu2020energybased}, where the differences of energies between in-distribution and OOD samples allow for distribution distinction. The energy-based model substitutes the softmax function with the Helmholtz free energy equation to extract a confidence score. They observed that examples with higher energy have a low likelihood of occurrence, concluding that they are likely OOD. The free energy expression is:
\begin{equation}
E(\boldsymbol{x} ; f)=-T \cdot \log \sum_{y \in \mathcal{Y}} e^{f_{y}(\boldsymbol{x}) / T}.
\end{equation}
Note that, differently from ODIN and MSP, they use the information of all of the logits output values through the sum operation. Besides, they apply input pre-processing for further separating OOD data from in-distribution. 

\subsection{Mahalanobis distance-based confidence score}

The Mahalanobis-based method in \citet{mahalanobis} fits the DNN training data features as class-conditional Gaussian distributions. These use the outputs of every DNN latent block to leverage useful information for discrimination. For a test sample $\boldsymbol{x}$, the confidence score from the $\ell$-th feature is calculated based on the Mahalanobis distance between $f^{(\ell)}(\boldsymbol{x})$ and the closest class-conditional distribution:
\begin{equation}
M_{\ell}(\boldsymbol{x})=\max _{y}-\left(f^{(\ell)}(\boldsymbol{x})-\widehat{\boldsymbol{\mu}}_{y}^{(\ell)}\right)^{\top} \widehat{\Sigma}_{\ell}^{-1}\left(f^{(\ell)}(\boldsymbol{x})-\widehat{\boldsymbol{\mu}}_{y}^{(\ell)}\right),
\end{equation}
where $f^{(\ell)}(\boldsymbol{x})$ is the $\ell$-th latent feature output, and $\widehat{\boldsymbol{\mu}}_{y}^{(\ell)}$ and $\widehat{\Sigma}_{\ell}$ are, respectively, the empirical class mean and covariance matrix estimates. The covariance matrix is often not full rank, so the pseudo-inverse is calculated instead of the inverse. In addition, input pre-processing and feature ensemble are also used to boost performance. A logistic regression model learns the multiplicative weights $\alpha_{\ell}$ for each layer score, which predicts $1$ for in-distribution and $0$ for OOD examples from a mixture validation dataset. Finally, the Mahalanobis-based discriminator is given by thresholding expression $\sum_\ell \alpha_\ell M_\ell(\boldsymbol{x})$.

\section{Additional Out-Of-Distribution detection results}\label{sec:sup_extended_ood_results}

\subsection{\textcolor{black}{Fisher-Rao distance versus Kullback-Leibler Divergence}}\label{sec:fr_vs_kl}
\textcolor{black}{
From \citet{Picot2021AdversarialRV}, the Kullback-Leibler divergence ($\textrm{KL}$) is connected to the Fisher-Rao distance between softmax probability distributions ($d_{R, \mathcal{D}}$) by the inequality:
\begin{equation}\label{eq:kl_rao}
1-\cos \left(\frac{d_{R, \mathcal{D}}\left(q_{\boldsymbol{\theta}}, q_{\boldsymbol{\theta}}^{\prime}\right)}{2}\right) \leq \frac{1}{2} \textrm{KL}\left(q_{\boldsymbol{\theta}}, q_{\boldsymbol{\theta}}^{\prime}\right). 
\end{equation}
To verify how the KL divergence would behave for OOD detection, we ran experiments with our \textsc{Black-Box} setting, where we calculated the class conditional centroids with the KL divergence. We calculated the divergence of the test sample w.r.t each of these centroids during test time, then aggregated the results with a sum or by taking the minimal value. The results are displayed in Table \ref{tab:kl}. We can conclude from these experiments that taking the sum of the outputs instead of the minimal value is overall advantageous for Fisher-Rao distance and KL divergence.
}
\begin{table}[ht]
\begin{center}
    \caption{\textcolor{black}{Performance comparison between the Fisher-Rao distance and the KL Divergence for OOD detection in a \textsc{Black-Box} setting. The numerical values in the Table are TNR at TPR-95\% in percentage for a DenseNet and ResNet models pre-trained on CIFAR-10, CIFAR-100 and SVHN datasets. \textsc{Fisher-Rao} (sum) corresponds to the \textsc{Igeood} score.}}
    \label{tab:kl}
\begin{tabular}{>{\color{black}}c>{\color{black}}l>{\color{black}}c>{\color{black}}c>{\color{black}}c}
\toprule
    & \multirowcell{2}{OOD\\dataset} & CIFAR-10 & CIFAR-100 & SVHN \\
   & & \multicolumn{3}{>{\color{black}}c}{\textsc{Fisher-Rao} (sum) / \textsc{Fisher-Rao} (min)  / KL (min) / KL (sum)} \\
    \midrule
\multirow{9}{*}{\rotatebox[origin=c]{90}{DenseNet}} & Chars &  55.1/45.0/\textbf{56.9}/54.6 & 17.2/14.6/20.1/17.1 & 47.9/46.6/\textbf{50.1}/46.9 \\
& Gaussian &  \textbf{99.9}/97.9/97.9/\textbf{99.9} &      0.0/0.0/0.0/0.0 & 98.0/97.2/73.1/\textbf{98.1} \\
& TinyImgNet &  87.8/73.6/72.3/\textbf{88.1} & \textbf{25.7}/18.1/15.7/25.4 & \textbf{85.1}/84.1/69.4/85.0 \\
& LSUN &  93.3/81.9/86.4/\textbf{93.4} & \textbf{25.4}/17.8/15.1/25.2 & \textbf{85.0}/83.9/66.2/85.4 \\
& Places365 &  52.2/49.7/\textbf{57.2}/51.5 & 20.7/\textbf{20.9}/18.2/20.8 & \textbf{71.9}/71.1/59.1/71.3 \\
& Textures &  35.8/46.9/\textbf{51.0}/34.8 & 22.8/19.5/17.6/\textbf{23.1} & 56.\textbf{5/57.}4/65.3/55.4 \\
& CIFAR-10  &  - & 17.0/\textbf{20.0}/18.5/17.7 & \textbf{67.0}/66.0/56.2/65.8 \\
& CIFAR-100 &  \textbf{50.8}/48.7/49.6/50.3 & - & 65.4/\textbf{65.0}/59.1/64.1\\
& SVHN &  50.1/47.9/\textbf{50.7}/49.6 & \textbf{36.7}/29.9/29.1/35.9 & - \\
\hline
& average &  \textbf{65.6}/61.4/65.3/65.3 & \textbf{20.7}/17.6/16.8/20.6 & \textbf{72.1}/71.4/62.3/71.5 \\
    \midrule
\multirow{9}{*}{\rotatebox[origin=c]{90}{ResNet}}               
                & Chars &  \textbf{51.1}/45.0/41.6/49.6 & 15.2/14.6/14.5/\textbf{15.5} &  \textbf{58.5}/57.4/46.2/58.4  \\
                & Gaussian &  \textbf{89.0}/86.4/62.3/86.8 &    0.6/1.7/\textbf{3.9}/0.9 & 87.3/\textbf{87.5}/76.7/87.0\\
               & TinyImageNet &  \textbf{58.2}/51.4/51.4/57.8 & \textbf{23.0}/17.8/10.4/21.6 &  \textbf{82.2}/81.6/68.8/81.9 \\
               & LSUN &  \textbf{62.0}/53.8/56.0/\textbf{62.0} & \textbf{20.6}/15.4/10.2/19.5  &  77.4/\textbf{77.5}/64.6/77.2 \\
               & Places365 &  \textbf{48.2}/40.0/39.6/48.1 &  16.9/17.3/16.7/\textbf{17.8} &  79.0/\textbf{79.1}/67.2/78.8 \\
               & Textures &   \textbf{50.3}/44.0/45.6/49.8  &  \textbf{23.4}/20.9/14.1/23.2 & \textbf{80.9}/\textbf{80.9}/72.8/80.6 \\
                & CIFAR-10 &  -                   &  18.0/\textbf{18.1}/16.8/18.8 & \textbf{81.2}/81.0/67.8/81.1 \\
               & CIFAR-100 &  \textbf{45.9}/38.9/36.8/45.6 & - &  \textbf{80.2}/79.8/66.2/79.9 \\
               & SVHN &  \textbf{48.8}/31.6/31.5/47.0 &  13.3/14.3/\textbf{15.7}/14.3 & - \\
\hline
                 & average &  \textbf{56.7}/48.9/45.6/55.8  &  \textbf{16.4}/15.0/12.8/\textbf{16.4} &  \textbf{78.3}/78.1/66.3/78.1 \\
\bottomrule
\end{tabular}
\end{center}
\end{table}

\subsection{Hyperparameters tuning}

For temperature $T$, we ran a Bayesian optimization for 500 epochs in the interval of temperature values between 1 and 1000, where the objective function was to maximize the TNR at TPR-95\% metric for the validation set. We took the best temperature among five runs with different random seeds. For the input pre-processing noise magnitude $\varepsilon$ tuning, we ran a grid search optimization with 21 equally spaced values in the interval $\left[0, 0.002\right]$. Table \ref{tab:best_hyperparameters} shows the best hyperparameters we found for the methods in the \textsc{Black-Box}, \textsc{Grey-Box}, and \textsc{White-Box} settings. 

\begin{table}[ht]
    \caption{Best temperatures $T$ for the \textsc{Black-Box} setup, best temperature and noise magnitude ($T$,~$\varepsilon$) for the \textsc{Grey-Box} setup, and best $\varepsilon$ for the Mahalanobis score and $(T,~\varepsilon)$ for \textsc{Igeood} and \textsc{Igeood+} in the \textsc{White-Box} setup with adversarial tuning.}
    \label{tab:best_hyperparameters}
\begin{center}
    \resizebox{\textwidth}{!}{%
    \begin{tabular}{ccccc|ccc|cc}
    \toprule
         & \multirowcell{2}{In-dist.\\dataset} & \multicolumn{3}{c|}{\textsc{Black-Box}} & \multicolumn{3}{c|}{\textsc{Grey-Box}} & \multicolumn{2}{c}{\textsc{White-Box}}\\
        Model &  & ODIN & Energy & \textsc{Igeood} & ODIN & Energy & \textsc{Igeood} & Maha. & \textsc{Igeood,+} \\
        \hline 
        \multirowcell{3}{DenseNet} & C-10 & 1000 & 4.6 & 5.3 & (1000, 0.0014) & (4.6, 0.0012) & (5.3, 0.0012) & 0 & (5, 0.0015) \\
        &  C-100 & 1000 & 1.1 & 2.1 & (1000, 0.0020) & (1.1, 0.0020) & (2.1, 0.0020) & 0 & (5, 0)\\
        & SVHN & 1 & 1.1 & 1.1 & (1, 0.0010) & (1.1, 0.0006) & (1.1, 0.0006) & 0.001 & (5, 0.0015)\\
        \hline
        \multirowcell{3}{ResNet} & C-10 & 1000 & 5.4 & 5.3 & (1000, 0.0014) & (5.4, 0.0012) & (5.3, 0.0012) & 0.0005 & (2, 0)\\
        &  C-100 & 1000 & 1 & 1  & (1000, 0.0020) & (9.1, 0.0024) & (12.7, 0.0024) & 0.0005 & (1, 0)\\
        & SVHN & 1000 & 1.7 & 1 & (1000, 0.0004) & (1.7, 0.0002) & (1.0, 0.0004) & 0 & (5, 0)\\
        \bottomrule
    \end{tabular}
    }
\end{center}
\end{table}

\subsection{Temperature scaling and noise magnitude plots}

\begin{figure}[ht]
    \centering
     \begin{subfigure}[b]{1\textwidth}
         \centering
         \includegraphics[width=\textwidth]{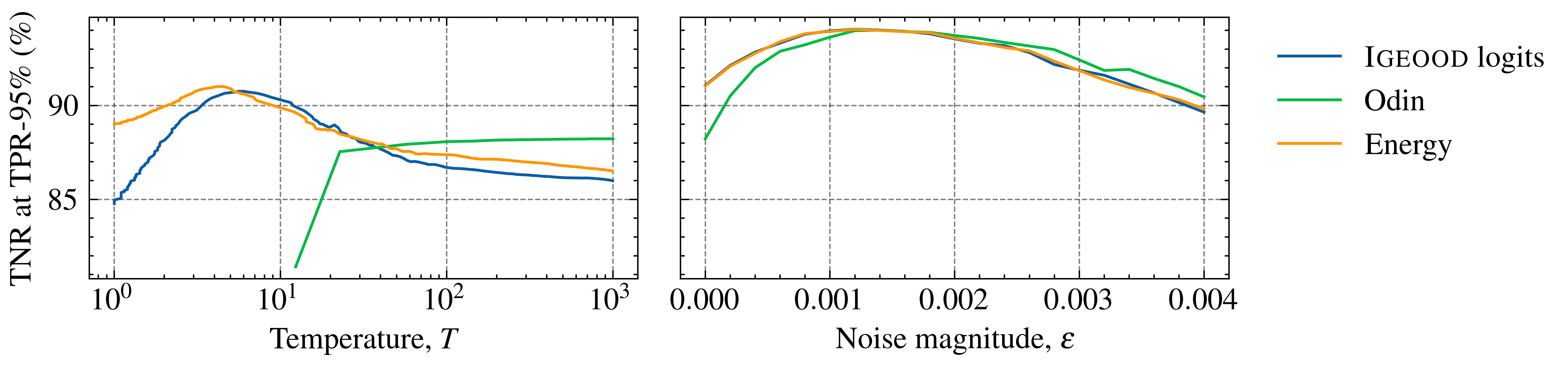}
         \caption{DenseNet on CIFAR-10.}
         \label{fig:t_eps_densenet10}
     \end{subfigure}
     \begin{subfigure}[b]{1\textwidth}
         \centering
         \includegraphics[width=\textwidth]{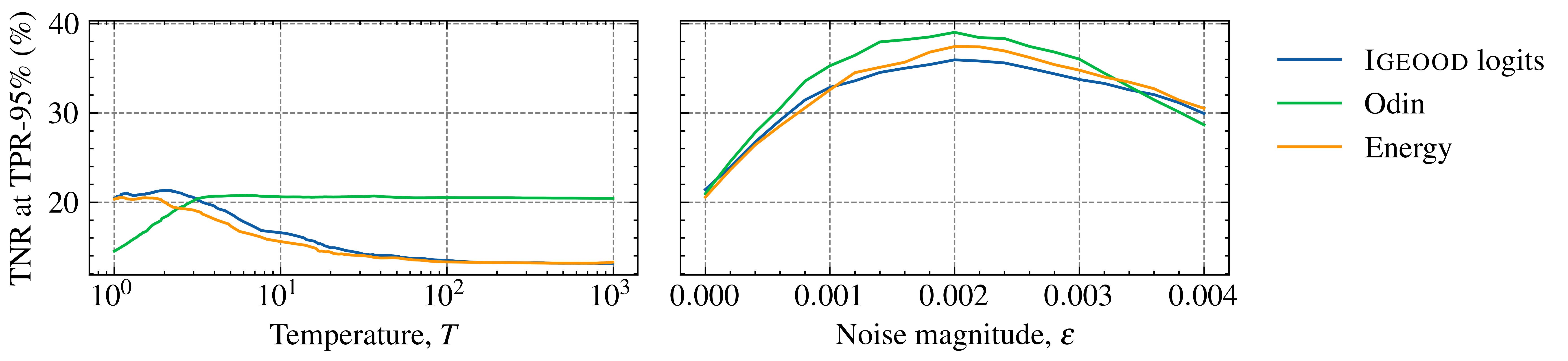}
         \caption{DenseNet on CIFAR-100.}
         \label{fig:t_eps_densenet100}
     \end{subfigure}
     \begin{subfigure}[b]{1\textwidth}
         \centering
         \includegraphics[width=\textwidth]{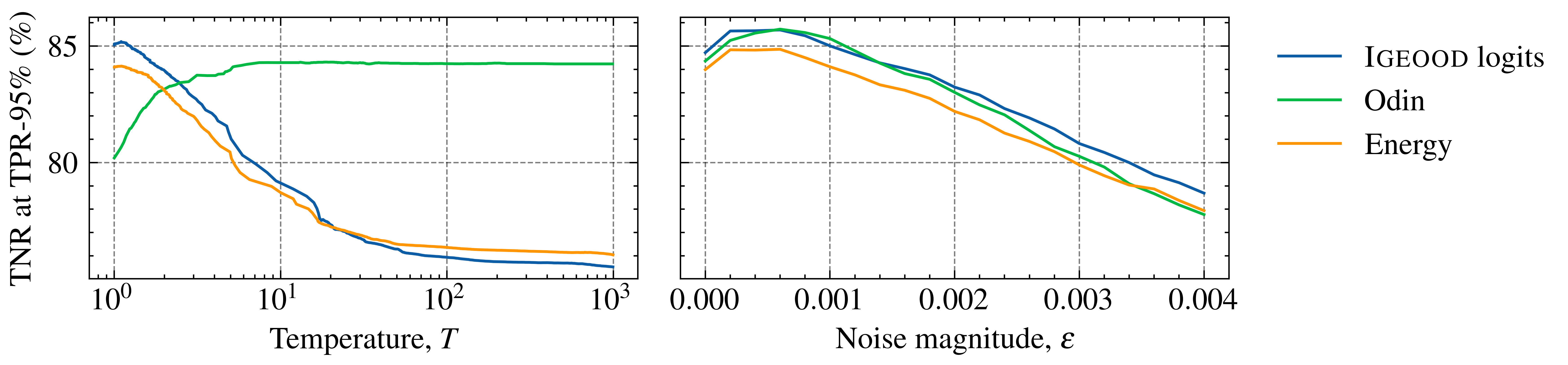}
         \caption{DenseNet on SVHN.}
         \label{fig:t_eps_densenet_svhn}
     \end{subfigure}
         \caption{OOD detection performance against temperature and noise magnitude parameters for ODIN \citep{odin}, Energy \citep{liu2020energybased} and \textsc{Igeood} (ours) on the iSUN \citep{isun} OOD dataset for a DenseNet-100 architecture.}
    \label{fig:t_eps_effect_1}
\end{figure}

\begin{figure}[ht]
    \centering
     \begin{subfigure}[b]{1\textwidth}
         \centering
         \includegraphics[width=\textwidth]{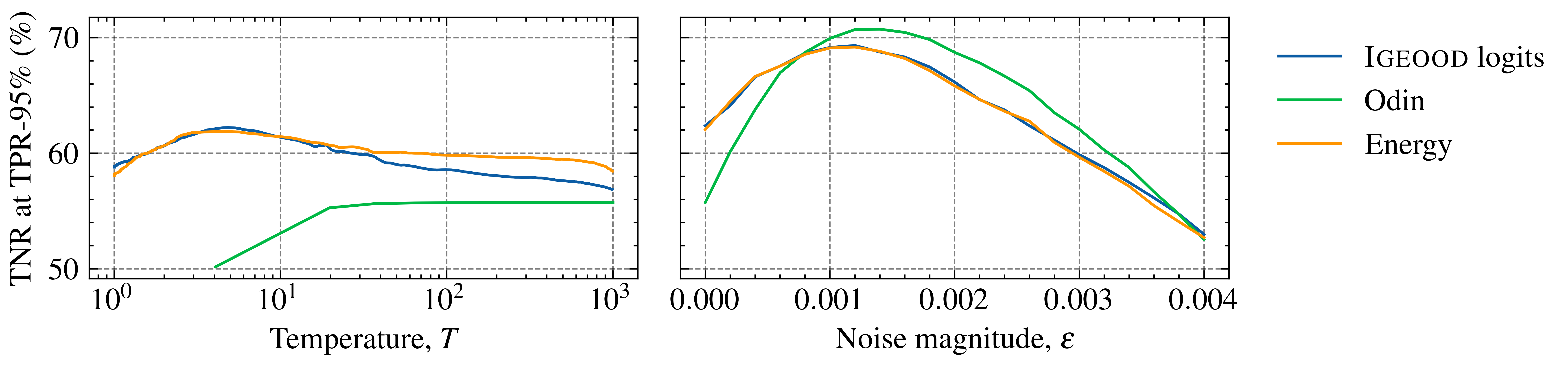}
         \caption{ResNet on CIFAR-10.}
         \label{fig:t_eps_resnet10}
     \end{subfigure}
     \begin{subfigure}[b]{1\textwidth}
         \centering
         \includegraphics[width=\textwidth]{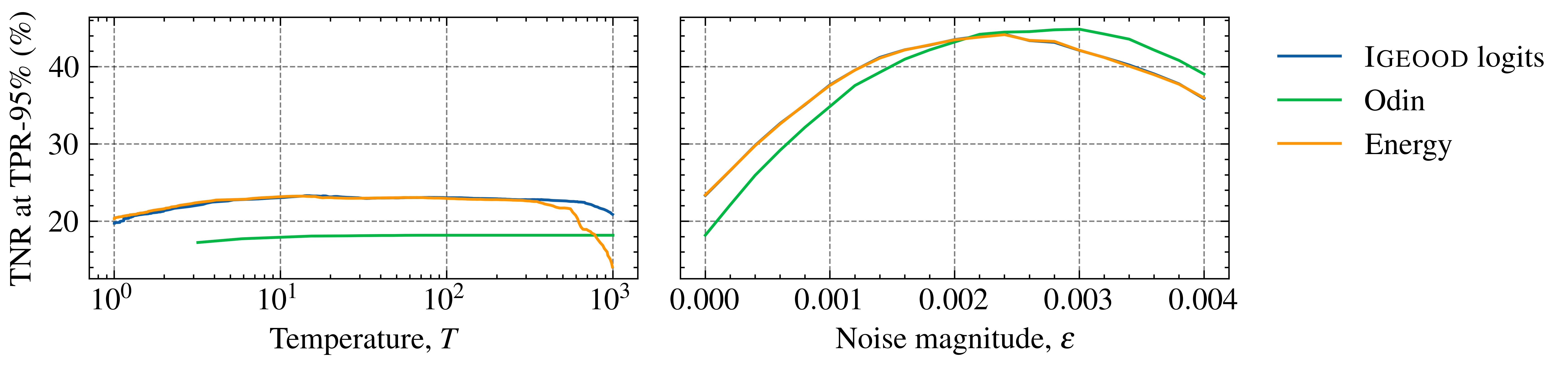}
         \caption{ResNet on CIFAR-100.}
         \label{fig:t_eps_resnet100}
     \end{subfigure}
     \begin{subfigure}[b]{1\textwidth}
         \centering
         \includegraphics[width=\textwidth]{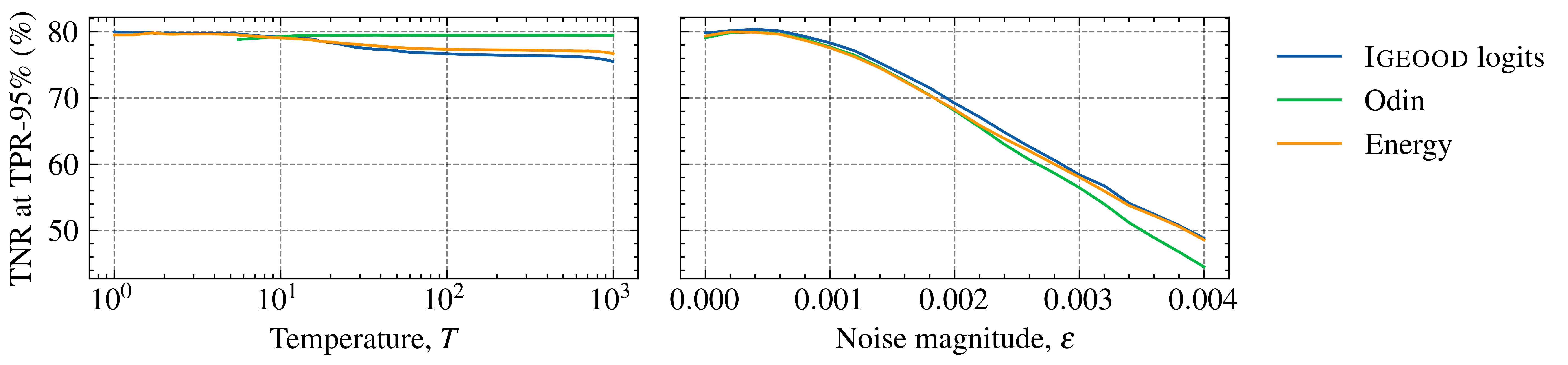}
         \caption{ResNet on SVHN.}
         \label{fig:t_eps_resnet_svhn}
     \end{subfigure}
         \caption{Temperature and noise magnitude tuning for OOD detection performance for ODIN \citep{odin}, Energy \citep{liu2020energybased} and \textsc{Igeood} (ours) on iSUN \citep{isun} OOD dataset for a ResNet-34 architecture.}
    \label{fig:t_eps_effect_2}
\end{figure}

In Figure \ref{fig:t_eps_effect_1} and \ref{fig:t_eps_effect_2}, we plot on the left hand side column the effect of the temperature parameter in the performance for the \textsc{Black-Box} setup. We set the noise magnitude to zero and measured the TNR at TPR-95\% for 500 different temperatures values found by a Bayesian optimization for a variety of DNN models. The performance is evaluated on the iSUN dataset. The right hand side column of Figure \ref{fig:t_eps_effect_1} and \ref{fig:t_eps_effect_2} show the effect of the noise magnitude parameter in the performance of \textsc{Igeood} score in the \textsc{Grey-Box} setup. We set the temperature to the best found in the \textsc{Black-Box} case. Then, we measured the OOD performance for 21 values of noise magnitude $\varepsilon$ equally spaced in the interval $\left[0, 0.004\right]$. The best couple $(T, \varepsilon)$ for each method and model is used to evaluate the \textsc{Grey-Box} performances. The best hyperparameters found are detailed in Table \ref{tab:best_hyperparameters}.

\subsection{\textcolor{black}{Consistency of \textsc{Igeood} score concerning the choice of the  validation data}}

\textcolor{black}{To verify the consistency of \textsc{Igeood} and other methods to the choice of validation data}, we measured the TNR at TPR-95\% after tuning our method in a \textsc{Black-Box} and \textsc{Grey-Box} scenario on nine validation datasets. In Table \ref{tab:robust}, the first column shows the validation dataset, while we used the remaining OOD datasets to evaluate performance. We obtained consistent results, ranging from 63.4\% to 72.0\% the average TNR at TPR-95\% in the \textsc{Black-Box} case and from 65.0\% to 73.4\% in the \textsc{Grey-Box} setting. We show that input pre-processing provides mild amelioration for our method and can be considered a fine-tuning step.

\begin{table}[ht]
    \caption{\textsc{Black-Box} and \textsc{Grey-Box} settings average performance across different OOD datasets for validation. The hyperparameters are tuned using one validation dataset (column 1), and evaluation is done on the remaining eight OOD test datasets. The DNN is DenseNet-BC-100 pre-trained on CIFAR-10, and the values are TNR at TPR-95\% in percentage.}
    \label{tab:robust}
    \begin{center}
    \begin{tabular}{lcccc|ccc}
    \toprule
     & \multicolumn{4}{c|}{\textsc{Black-Box}} & \multicolumn{3}{c}{\textsc{Grey-Box}}\\
    Validation set &  Baseline & ODIN & Energy & \textsc{Igeood} & ODIN & Energy & \textsc{Igeood}  \\
\midrule
           iSUN & 52.5 &  64.3 &    64.9 &          \textbf{65.6} & \textbf{66.8} &    64.8 &          65.3\\
           Chars & 55.0 &  70.8 &    71.1 &          \textbf{71.4} & 72.5 &    72.0 &          \textbf{73.4} \\
           CIFAR-100 & 55.4 &  68.6 &    69.1 &          \textbf{72.0} & 68.6 &    \textbf{71.7} &          71.3 \\
           Gaussian & 49.4 &  62.8 &    \textbf{65.6} &          63.4 &  \textbf{70.4} &    64.0 &          68.0  \\
           TinyImgNet & 53.0 &  64.7 &    \textbf{65.2} &          63.5 &  \textbf{67.0} &    65.0 &          65.5  \\
           LSUN & 52.1 &  \textbf{63.9} &    63.7 &          63.6  & \textbf{66.6} &    65.3 &          65.0 \\
           Places365 & 55.3 &  68.5 &    69.0 &          \textbf{71.8} &  70.0 &    \textbf{71.5} &          70.9\\
           SVHN & 55.4 &  68.7 &    69.3 &          \textbf{69.5} &  70.0 &    69.4 &          \textbf{70.1} \\
           Textures & 55.4 &  71.2 &    \textbf{73.1} &          71.4 &  71.5 &    \textbf{72.4} &          71.6 \\
        \hline
        average and std. & 53.7\textcolor{black}{\scriptsize{$\pm$2.0}} &  67.1\textcolor{black}{\scriptsize{$\pm$3.0}} &    67.9\textcolor{black}{\scriptsize{$\pm$}3.0} &          \textbf{68.0}\textcolor{black}{\scriptsize{$\pm$}3.7} & \textbf{69.3}\textcolor{black}{\scriptsize{$\pm$}2.0} & 68.4\textcolor{black}{\scriptsize{$\pm$}3.4} & 69.0\textcolor{black}{\scriptsize{$\pm$}3.0}\\
        \bottomrule
    \end{tabular}
    \end{center}
\end{table}

\subsection{Error bars and standard deviation}

We conduct all of our experiments during inference time. Provided that we fix the DNN, the in-distribution, and the out-of-distribution datasets, there is not a source of randomness to our algorithm because the weights $\alpha$ of the feature ensemble method and centroids are initialized deterministically. Thus, the OOD scores for the same experimental setting do not change. To confirm this, we ran the same experiment five times and obtained the same results in all of them. However, if we allow for retraining the DNN from scratch, we might obtain different parameters, leading to slightly different model accuracy and potentially OOD detection performance. With this in mind, we retrained a DenseNet-BC-100 model on CIFAR-10 five times with five different random seeds. The results for OOD detection in a \textsc{Black-Box} setting for the 5 models can be found in Table \ref{tab:seed_model_error}.

\begin{table}
  \begin{center}
    \caption{Experiment using five different training seeds for DenseNet-100 on CIFAR-10 for the \textsc{Black-Box} scenario. The average test accuracy of the 5 models is 94.58\%{\scriptsize $\pm$}0.13\%. All values are percentages.}
    \label{tab:seed_model_error}
    \begin{tabular}{lcc}
      \toprule
      {} &  TNR at TPR-95\% & AUROC \\
       Dataset & \multicolumn{2}{c}{Baseline / ODIN / Energy / \textsc{Igeood}} \\
      \hline
Chars              
&  34.1\textcolor{black}{\scriptsize{$\pm$21}}/54.3\textcolor{black}{\scriptsize{$\pm$20}}/59.0\textcolor{black}{\scriptsize{$\pm$}14}/\textbf{59.4}\textcolor{black}{\scriptsize{$\pm$14}}
&  88.6\textcolor{black}{\scriptsize{$\pm$4.3}}/90.4\textcolor{black}{\scriptsize{$\pm$3.5}}/90.2\textcolor{black}{\scriptsize{$\pm$3.9}}/\textbf{90.6}\textcolor{black}{\scriptsize{$\pm$3.4}} \\
CIFAR-100   
&  37.1\textcolor{black}{\scriptsize{$\pm$0.5}}/\textbf{47.8}\textcolor{black}{\scriptsize{$\pm$1.4}}/44.8\textcolor{black}{\scriptsize{$\pm$2.3}}/45.2\textcolor{black}{\scriptsize{$\pm$2.2}} 
&  88.2\textcolor{black}{\scriptsize{$\pm$0.3}}/\textbf{88.8}\textcolor{black}{\scriptsize{$\pm$0.7}}/88.0\textcolor{black}{\scriptsize{$\pm$1.0}}/88.3\textcolor{black}{\scriptsize{$\pm$0.8}} \\
Gaussian 
&  42.7\textcolor{black}{\scriptsize{$\pm$49}}/74.0\textcolor{black}{\scriptsize{$\pm$28}}/79.6\textcolor{black}{\scriptsize{$\pm$23}}/\textbf{80.2}\textcolor{black}{\scriptsize{$\pm$23}} 
&  93.7\textcolor{black}{\scriptsize{$\pm$4.3}}/96.6\textcolor{black}{\scriptsize{$\pm$2.4}}/\textbf{96.7}\textcolor{black}{\scriptsize{$\pm$1.8}}/\textbf{96.7}\textcolor{black}{\scriptsize{$\pm$1.8}} \\
TinyImgNet     
&  50.6\textcolor{black}{\scriptsize{$\pm$4.3}}/76.1\textcolor{black}{\scriptsize{$\pm$4.4}}/78.3\textcolor{black}{\scriptsize{$\pm$5.0}}/\textbf{78.4}\textcolor{black}{\scriptsize{$\pm$5.0}} 
&  92.5\textcolor{black}{\scriptsize{$\pm$1.0}}/95.7\textcolor{black}{\scriptsize{$\pm$1.0}}/96.0\textcolor{black}{\scriptsize{$\pm$1.1}}/\textbf{96.1}\textcolor{black}{\scriptsize{$\pm$1.0}} \\
LSUN            
&  58.0\textcolor{black}{\scriptsize{$\pm$3.6}}/85.2\textcolor{black}{\scriptsize{$\pm$3.5}}/87.3\textcolor{black}{\scriptsize{$\pm$4.7}}/\textbf{87.5}\textcolor{black}{\scriptsize{$\pm$4.5}} 
&  94.2\textcolor{black}{\scriptsize{$\pm$0.6}}/97.4\textcolor{black}{\scriptsize{$\pm$0.6}}/97.6\textcolor{black}{\scriptsize{$\pm$0.7}}/\textbf{97.7}\textcolor{black}{\scriptsize{$\pm$0.7}} \\
Places365              
&  9.30\textcolor{black}{\scriptsize{$\pm$1.5}}/\textbf{54.2}\textcolor{black}{\scriptsize{$\pm$2.5}}/52.7\textcolor{black}{\scriptsize{$\pm$4.2}}/53.2\textcolor{black}{\scriptsize{$\pm$3.9}} 
&  88.4\textcolor{black}{\scriptsize{$\pm$0.4}}/\textbf{89.9}\textcolor{black}{\scriptsize{$\pm$1.2}}/89.4\textcolor{black}{\scriptsize{$\pm$1.7}}/89.7\textcolor{black}{\scriptsize{$\pm$1.5}} \\
SVHN                   
&  36.0\textcolor{black}{\scriptsize{$\pm$3.0}}/\textbf{48.1}\textcolor{black}{\scriptsize{$\pm$7.1}}/46.1\textcolor{black}{\scriptsize{$\pm$9.9}}/46.5\textcolor{black}{\scriptsize{$\pm$9.6}} 
&  \textbf{86.8}\textcolor{black}{\scriptsize{$\pm$2.0}}/86.6\textcolor{black}{\scriptsize{$\pm$4.8}}/85.9\textcolor{black}{\scriptsize{$\pm$5.7}}/86.4\textcolor{black}{\scriptsize{$\pm$5.0}} \\

Textures               
&  35.6\textcolor{black}{\scriptsize{$\pm$1.7}}/\textbf{38.2}\textcolor{black}{\scriptsize{$\pm$1.4}}/33.3\textcolor{black}{\scriptsize{$\pm$2.8}}/34.0\textcolor{black}{\scriptsize{$\pm$2.6}} 
&  \textbf{87.2}\textcolor{black}{\scriptsize{$\pm$0.6}}/83.3\textcolor{black}{\scriptsize{$\pm$1.0}}/80.8\textcolor{black}{\scriptsize{$\pm$1.9}}/82.1\textcolor{black}{\scriptsize{$\pm$1.3}} \\
\hline
average and std.
&  41.7\textcolor{black}{\scriptsize{$\pm$10}}/59.7\textcolor{black}{\scriptsize{$\pm$8.6}}/60.1\textcolor{black}{\scriptsize{$\pm$8.3}}/\textbf{60.6}\textcolor{black}{\scriptsize{$\pm$8.0}} 
&  90.0\textcolor{black}{\scriptsize{$\pm$1.7}}/\textbf{91.1}\textcolor{black}{\scriptsize{$\pm$1.9}}/90.6\textcolor{black}{\scriptsize{$\pm$2.2}}/90.9\textcolor{black}{\scriptsize{$\pm$1.9}}	 \\
      \bottomrule
    \end{tabular}
  \end{center}
\end{table}

\begin{table}[ht]
    \begin{center}
      \caption{\textcolor{black}{Average and standard deviation OOD detection performance across eight OOD datasets for each model and in-distribution dataset in a \textsc{Grey-Box} setting.} }
      \label{tab:grey_box_avg}
  \begin{tabular}{>{\color{black}}c>{\color{black}}c>{\color{black}}c>{\color{black}}c}
  \toprule
    & & TNR at TPR-95\% & AUROC \\
    Model & In-dist. & \multicolumn{2}{>{\color{black}}c}{ODIN  / Energy / \textsc{Igeood}} \\
      \hline
  \multirowcell{3}{DenseNet} 
& C-10  &  
\textbf{66.8}\textcolor{black}{{\scriptsize$\pm$23}}/64.8\textcolor{black}{{\scriptsize$\pm$25}}/65.3\textcolor{black}{{\scriptsize$\pm$24}}   &   \textbf{91.9}\textcolor{black}{{\scriptsize$\pm$6.2}}/91.5\textcolor{black}{{\scriptsize$\pm$6.4}}/\textbf{91.9}\textcolor{black}{{\scriptsize$\pm$6.0}} \\
& C-100 & 
\textbf{25.5}\textcolor{black}{{\scriptsize$\pm$14}}/24.8\textcolor{black}{{\scriptsize$\pm$13}}/25.0\textcolor{black}{{\scriptsize$\pm$13}}   &   76.6\textcolor{black}{{\scriptsize$\pm$12}}/76.4\textcolor{black}{{\scriptsize$\pm$12}}/\textbf{78.2}\textcolor{black}{{\scriptsize$\pm$8.2}}    \\  
& SVHN        & 
\textbf{75.4}\textcolor{black}{{\scriptsize$\pm$15}}/70.6\textcolor{black}{{\scriptsize$\pm$17}}/72.4\textcolor{black}{{\scriptsize$\pm$16}}   &   \textbf{91.6}\textcolor{black}{{\scriptsize$\pm$5.4}}/89.2\textcolor{black}{{\scriptsize$\pm$6.9}}/90.0\textcolor{black}{{\scriptsize$\pm$6.3}}  \\
\hline
\multirowcell{3}{ResNet} 
& C-10      & 
57.3\textcolor{black}{{\scriptsize$\pm$20}}/57.7\textcolor{black}{{\scriptsize$\pm$19}}/\textbf{57.8}\textcolor{black}{{\scriptsize$\pm$19}}   &   \textbf{89.2}\textcolor{black}{{\scriptsize$\pm$5.4}}/88.7\textcolor{black}{{\scriptsize$\pm$5.3}}/89.0\textcolor{black}{{\scriptsize$\pm$5.2}}   \\
& C-100     & 
\textbf{31.1}\textcolor{black}{{\scriptsize$\pm$22}}/30.2\textcolor{black}{{\scriptsize$\pm$22}}/30.2\textcolor{black}{{\scriptsize$\pm$22}}   &   \textbf{76.9}\textcolor{black}{{\scriptsize$\pm$11}}/74.4\textcolor{black}{{\scriptsize$\pm$12}}/74.3\textcolor{black}{{\scriptsize$\pm$12}}  \\
& SVHN          & 78.5\textcolor{black}{{\scriptsize$\pm$7.8}}/78.5\textcolor{black}{{\scriptsize$\pm$7.9}}/\textbf{78.8}\textcolor{black}{{\scriptsize$\pm$7.8}}   &   90.4\textcolor{black}{{\scriptsize$\pm$3.4}}/\textbf{90.9}\textcolor{black}{{\scriptsize$\pm$3.4}}/90.7\textcolor{black}{{\scriptsize$\pm$3.3}}  \\
\hline
\multicolumn{2}{>{\color{black}}c}{Average and Std.}                    & \textbf{55.8}\textcolor{black}{{\scriptsize$\pm$21}}/54.4\textcolor{black}{{\scriptsize$\pm$20}}/54.9\textcolor{black}{{\scriptsize$\pm$20}}   &   \textbf{86.1}\textcolor{black}{{\scriptsize$\pm$6.7}}/85.2\textcolor{black}{{\scriptsize$\pm$7.0}}/85.7\textcolor{black}{{\scriptsize$\pm$6.8}}      \\
  \bottomrule
  \end{tabular}
    \end{center}
\end{table}

\subsection{\textcolor{black}{\textsc{Igeood} compared to other \textsc{White-Box} methods.}}\label{sec:sup_igeood_vs_others}

\textcolor{black}{Even though \citet{mahalanobis} shares the closest setup to ours, recent literature also shows promising results for OOD detection in a \textsc{White-Box} setting, achieving state-of-the-art in a few benchmarks. Notably, the works from \citet{gram_matrice, Hsu2020GeneralizedOD, Zisselman2020DeepRF} achieve remarkable performance in a range of benchmarks. Thus, we gathered the reported results from the original works and displayed them in Table \ref{tab:other_comparisons} and \ref{tab:other_comparisons_adv}, which considers that a few OOD samples and only adversarial samples are available for tuning, respectively. We highlight that \citet{gram_matrice} extracts, in addition to the outputs of the blocks, intra-block features for the ResNet and DenseNet models.}

\begin{table}[ht]
\begin{center}
    \caption{\textcolor{black}{TNR at TPR-95\% (\%) performance in a \textsc{White-Box} setting considering the original results from \citet{mahalanobis} and \cite{Zisselman2020DeepRF} with access to OOD samples. The models are DenseNet-BC-100 and ResNet-34 pre-trained on CIFAR-10, CIFAR-100 and SVHN.}}
    \label{tab:other_comparisons}
\begin{tabular}{>{\color{black}}c>{\color{black}}l>{\color{black}}c>{\color{black}}c>{\color{black}}c}
\toprule
    & \multirowcell{2}{OOD\\dataset} & CIFAR-10 & CIFAR-100 & SVHN \\
    & & \multicolumn{3}{>{\color{black}}c}{Mahalanobis / Res-Flow / \textsc{Igeood} / \textsc{Igeood+}} \\
    \midrule
\multirow{5}{*}{\rotatebox[origin=c]{90}{DenseNet}}
&  iSUN  &  95.3/ \ \  -  \ \ /97.7/\textbf{99.8} & 87.0/ \ \  -  \ \ /93.8/\textbf{99.7} & \textbf{99.9}/ \ \  -  \ \ /98.3/\textbf{99.9} \\
&  LSUN  &  97.2/98.2/98.5/\textbf{99.9} & 91.4/96.3/95.2/\textbf{99.9} & \textbf{99.9}/\textbf{100}/97.1/\textbf{99.9} \\
&  TinyImgNet  &  95.0/96.4/95.7/\textbf{99.8} & 86.6/93.0/94.5/\textbf{99.5} & \textbf{99.9}/\textbf{100}/98.2/\textbf{99.9} \\
&  SVHN/C-10  &  90.8/94.9/98.9/\textbf{99.9} & 82.5/84.9/93.3/\textbf{99.6} & 96.8/\textbf{99.0}/91.6/98.3 \\
\cline{2-5}
& average & 94.6/96.5/97.7/\textbf{99.8} & 86.9/91.4/94.2/\textbf{99.7} & 99.1/\textbf{99.6}/96.3/\textbf{99.5}  \\
\midrule
\multirow{5}{*}{\rotatebox[origin=c]{90}{ResNet}}
&  iSUN  &  97.8/ \ \  -  \ \ /97.2/\textbf{99.9} & 89.9/ \ \  -  \ \ /93.4/\textbf{99.8} & 99.7/ \ \  -  \ \ /99.8/\textbf{100} \\
&  LSUN  &  98.8/99.0/98.4/\textbf{100} & 90.9/96.2/94.3/\textbf{100} & \textbf{99.9}/\textbf{100}/99.7/\textbf{99.9} \\
&  TinyImgNet  &  97.1/97.8/96.3/\textbf{99.6} & 90.9/94.6/90.1/\textbf{99.6} & \textbf{99.9}/\textbf{100}/99.7/\textbf{99.9} \\
&  SVHN/C-10  &  87.8/96.5/98.8/\textbf{99.8} & 91.9/93.0/91.6/\textbf{99.7} & 98.4/99.4/97.7/\textbf{99.7} \\
\cline{2-5}
& average & 95.4/97.8/97.7/\textbf{99.8} & 90.9/94.6/92.35/\textbf{99.8} & 99.5/99.8/99.2/\textbf{99.9} \\
    \bottomrule
\end{tabular}
\end{center}
\end{table}

\begin{table}[ht]
\begin{center}
    \caption{\textcolor{black}{TNR at TPR-95\% (\%) performance in a \textsc{White-Box} setting considering the original results from \citet{mahalanobis, gram_matrice, Hsu2020GeneralizedOD, Zisselman2020DeepRF} without access to OOD samples for hyperparameter tuning.}}
    \label{tab:other_comparisons_adv}
  \resizebox{\textwidth}{!}{%
\begin{tabular}{>{\color{black}}c>{\color{black}}l>{\color{black}}c>{\color{black}}c>{\color{black}}c}
\toprule
    & \multirowcell{2}{OOD\\dataset} & CIFAR-10 & CIFAR-100 & SVHN \\
    & & \multicolumn{3}{>{\color{black}}c}{Mahalanobis / Gram Matrix / DeConf-C / Res-Flow / \textsc{Igeood} / \textsc{Igeood+}} \\
    \midrule
\multirow{5}{*}{\rotatebox[origin=c]{90}{DenseNet}}
&  iSUN  & 94.3/99.0/\textbf{99.4}/ \ \  -  \ \ /94.5/95.8 & 84.8/95.9/\textbf{98.4}/ \ \  -  \ \ /93.8/92.2 & \textbf{99.9}/99.4/ \ \  -  \ \ / \ \  -  \ \ /98.2/98.6 \\
&  LSUN  &  97.2/\textbf{99.5}/99.4/98.1/96.4/97.2 & 91.4/97.2/\textbf{98.7}/95.8/95.1/94.4 & \textbf{100}/99.5/ \ \  -  \ \ /\textbf{100}/97.3/97.0 \\
&  TinyImgNet  &  94.9/98.8/\textbf{99.1}/96.1/93.4/94.5 & 87.2/95.7/\textbf{98.6}/91.5/94.3/94.0 & \textbf{99.9}/99.1/ \ \  -  \ \ /\textbf{99.9}/98.1/96.8 \\
&  SVHN/C-10  &  89.9/96.1/\textbf{98.8}/86.1/94.3/95.7 & 62.2/89.3/\textbf{95.9}/48.9/90.1/90.6 & \textbf{90.0}/80.4/ \ \  -  \ \ /\textbf{90.0}/89.5/86.6 \\
\cline{2-5}
& average & 94.1/98.3/\textbf{99.2}/93.4/94.6/95.8 & 81.4/94.5/\textbf{97.9}/78.7/93.3/92.8 & \textbf{97.4}/94.6/ \ \  - \ \ /96.6/95.8/94.8 \\
\midrule
\multirow{5}{*}{\rotatebox[origin=c]{90}{ResNet}}
&  iSUN  &  96.8/\textbf{99.3}/88.8/ \ \  -  \ \ /95.3/95.0 & 87.9/\textbf{94.8}/75.3/ \ \  -  \ \ /89.4/91.0 & \textbf{100}/99.4/ \ \  -  \ \ / \ \  -  \ \ /\textbf{99.8}/\textbf{99.9} \\
&  LSUN  &  98.1/\textbf{99.6}/90.9/99.1/97.7/97.7 & 56.6/\textbf{96.6}/76.8/70.4/88.6/93.9 & \textbf{99.9}/99.6/ \ \  -  \ \ /\textbf{100}/\textbf{99.8}/\textbf{100} \\
&  TinyImgNet  &  95.5/\textbf{98.7}/81.4/98.0/94.3/94.2 & 70.3/\textbf{94.8}/76.5/77.5/86.2/90.1 & 99.2/99.3/ \ \  -  \ \ /\textbf{99.9}/\textbf{99.6}/\textbf{99.6} \\
&  SVHN/C-10  &  75.8/97.6/89.5/91.0/\textbf{98.2}/97.7 & 41.9/\textbf{80.8}/55.1/74.1/75.2/78.5 & 94.1/85.8/ \ \  -  \ \ /96.6/96.7/\textbf{97.3} \\
\cline{2-5}
& average & 91.5/\textbf{98.8}/87.6/96.0/96.3/96.2 & 64.2/\textbf{91.7}/71.0/74.0/84.8/88.4 & 98.3/96.0/ \ \ - \  /98.8/\textbf{99.0}/\textbf{99.2} \\
    \bottomrule
\end{tabular}
}
\end{center}
\end{table}

\subsection{Extended OOD detection results}

We show in Table \ref{tab:black_box_extended} extended OOD detection results of Table \ref{tab:black_box_avg}. It contains the OOD detection performance for each model, in-distribution dataset and OOD dataset in a \textsc{Black-Box} setting. In Table \ref{tab:grey_box_extended}, we show the performance of ODIN, energy-based, and \textsc{Igeood} scores in the task of OOD detection in a \textsc{Grey-Box} setup for each OOD dataset. In Table \ref{tab:wb_ood_extended} and \ref{tab:wb_adv_extended}, we show additional results referring to the right-hand column and left-hand column of Table \ref{tab:white_box_avg}, respectively.

\begin{table}[ht]
  \begin{center}
    \caption{Extended \textsc{Black-Box} results for Table\ref{tab:black_box_avg}. Parameter tuning on iSUN dataset.}
    \label{tab:black_box_extended}
  \resizebox{0.9\textwidth}{!}{%

\begin{tabular}{ccccc}
\toprule
    \multirowcell{2}{In-dist.\\(model)} & \multirowcell{2}{OOD\\dataset} & TNR at TPR-95\% & AUROC & AUPR \\
    \cmidrule(r){3-5}
    &  & \multicolumn{3}{c}{Baseline / ODIN / Energy / \textsc{Igeood}} \\
    \hline
    \multirow{8}{*}{\multirowcell{2}{CIFAR-10\\(DenseNet)}} 
& Chars &  43.5/\textbf{57.2}/54.6/55.0 &  90.2/\textbf{91.2}/90.4/90.5 &  93.0/\textbf{93.1}/92.5/92.7 \\
& CIFAR-100 &  40.6/\textbf{53.1}/50.5/50.7 &  89.4/\textbf{90.4}/89.7/89.8 &  90.5/\textbf{90.7}/90.1/90.2 \\
& Gaussian &  88.1/99.8/\textbf{99.9}/\textbf{99.9} &  97.6/\textbf{98.9}/98.5/98.5 &  98.3/\textbf{99.3}/99.1/99.1 \\
& TinyImgNet &  59.4/85.0/\textbf{88.0}/87.8 &  94.1/97.3/\textbf{97.6}/\textbf{97.6} &  95.4/97.7/\textbf{97.9}/\textbf{97.9} \\
& LSUN &  66.9/91.4/\textbf{93.3}/\textbf{93.3} &  95.5/98.3/\textbf{98.5}/\textbf{98.5} &  96.5/98.5/\textbf{98.7}/\textbf{98.7} \\
& Places365 &  40.8/\textbf{54.2}/51.5/52.0 &  88.8/\textbf{90.2}/89.5/89.7 &  74.4/\textbf{74.8}/73.9/74.3 \\
& SVHN &  40.4/\textbf{52.0}/49.6/50.1 &  89.9/\textbf{90.9}/90.2/90.3 &  \textbf{84.6}/\textbf{84.6}/83.5/83.7 \\
& Textures &  40.5/\textbf{42.1}/34.9/35.6 &  \textbf{88.5}/85.1/82.4/83.2 &  \textbf{93.1}/88.3/86.5/87.8 \\
    \cline{2-5}
     & average &  52.5/\textbf{66.8}/65.3/65.6 &  91.7/\textbf{92.8}/92.1/92.3 &  90.7/\textbf{90.9}/90.3/90.6 \\
    \cline{1-5}
    \multirow{8}{*}{\multirowcell{2}{CIFAR-100\\(DenseNet)}} 
& Chars &  15.1/\textbf{17.8}/17.0/17.2 &  72.8/\textbf{78.0}/77.9/77.8 &  79.6/83.8/\textbf{83.9}/83.8 \\
& CIFAR-10 &  17.7/\textbf{18.1}/17.1/17.0 &  \textbf{75.6}/74.8/74.4/75.4 &  \textbf{78.3}/74.3/74.0/76.2 \\
& Gaussian &      0.0/0.0/0.0/0.0 &  30.2/19.4/19.5/\textbf{30.7} &  53.2/44.0/44.1/\textbf{53.5} \\
& TinyImgNet &  16.7/24.7/25.0/\textbf{25.7} &  72.0/79.4/\textbf{79.6}/79.5 &  74.8/80.7/\textbf{80.9}/80.7 \\
& LSUN &  15.5/23.2/24.2/\textbf{25.4} &  70.9/80.4/\textbf{80.8}/80.6 &  74.4/82.6/\textbf{82.9}/82.5 \\
& Places365 &  18.8/21.2/\textbf{20.6}/\textbf{20.6} &  75.9/\textbf{78.0}/77.7/\textbf{78.0} &  54.2/54.5/54.3/\textbf{55.7} \\
& SVHN &  25.7/36.4/36.5/\textbf{36.7} &  82.8/\textbf{88.4}/\textbf{88.4}/88.2 &  75.4/\textbf{82.5}/82.4/82.4 \\
& Textures &  18.0/22.4/22.2/\textbf{22.8} &  72.7/74.4/74.3/\textbf{75.7} &  80.8/79.1/79.0/\textbf{81.5} \\
    \cline{2-5}
     & average &  15.9/20.5/20.3/\textbf{20.7} &  69.1/71.6/71.6/\textbf{73.2} &  71.3/72.7/72.7/\textbf{74.5} \\
    \cline{1-5}
    \multirow{8}{*}{\multirowcell{2}{SVHN\\(DenseNet)}} 
& Chars &  46.4/27.0/45.0/\textbf{47.9} &  \textbf{83.9}/52.2/79.6/80.9 &  \textbf{91.8}/70.2/88.9/89.6 \\
& CIFAR-10 &  61.8/66.0/64.7/\textbf{67.0} &  \textbf{92.3}/90.9/90.3/90.9 &  \textbf{96.2}/95.1/94.6/95.0 \\
& CIFAR-100 &  61.3/64.4/63.0/\textbf{65.4} &  \textbf{91.9}/90.3/89.5/90.3 &  \textbf{95.7}/94.3/93.8/94.3 \\
& Gaussian &  93.6/97.8/97.9/\textbf{98.0} &  97.4/\textbf{98.0}/\textbf{98.0}/\textbf{98.0} &  99.2/\textbf{99.4}/\textbf{99.4}/\textbf{99.4} \\
& TinyImgNet &  80.4/84.4/84.1/\textbf{85.1} &  \textbf{95.5}/95.3/94.9/95.3 &  \textbf{97.9}/97.4/97.2/97.5 \\
& LSUN &  80.1/84.4/84.3/\textbf{85.0} &  \textbf{95.5}/95.3/95.1/95.3 &  \textbf{98.0}/97.6/97.5/97.6 \\
& Places365 &  66.8/71.0/69.9/\textbf{71.9} &  \textbf{93.0}/91.9/91.3/91.9 &  \textbf{89.1}/85.8/84.6/85.8 \\
& Textures &  56.4/55.2/52.4/\textbf{56.5} &  \textbf{88.9}/84.6/82.5/84.9 &  \textbf{95.5}/93.3/92.2/93.4 \\
    \cline{2-5}
     & average &  68.3/68.8/70.2/\textbf{72.1} &  \textbf{92.3}/87.3/90.2/90.9 &  \textbf{95.4}/91.6/93.5/94.1 \\
    \cline{1-5}
    \multirow{8}{*}{\multirowcell{2}{CIFAR-10\\(ResNet)}} 
& Chars &  36.8/45.8/50.7/\textbf{51.1} &  89.4/90.1/90.3/\textbf{90.4} &  \textbf{92.7}/92.7/92.5/\textbf{92.7} \\
& CIFAR-100 &  33.6/41.8/45.6/\textbf{45.9} &  86.4/87.0/87.0/\textbf{87.1} &  \textbf{87.0}/86.6/86.2/86.4 \\
& Gaussian &  81.5/\textbf{89.9}/88.2/89.0 &  96.9/\textbf{97.3}/96.7/96.7 &  97.9/\textbf{98.3}/98.0/98.0 \\
& TinyImgNet &  42.1/53.4/57.7/\textbf{58.2} &  90.3/91.5/91.6/\textbf{91.7} &  91.8/\textbf{92.2}/92.1/\textbf{92.2} \\
& LSUN &  41.2/55.1/61.7/\textbf{62.0} &  90.1/91.5/92.0/\textbf{92.1} &  91.5/92.1/\textbf{92.3}/\textbf{92.3} \\
& Places365 &  32.9/42.4/\textbf{48.2}/\textbf{48.2} &  85.8/86.6/\textbf{86.9}/\textbf{86.9} &  \textbf{67.1}/66.1/65.6/65.7 \\
& SVHN &  27.7/39.9/48.6/\textbf{48.8} &  89.2/90.2/90.5/\textbf{90.6} &  \textbf{85.8}/85.1/84.1/84.5 \\
& Textures &  37.9/46.6/49.6/\textbf{50.3} &  89.0/\textbf{89.1}/88.4/88.6 &  \textbf{93.9}/93.3/92.4/92.6 \\
    \cline{2-5}
     & average &  41.7/51.9/56.3/\textbf{56.7} &  89.6/90.4/90.4/\textbf{90.5} &  \textbf{88.5}/88.3/87.9/88.1 \\
    \cline{1-5}
    \multirow{8}{*}{\multirowcell{2}{CIFAR-100\\(ResNet)}} 
& Chars &  14.3/\textbf{15.3}/15.1/15.2 &  72.7/73.0/73.1/\textbf{73.4} &  \textbf{77.8}/77.2/77.3/77.7 \\
& CIFAR-10 &  18.1/\textbf{18.4}/17.4/18.0 &  76.5/76.6/76.5/\textbf{76.8} &  \textbf{78.3}/77.8/77.8/78.1 \\
& Gaussian &      \textbf{1.6}/0.8/0.3/0.6 &  72.8/76.0/\textbf{76.7}/\textbf{76.7} &  81.6/83.8/\textbf{84.4}/84.3 \\
& TinyImgNet &  17.8/20.7/\textbf{23.8}/23.0 &  73.3/76.6/\textbf{77.4}/76.9 &  76.4/78.7/\textbf{79.2}/78.8 \\
& LSUN &  15.5/18.8/\textbf{21.5}/20.6 &  70.8/74.1/\textbf{74.9}/74.4 &  72.9/75.2/\textbf{75.7}/75.2 \\
& Places365 &  \textbf{17.3}/17.2/16.2/16.9 &  \textbf{74.1}/73.2/72.9/73.4 &  \textbf{44.6}/42.1/41.9/42.7 \\
& SVHN &  \textbf{14.2}/13.8/12.5/13.3 &  \textbf{74.8}/74.1/73.9/74.5 &  \textbf{59.4}/56.9/56.8/57.7 \\
& Textures &  20.9/22.8/\textbf{23.3}/\textbf{23.3} &  77.0/78.0/78.2/\textbf{78.3} &  85.9/86.2/86.3/\textbf{86.4} \\
    \cline{2-5}
     & average &  15.0/16.0/16.3/\textbf{16.4} &  74.0/75.2/75.4/\textbf{75.5} &  72.1/72.3/72.4/\textbf{72.6} \\
    \cline{1-5}
    \multirow{8}{*}{\multirowcell{2}{SVHN\\(ResNet)}} 
& Chars &  56.9/58.2/58.4/\textbf{58.5} &  \textbf{85.1}/83.7/83.7/84.0 &  \textbf{92.3}/91.0/91.0/91.3 \\
& CIFAR-10 &  79.0/80.6/81.0/\textbf{81.3} &  \textbf{93.0}/92.1/92.2/92.5 &  \textbf{94.7}/93.4/93.4/93.7 \\
& CIFAR-100 &  78.0/79.6/79.8/\textbf{80.2} &  \textbf{92.7}/91.9/91.9/92.2 &  \textbf{94.7}/93.4/93.4/93.7 \\
& Gaussian &  85.3/86.7/86.9/\textbf{87.3} &  \textbf{95.9}/95.7/95.7/\textbf{95.9} &  \textbf{97.7}/97.1/97.1/97.3 \\
& TinyImgNet &  79.8/81.4/81.8/\textbf{82.2} &  \textbf{93.5}/92.9/92.9/93.2 &  \textbf{95.3}/94.3/94.3/94.5 \\
& LSUN &  \textbf{74.9}/76.8/77.2/77.4 &  \textbf{91.5}/90.5/90.6/90.9 &  \textbf{93.5}/92.2/92.1/92.4 \\
& Places365 &  77.0/78.4/78.8/\textbf{79.0} &  \textbf{92.0}/91.0/91.1/91.4 &  \textbf{81.4}/77.9/77.8/78.6 \\
& Textures &  78.5/80.0/80.4/\textbf{80.9} &  \textbf{93.7}/93.0/93.0/93.4 &  \textbf{97.6}/97.0/97.0/97.2 \\
    \cline{2-5}
     & average &  76.2/77.7/78.0/\textbf{78.4} &  \textbf{92.2}/91.3/91.4/91.7 &  \textbf{93.4}/92.0/92.0/92.4 \\
\cline{1-5}
\multicolumn{2}{c}{Average of the average values}     &  46.0/51.6/52.1/\textbf{52.6}   &   85.3/85.4/85.1/\textbf{85.8}   &   \textbf{85.2}/84.7/84.4/85.1 \\
\bottomrule
\end{tabular}
}
  \end{center}
\end{table}

\begin{table}[ht]
  \begin{center}
    \caption{Extended \textsc{Grey-Box} results. Parameter tuning on iSUN dataset.}
    \label{tab:grey_box_extended}
  \resizebox{0.85\textwidth}{!}{%
\begin{tabular}{ccccc}
\toprule
    \multirowcell{2}{In-dist.\\(model)} & \multirowcell{2}{OOD\\dataset} & TNR at TPR-95\% & AUROC & AUPR \\
    \cmidrule(r){3-5}
    &  & \multicolumn{3}{c}{ODIN / Energy / \textsc{Igeood}} \\
\hline
\multirow{8}{*}{\multirowcell{2}{CIFAR-10\\(DenseNet)}} 
& Chars &     \textbf{52.5}/49.2/50.0 &   \textbf{89.1}/88.2/88.7 &   90.7/90.2/\textbf{90.8} \\
& CIFAR-100 &     48.9/49.7/\textbf{50.0} &   88.3/88.5/\textbf{88.7} &   88.4/88.6/\textbf{88.9} \\
& Gaussian &  \textbf{100}/\textbf{100}/\textbf{100} &  \textbf{100}/99.9/99.9 &  \textbf{100}/99.9/99.9 \\
& TinyImgNet &     \textbf{92.6}/92.4/92.4 &   98.5/98.5/98.5 &   \textbf{98.6}/98.5/98.5 \\
& LSUN &     96.2/96.2/96.2 &   \textbf{99.2}/99.1/\textbf{99.2} &   \textbf{99.3}/99.2/99.2 \\
& Places365 &     \textbf{52.4}/52.0/\textbf{52.4} &   89.0/88.9/\textbf{89.2} &   88.7/88.7/\textbf{89.0} \\
& SVHN &     \textbf{49.3}/41.8/42.5 &   \textbf{89.7}/88.2/88.6 &   \textbf{82.1}/80.6/81.4 \\
& Textures &     \textbf{42.9}/37.0/38.7 &   81.4/80.8/\textbf{82.3} &   84.7/84.6/\textbf{86.5} \\
\cline{2-5}
& average &     \textbf{66.8}/64.8/65.3 &   \textbf{91.9}/91.5/\textbf{91.9} &   91.6/91.3/\textbf{91.8} \\
\cline{1-5}
\multirow{8}{*}{\multirowcell{2}{CIFAR-100\\(DenseNet)}} 
            & Chars &     \textbf{19.8}/19.2/19.5 &   \textbf{76.6}/76.3/75.8 &   \textbf{80.8}/80.6/79.7 \\
            & CIFAR-10 &     15.3/16.4/\textbf{16.9} &   72.4/72.8/\textbf{74.9} &   72.6/72.8/\textbf{76.1} \\
            & Gaussian &        0.0/0.0/0.0 &   47.1/46.9/\textbf{59.9} &   65.5/65.4/\textbf{74.9} \\
            & TinyImgNet &     \textbf{43.8}/42.2/40.1 &   \textbf{86.5}/86.2/84.8 &   \textbf{87.2}/87.0/85.3 \\
            & LSUN &     \textbf{42.2}/40.6/38.9 &   \textbf{86.8}/86.4/84.7 &   \textbf{87.8}/87.7/85.7 \\
            & Places365 &     23.4/23.8/\textbf{23.9} &   79.1/79.1/\textbf{79.4} &   79.4/79.4/\textbf{80.0} \\
            & SVHN &     34.7/31.4/\textbf{35.7} &   87.7/87.2/\textbf{88.1} &   81.0/80.8/\textbf{81.8} \\
            & Textures &     \textbf{24.9}/24.7/\textbf{24.9} &   76.7/76.6/\textbf{77.8} &   81.7/81.6/\textbf{83.8} \\
\cline{2-5}
& average &     \textbf{25.5}/24.8/25.0 &   76.6/76.4/\textbf{78.2} &   79.5/79.4/\textbf{80.9} \\
\cline{1-5}
\multirow{8}{*}{\multirowcell{2}{SVHN\\(DenseNet)}} 
            & Chars &     \textbf{53.3}/45.8/48.3 &   \textbf{81.8}/78.0/79.3 &   \textbf{90.4}/88.1/88.8 \\
            & CIFAR-10 &     \textbf{69.8}/65.1/67.1 &   \textbf{91.3}/89.1/89.9 &   \textbf{95.2}/93.7/94.2 \\
            & CIFAR-100 &     \textbf{69.8}/64.2/66.3 &   \textbf{91.1}/88.5/89.4 &   \textbf{94.8}/93.0/93.6 \\
            & Gaussian &     \textbf{99.4}/99.0/99.1 &   \textbf{98.9}/98.5/98.6 &   \textbf{99.7}/99.5/99.6 \\
            & TinyImgNet &     \textbf{88.4}/85.5/86.4 &   \textbf{96.1}/94.9/95.2 &   \textbf{97.9/}97.0/97.3 \\
            & LSUN &     \textbf{88.2}/85.8/86.5 &   \textbf{96.2}/95.1/95.4 &   \textbf{98.0}/97.3/97.5 \\
            & Places365 &     \textbf{74.8}/69.9/71.8 &   \textbf{92.4}/90.3/91.0 &   \textbf{95.8}/94.2/94.7 \\
            & Textures &     \textbf{59.4}/49.6/53.7 &   \textbf{84.9}/79.0/81.4 &   \textbf{93.4}/90.4/91.6 \\
\cline{2-5}
& average &     \textbf{75.4}/70.6/72.4 &   \textbf{91.6}/89.2/90.0 &   \textbf{95.7}/94.2/94.7 \\
\cline{1-5}
\multirow{8}{*}{\multirowcell{2}{CIFAR-10\\(ResNet)}} 
            & Chars &     54.7/54.2/\textbf{54.8} &   \textbf{88.3}/87.8/88.2 &   \textbf{90.7}/90.4/90.6 \\
            & CIFAR-100 &     38.9/\textbf{41.7}/41.5 &   83.8/83.8/\textbf{84.1} &   83.7/83.5/\textbf{83.8} \\
            & Gaussian &  \textbf{100}/\textbf{100}/\textbf{100} &  \textbf{100}/99.5/99.5 &  \textbf{100}/99.7/99.7 \\
            & TinyImgNet &     \textbf{68.7}/66.6/67.0 &   \textbf{93.1}/92.4/92.6 &   \textbf{93.3}/92.7/92.9 \\
            & LSUN &     \textbf{70.4}/68.8/68.9 &   \textbf{93.2}/92.8/92.9 &   \textbf{93.3}/92.9/93.0 \\
            & Places365 &     40.4/\textbf{43.2}/43.0 &   84.1/84.1/\textbf{84.2} &   \textbf{84.0}/83.9/\textbf{84.0} \\
            & SVHN &     38.6/39.9/\textbf{40.0} &   \textbf{84.8}/84.0/84.6 &   \textbf{76.4}/74.6/75.7 \\
            & Textures &     46.8/47.1/\textbf{47.5} &   \textbf{86.1}/85.2/85.7 &   \textbf{91.4}/90.7/91.0 \\
\cline{2-5}
& average &     57.3/57.7/\textbf{57.8} &   \textbf{89.2}/88.7/89.0 &   \textbf{89.1}/88.5/88.8 \\
\cline{1-5}
\multirow{8}{*}{\multirowcell{2}{CIFAR-100\\(ResNet)}} & Chars &     14.8/13.8/13.8 &   68.0/65.1/65.0 &   71.0/68.5/68.4 \\
            & CIFAR-10 &     \textbf{14.3}/11.8/11.8 &   \textbf{70.8}/65.1/64.9 &   \textbf{70.1}/64.7/64.5 \\
            & Gaussian &     \textbf{78.7}/73.8/74.3 &   \textbf{96.8}/95.9/96.0 &   \textbf{97.7}/97.4/97.4 \\
            & TinyImgNet &     44.9/49.2/\textbf{49.3} &   86.4/\textbf{86.5}/\textbf{86.5} &   \textbf{86.1}/85.6/85.6 \\
            & LSUN &     40.2/44.3/\textbf{44.4} &   83.5/\textbf{83.8}/\textbf{83.8} &   82.5/82.5/82.5 \\
            & Places365 &     \textbf{15.9}/10.2/10.1 &   \textbf{67.7}/59.7/59.5 &   \textbf{64.5}/58.2/58.0 \\
            & SVHN &     10.2/\textbf{10.6}/\textbf{10.6} &   64.3/\textbf{65.4}/65.3 &   42.2/\textbf{43.6}/43.5 \\
            & Textures &     \textbf{29.5}/27.6/27.7 &   \textbf{77.9}/73.7/73.6 &   \textbf{84.4}/81.1/81.0 \\
\cline{2-5}
& average &     \textbf{31.1}/30.2/30.2 &   \textbf{76.9}/74.4/74.3 &   \textbf{74.8}/72.7/72.6 \\
\cline{1-5}
\multirow{8}{*}{\multirowcell{2}{SVHN\\(ResNet)}} 
            & Chars &     59.9/59.4/\textbf{60.0} &   82.8/\textbf{83.2}/\textbf{83.2} &   90.4/\textbf{90.7}/90.6 \\
            & CIFAR-10 &     80.9/81.0/\textbf{81.3} &   90.9/\textbf{91.5}/91.2 &   92.2/\textbf{92.8}/92.5 \\
            & CIFAR-100 &     80.2/80.2/\textbf{80.5} &   90.8/\textbf{91.4}/91.1 &   92.3/\textbf{92.9}/92.6 \\
            & Gaussian &     \textbf{89.4}/88.7/\textbf{89.4} &   95.9/95.9/95.9 &   \textbf{97.3}/97.2/97.2 \\
            & TinyImgNet &     82.6/82.3/\textbf{82.9} &   92.2/\textbf{92.6}/92.5 &   93.5/\textbf{93.9}/93.8 \\
            & LSUN &     77.8/77.6/\textbf{78.3} &   89.6/\textbf{90.1}/89.9 &   91.3/\textbf{91.7}/91.6 \\
            & Places365 &     78.7/78.8/\textbf{79.0} &   89.8/\textbf{90.6}/90.2 &   91.2/\textbf{91.9}/91.5 \\
            & Textures &     78.6/\textbf{79.8}/79.2 &   91.3/\textbf{92.2}/91.7 &   96.1/\textbf{96.6}/96.3 \\
\cline{2-5}
& average &     78.5/78.5/\textbf{78.8} &   90.4/\textbf{90.9}/90.7 &   93.0/\textbf{93.5}/93.3 \\
\cline{1-5}
\multicolumn{2}{c}{Average of average values} & \textbf{55.8}/54.4/54.9   &   \textbf{86.1}/85.2/85.7   &   \textbf{87.3}/86.6/87.0\\
\bottomrule
\end{tabular}
}
  \end{center}
\end{table}

\begin{table}[ht]
  \begin{center}
    \caption{\textsc{White-Box} extended results. Validation on OOD data.}
    \label{tab:wb_ood_extended}
  \resizebox{0.8\textwidth}{!}{%
\begin{tabular}{ccccc}
\toprule
    \multirowcell{2}{In-dist.\\(model)} & \multirowcell{2}{OOD\\dataset} & TNR at TPR-95\% & AUROC & AUPR \\
    \cmidrule(r){3-5}
    &  & \multicolumn{3}{c}{Mahalanobis \citep{mahalanobis} / \textsc{Igeood+}} \\
    \hline
\multirow{10}{*}{\multirowcell{2}{CIFAR-10\\(DenseNet)}} 
                                 & Chars &    91.3/\textbf{99.4} &    97.5/\textbf{99.9} &    97.7/\textbf{99.9} \\
                                 & CIFAR-100 &    21.4/\textbf{56.6} &    67.3/\textbf{90.7} &    64.4/\textbf{90.8} \\
                                 & TinyImgNet &    96.9/\textbf{99.8} &    99.3/\textbf{99.9} &    99.3/\textbf{99.9} \\
                                 & LSUN &    98.2/\textbf{99.9} &   99.5/\textbf{100} &   99.5/\textbf{100} \\
                                 & Places365 &    18.1/\textbf{80.2} &    72.7/\textbf{95.7} &    72.8/\textbf{95.4} \\
                                 & SVHN &    90.1/\textbf{99.9} &   97.3/\textbf{100} &   97.3/\textbf{100} \\
                                 & Textures &    84.1/\textbf{97.4} &    95.6/\textbf{99.5} &    94.7/\textbf{99.5} \\
                                 & Gaussian &  \textbf{100}/\textbf{100} &  \textbf{100}/\textbf{100} &  \textbf{100}/\textbf{100} \\
                                 & iSUN &    97.3/\textbf{99.8} &   99.4/\textbf{100} &   99.4/\textbf{100} \\
        \cline{2-5}                         
                                 & average &  77.5\textcolor{black}{\scriptsize{$\pm$31}}/\textbf{92.6}\textcolor{black}{\scriptsize{$\pm$14}}  &  92.1\textcolor{black}{\scriptsize{$\pm$12}}/\textbf{98.4}\textcolor{black}{\scriptsize{$\pm$3.0}}  &  91.7\textcolor{black}{\scriptsize{$\pm$13}}/\textbf{98.4}\textcolor{black}{\scriptsize{$\pm$3.0}} \\
\cline{1-5}
\multirow{10}{*}{\multirowcell{2}{CIFAR-100\\(DenseNet)}} 
                                 & Chars &    62.9/\textbf{97.5} &    94.0/\textbf{99.4} &    95.8/\textbf{99.4} \\
                                 & CIFAR-10 &     9.1/\textbf{22.7} &    60.8/\textbf{80.7} &    60.1/\textbf{83.0} \\
                                 & TinyImgNet &    87.1/\textbf{99.5} &    97.4/\textbf{99.9} &    97.4/\textbf{99.9} \\
                                 & LSUN &    91.1/\textbf{99.9} &   97.8/\textbf{100} &   98.1/\textbf{100} \\
                                 & Places365 &     5.9/\textbf{58.2} &    54.8/\textbf{90.0} &    54.7/\textbf{89.2} \\
                                 & SVHN &    79.0/\textbf{99.6} &    96.8/\textbf{99.9} &    94.1/\textbf{99.9} \\
                                 & Textures &    70.3/\textbf{90.2} &    91.4/\textbf{98.1} &    94.3/\textbf{98.2} \\
                                 & Gaussian &  \textbf{100}/\textbf{100} &  \textbf{100}/\textbf{100} &  \textbf{100}/\textbf{100} \\
                                 & iSUN &    86.4/\textbf{99.7} &    96.8/\textbf{99.9} &    97.7/\textbf{99.9} \\
                                 \cline{2-5}
                                 & average &  67.7\textcolor{black}{\scriptsize{$\pm$28}}/\textbf{90.2}\textcolor{black}{\scriptsize{$\pm$21}}  &  87.8\textcolor{black}{\scriptsize{$\pm$13}}/\textbf{97.7}\textcolor{black}{\scriptsize{$\pm$5.0}}  &  88.0\textcolor{black}{\scriptsize{$\pm$12}}/\textbf{97.8}\textcolor{black}{\scriptsize{$\pm$5.0}} \\
\cline{1-5}
\multirow{10}{*}{\multirowcell{2}{SVHN\\(DenseNet)}} 
                                 & Chars &    78.7/\textbf{92.2} &    96.1/\textbf{98.4} &    \textbf{98.9}/98.5 \\
                                 & CIFAR-10 &    91.6/\textbf{98.3} &    98.0/\textbf{99.6} &    99.4/\textbf{99.6} \\
                                 & CIFAR-100 &    92.9/\textbf{95.3} &    98.2/\textbf{99.1} &    \textbf{99.4}/99.2 \\
                                 & TinyImgNet &    \textbf{99.9}/\textbf{99.9} &    99.8/\textbf{99.9} &    \textbf{99.9}/\textbf{99.9} \\
                                 & LSUN &   \textbf{99.9}/\textbf{99.9} &   99.8/\textbf{100} &   99.7/\textbf{100} \\
                                 & Places365 &    94.7/\textbf{98.3} &    98.3/\textbf{99.6} &    98.4/\textbf{99.7} \\
                                 & Textures &    98.2/\textbf{98.5} &    99.4/\textbf{99.6} &    \textbf{99.9}/99.6 \\
                                 & Gaussian &  \textbf{100}/\textbf{100} &  \textbf{100}/\textbf{100} &  \textbf{100}/\textbf{100} \\
                                 & iSUN &    \textbf{99.9}/\textbf{99.9} &    99.8/\textbf{99.9} &    \textbf{99.9}/\textbf{99.9} \\
                                 \cline{2-5}
                                 & average&  95.1\textcolor{black}{\scriptsize{$\pm$8.0}}/\textbf{98.0}\textcolor{black}{\scriptsize{$\pm$2.0}}  &  98.8\textcolor{black}{\scriptsize{$\pm$1.0}}/\textbf{99.6}\textcolor{black}{\scriptsize{$\pm$0.1}}  &  99.5\textcolor{black}{\scriptsize{$\pm$1.0}}/\textbf{99.6}\textcolor{black}{\scriptsize{$\pm$0.1}} \\
\cline{1-5}
\multirow{10}{*}{\multirowcell{2}{CIFAR-10\\(ResNet)}} 
                                 & Chars &    93.6/\textbf{99.3} &    98.6/\textbf{99.8} &    99.1/\textbf{99.8} \\
                                 & CIFAR-100 &    44.9/\textbf{51.3} &    87.4/\textbf{90.9} &    87.8/\textbf{91.7} \\
                                 & TinyImgNet &    96.8/\textbf{99.6} &    99.4/\textbf{99.9} &    99.4/\textbf{99.9} \\
                                 & LSUN &    98.3/\textbf{99.9} &   99.6/\textbf{100} &   99.6/\textbf{100} \\
                                 & Places365 &    45.8/\textbf{77.6} &    88.1/\textbf{95.6} &    88.1/\textbf{95.5} \\
                                 & SVHN &    96.1/\textbf{99.8} &    99.0/\textbf{99.9} &    98.1/\textbf{99.9} \\
                                 & Textures &    84.3/\textbf{97.0} &    97.3/\textbf{99.4} &    98.6/\textbf{99.4} \\
                                 & Gaussian &  \textbf{100}/\textbf{100} &  \textbf{100}/\textbf{100} &  \textbf{100}/\textbf{100} \\
                                 & iSUN &    97.2/\textbf{99.9} &   99.4/\textbf{100} &   99.5/\textbf{100} \\
                                 \cline{2-5}
                                 & average &  84.1\textcolor{black}{\scriptsize{$\pm$23}}/\textbf{91.6}\textcolor{black}{\scriptsize{$\pm$16}}  &  96.5\textcolor{black}{\scriptsize{$\pm$4.0}}/\textbf{98.4}\textcolor{black}{\scriptsize{$\pm$3.0}}  &  96.7\textcolor{black}{\scriptsize{$\pm$4.0}}/\textbf{98.5}\textcolor{black}{\scriptsize{$\pm$3.0}} \\
\cline{1-5}
\multirow{10}{*}{\multirowcell{2}{CIFAR-100\\(ResNet)}} 
                                 & Chars &    63.8/\textbf{97.8} &    94.0/\textbf{99.5} &    96.0/\textbf{99.5} \\
                                 & CIFAR-10 &    18.0/\textbf{30.8} &    76.6/\textbf{85.3} &    76.4/\textbf{87.8} \\
                                 & TinyImgNet &    90.1/\textbf{99.6} &    97.9/\textbf{99.9} &    98.0/\textbf{99.9} \\
                                 & LSUN &   92.4/\textbf{100} &   98.3/\textbf{100} &   98.5/\textbf{100} \\
                                 & Places365 &    23.5/\textbf{59.1} &    76.8/\textbf{91.2} &    76.0/\textbf{91.4} \\
                                 & SVHN &    88.4/\textbf{99.7} &    97.7/\textbf{99.9} &    95.2/\textbf{99.9} \\
                                 & Textures &    71.6/\textbf{90.7} &    93.9/\textbf{98.2} &    96.6/\textbf{98.1} \\
                                 & Gaussian &  \textbf{100}/\textbf{100} &  \textbf{100}/\textbf{100} &  \textbf{100}/\textbf{100} \\
                                 & iSUN &    89.4/\textbf{99.8} &    97.7/\textbf{99.9} &    98.0/\textbf{99.9} \\
                                 \cline{2-5}
                                 & average &  70.8\textcolor{black}{\scriptsize{$\pm$30}}/\textbf{86.4}\textcolor{black}{\scriptsize{$\pm$23}}  &  92.5\textcolor{black}{\scriptsize{$\pm$10}}/\textbf{97.1}\textcolor{black}{\scriptsize{$\pm$5.0}}  &  92.7\textcolor{black}{\scriptsize{$\pm$10}}/\textbf{97.4}\textcolor{black}{\scriptsize{$\pm$4.0}} \\
\cline{1-5}
\multirow{10}{*}{\multirowcell{2}{SVHN\\(ResNet)}} 
                                 & Chars &    84.9/\textbf{92.4} &    97.0/\textbf{98.4} &    \textbf{99.0}/98.5 \\
                                 & CIFAR-10 &    98.0/\textbf{99.7} &    99.2/\textbf{99.9} &    99.7/\textbf{99.9} \\
                                 & CIFAR-100 &    98.3/\textbf{99.1} &    99.3/\textbf{99.7} &    \textbf{99.8}/\textbf{99.8} \\
                                 & TinyImgNet &    \textbf{99.9}/\textbf{99.9} &   99.9/\textbf{100} &   \textbf{100}/\textbf{100} \\
                                 & LSUN &    \textbf{99.9}/\textbf{99.9} &   99.9/\textbf{100} &   \textbf{100}/\textbf{100} \\
                                 & Places365 &    98.4/\textbf{99.6} &    99.3/\textbf{99.9} &    99.8/\textbf{99.9} \\
                                 & Textures &    99.0/\textbf{99.9} &    99.7/\textbf{99.9} &    \textbf{99.9}/\textbf{99.9} \\
                                 & Gaussian &  \textbf{100}/\textbf{100} &  \textbf{100}/\textbf{100} &  \textbf{100}/\textbf{100} \\
                                 & iSUN &  \textbf{100}/\textbf{100} &   99.9/\textbf{100} &   \textbf{100}/\textbf{100} \\
                                 \cline{2-5}
                                 & average &  97.6\textcolor{black}{\scriptsize{$\pm$6.0}}/\textbf{98.9}\textcolor{black}{\scriptsize{$\pm$2.0}}  &  99.4\textcolor{black}{\scriptsize{$\pm$1.0}}/\textbf{99.7}\textcolor{black}{\scriptsize{$\pm$0.1}}  &  \textbf{99.8}\textcolor{black}{\scriptsize{$\pm$1.0}}/\textbf{99.8}\textcolor{black}{\scriptsize{$\pm$0.1}} \\
\bottomrule
\multicolumn{2}{c}{Avg. and std. of avg. values} & 82.1\textcolor{black}{\scriptsize{$\pm$11}}/\textbf{92.9}\textcolor{black}{\scriptsize{$\pm$4.0}} & 94.5\textcolor{black}{\scriptsize{$\pm$4.0}}/\textbf{98.5}\textcolor{black}{\scriptsize{$\pm$1.0}} & 94.7\textcolor{black}{\scriptsize{$\pm$4.0}}/\textbf{98.6}\textcolor{black}{\scriptsize{$\pm$1.0}} \\
    \bottomrule
\end{tabular}
}
  \end{center}
\end{table}

\begin{table}[ht]
\begin{center}
    \caption{\textsc{White-Box} extended results. Validation on adversarial (FGSM) data.}
    \label{tab:wb_adv_extended}
  \resizebox{0.8\textwidth}{!}{%
\begin{tabular}{ccccc}
\toprule
    \multirowcell{2}{In-dist.\\(model)} & \multirowcell{2}{OOD\\dataset} & TNR at TPR-95\% & AUROC & AUPR \\
    \cmidrule(r){3-5}
    &  & \multicolumn{3}{c}{Mahalanobis \citep{mahalanobis} / \textsc{Igeood}} \\
    \hline
\multirow{9}{*}{\multirowcell{2}{CIFAR-10\\(DenseNet)}}  
                                 & Chars &    \textbf{88.5}/87.3 &    \textbf{97.7}/97.7 &    \textbf{98.3}/98.3 \\
& CIFAR-100 &    21.5/\textbf{26.4} &    68.0/\textbf{77.7} &    66.3/\textbf{75.5} \\
                                 & TinyImageNet &    \textbf{93.9}/93.4 &    98.6/\textbf{98.7} &    98.6/\textbf{98.7} \\
                                 & LSUN &    96.3/\textbf{96.4} &    99.1/\textbf{99.2} &    99.1/\textbf{99.2} \\
                                 & Places365 &    17.8/\textbf{23.2} &    70.0/\textbf{77.9} &    40.1/\textbf{76.3} \\
                                 & SVHN &    87.0/\textbf{94.3} &    97.2/\textbf{98.7} &    93.7/\textbf{97.3} \\
                                 & Textures &    83.6/\textbf{86.0} &    95.8/\textbf{97.2} &    97.3/\textbf{98.2} \\
                                 & Gaussian &  \textbf{100}/\textbf{100} &  \textbf{100}/\textbf{100} &  \textbf{100}/\textbf{100} \\
                                 & iSUN &    94.3/\textbf{94.5} &    98.8/\textbf{98.9} &    98.9/\textbf{99.0} \\
\cline{2-5}
 & average &  75.9\textcolor{black}{\scriptsize{$\pm$30}}/\textbf{77.9}\textcolor{black}{\scriptsize{$\pm$29}}  &  91.7\textcolor{black}{\scriptsize{$\pm$12}}/\textbf{94.0}\textcolor{black}{\scriptsize{$\pm$9.0}}  &  88.0\textcolor{black}{\scriptsize{$\pm$20}}/\textbf{93.6}\textcolor{black}{\scriptsize{$\pm$10}} \\
\cline{1-5}
\multirow{9}{*}{\multirowcell{2}{CIFAR-100\\(DenseNet)}}  & CIFAR-10 &      1.1/\textbf{5.7} &    43.5/\textbf{62.6} &    46.7/\textbf{62.6} \\
                                 & Chars &    53.9/\textbf{59.6} &    \textbf{92.2}/92.0 &    \textbf{94.5}/93.9 \\
                                 & TinyImageNet &    86.4/\textbf{94.3} &    97.4/\textbf{98.8} &    97.5/\textbf{98.9} \\
                                 & LSUN &    88.6/\textbf{95.1} &    97.6/\textbf{98.9} &    97.9/\textbf{98.9} \\
                                 & Places365 &     5.5/\textbf{13.0} &    56.6/\textbf{71.0} &    57.5/\textbf{71.0} \\
                                 & SVHN &    56.1/\textbf{90.1} &    91.8/\textbf{98.0} &    85.4/\textbf{96.2} \\
                                 & Textures &    67.5/\textbf{86.7} &    91.2/\textbf{97.4} &    94.4/\textbf{98.4} \\
                                 & Gaussian &  \textbf{100}/\textbf{100} &  \textbf{100}/\textbf{100} &  \textbf{100}/\textbf{100} \\
                                 & iSUN &    84.8/\textbf{93.8} &    97.2/\textbf{98.7} &    97.6/\textbf{98.8} \\
            \cline{2-5}
 & average &  60.4\textcolor{black}{\scriptsize{$\pm$34}}/\textbf{70.9}\textcolor{black}{\scriptsize{$\pm$35}}  &  85.3\textcolor{black}{\scriptsize{$\pm$19}}/\textbf{90.8}\textcolor{black}{\scriptsize{$\pm$13}}  &  85.7\textcolor{black}{\scriptsize{$\pm$19}}/\textbf{91.0}\textcolor{black}{\scriptsize{$\pm$13}} \\
\cline{1-5}
\multirow{9}{*}{\multirowcell{2}{SVHN\\(DenseNet)}} & CIFAR-10 &    \textbf{90.6}/89.5 &    97.7/\textbf{97.8} &    99.1/\textbf{99.2} \\
                                 & CIFAR-100 &    \textbf{91.8}/88.4 &    \textbf{98.0}/97.7 &    \textbf{99.2}/99.1 \\
                                 & Chars &    \textbf{72.3}/70.5 &    \textbf{95.2}/94.5 &    \textbf{98.5}/98.3 \\
                                 & TinyImageNet &    \textbf{99.5}/98.1 &    \textbf{99.6}/99.3 &    99.5/\textbf{99.8} \\
                                 & LSUN &    \textbf{99.9}/97.3 &    \textbf{99.8}/99.1 &    \textbf{99.9}/99.7 \\
                                 & Places365 &    \textbf{94.3}/91.9 &    \textbf{98.3}/98.2 &    98.1/\textbf{99.3} \\
                                 & Textures &    95.3/\textbf{97.1} &    98.8/\textbf{99.3} &    99.6/\textbf{99.8} \\
                                 & Gaussian &  \textbf{100}/\textbf{100} &   \textbf{100}/99.9 &  \textbf{100}/\textbf{100} \\
                                 & iSUN &    \textbf{99.9}/98.2 &    \textbf{99.8}/99.3 &    \textbf{99.9}/99.8 \\
            \cline{2-5}
 & average &  \textbf{93.7}\textcolor{black}{\scriptsize{$\pm$8.0}}/92.3\textcolor{black}{\scriptsize{$\pm$9.0}}  &  \textbf{98.6}\textcolor{black}{\scriptsize{$\pm$1.0}}/98.3\textcolor{black}{\scriptsize{$\pm$2.0}}  &  \textbf{99.3}\textcolor{black}{\scriptsize{$\pm$1.0}}/99.4\textcolor{black}{\scriptsize{$\pm$0.5}} \\
\cline{1-5}
\multirow{9}{*}{\multirowcell{2}{CIFAR-10\\(ResNet)}} & CIFAR-100 &    \textbf{36.5}/21.5 &    \textbf{84.5}/63.3 &    \textbf{84.3}/58.1 \\
                                 & Chars &    82.0/\textbf{90.9} &    96.9/\textbf{98.3} &    97.7/\textbf{98.7} \\
                                 & TinyImageNet &    \textbf{96.2}/94.3 &    \textbf{99.2}/98.0 &    \textbf{99.2}/96.7 \\
                                 & LSUN &    \textbf{98.2}/97.7 &    \textbf{99.5}/99.2 &    \textbf{99.5}/98.9 \\
                                 & Places365 &    \textbf{34.8}/15.9 &    \textbf{85.0}/60.1 &    \textbf{84.2}/24.4 \\
                                 & SVHN &    81.0/\textbf{98.2} &    96.6/\textbf{99.3} &    93.7/\textbf{97.5} \\
                                 & Textures &    \textbf{81.7}/81.6 &    \textbf{96.7}/93.4 &    \textbf{98.2}/94.3 \\
                                 & Gaussian &  \textbf{100}/\textbf{100} &  \textbf{100}/\textbf{100} &  \textbf{100}/\textbf{100} \\
                                 & iSUN &    \textbf{96.8}/95.3 &    \textbf{99.3}/98.6 &    \textbf{99.3}/98.1 \\
            \cline{2-5}
 & average &  \textbf{78.6}\textcolor{black}{\scriptsize{$\pm$24}}/77.3\textcolor{black}{\scriptsize{$\pm$32}}  &  \textbf{95.3}\textcolor{black}{\scriptsize{$\pm$6.0}}/90.0\textcolor{black}{\scriptsize{$\pm$15}}  &  \textbf{95.1}\textcolor{black}{\scriptsize{$\pm$6.0}}/85.2\textcolor{black}{\scriptsize{$\pm$25}} \\
\cline{1-5}
\multirow{9}{*}{\multirowcell{2}{CIFAR-100\\(ResNet)}}& CIFAR-10 &      3.0/\textbf{5.0} &    \textbf{61.0}/59.6 &    \textbf{63.7}/60.6 \\
                                 & Chars &    39.9/\textbf{55.1} &    85.6/\textbf{90.4} &    88.1/\textbf{92.5} \\
                                 & TinyImageNet &    \textbf{88.7}/86.2 &    \textbf{97.6}/97.3 &    \textbf{97.6}/97.3 \\
                                 & LSUN &    \textbf{91.3}/88.6 &    \textbf{98.0}/97.8 &    \textbf{98.3}/98.0 \\
                                 & Places365 &      8.0/\textbf{8.6} &    \textbf{67.9}/63.0 &    \textbf{66.8}/61.7 \\
                                 & SVHN &    31.6/\textbf{75.2} &    82.9/\textbf{95.8} &    68.8/\textbf{92.7} \\
                                 & Textures &    65.9/\textbf{78.1} &    91.9/\textbf{95.6} &    95.2/\textbf{97.6} \\
                                 & Gaussian &  \textbf{100}/\textbf{100} &  \textbf{100}/\textbf{100} &  \textbf{100}/\textbf{100} \\
                                 & iSUN &    87.9/\textbf{89.4} &    97.4/\textbf{97.8} &    97.6/\textbf{97.7} \\
            \cline{2-5}
 & average &  57.4\textcolor{black}{\scriptsize{$\pm$36}}/\textbf{65.1}\textcolor{black}{\scriptsize{$\pm$33}}  &  \textbf{86.9}\textcolor{black}{\scriptsize{$\pm$13}}/88.6\textcolor{black}{\scriptsize{$\pm$15}}  &  86.2\textcolor{black}{\scriptsize{$\pm$14}}/\textbf{88.7}\textcolor{black}{\scriptsize{$\pm$15}} \\
\cline{1-5}
\multirow{9}{*}{\multirowcell{2}{SVHN\\(ResNet)}}  & CIFAR-10 &    \textbf{97.1}/96.7 &    99.1/\textbf{99.2} &    \textbf{99.7}/99.7 \\
                                 & CIFAR-100 &    \textbf{97.5}/96.2 &    \textbf{99.1}/99.1 &    \textbf{99.7}/99.6 \\
                                 & Chars &    \textbf{75.4}/55.1 &    \textbf{95.3}/89.1 &    \textbf{98.5}/96.0 \\
                                 & TinyImageNet &    \textbf{99.9}/99.6 &    \textbf{99.9}/\textbf{99.9} &    \textbf{99.9}/\textbf{99.9} \\
                                 & LSUN &   \textbf{100}/99.8 &    \textbf{99.9}/\textbf{99.9} &  \textbf{100}/\textbf{100} \\
                                 & Places365 &    \textbf{98.1}/97.0 &    \textbf{99.2}/99.2 &    \textbf{99.2}/99.0 \\
                                 & Textures &    \textbf{98.9}/98.4 &    \textbf{99.6}/99.6 &    \textbf{99.9}/\textbf{99.9} \\
                                 & Gaussian &  \textbf{100}/\textbf{100} &   99.9/\textbf{100} &  \textbf{100}/\textbf{100} \\
                                 & iSUN &   \textbf{100}/99.8 &    99.8/\textbf{99.9} &   99.9/\textbf{100} \\
\cline{2-5}
 & average &  \textbf{96.3}\textcolor{black}{\scriptsize{$\pm$8.0}}/93.6\textcolor{black}{\scriptsize{$\pm$14}}  &  \textbf{99.1}\textcolor{black}{\scriptsize{$\pm$1.0}}/98.4\textcolor{black}{\scriptsize{$\pm$3.0}}  &  \textbf{99.6}\textcolor{black}{\scriptsize{$\pm$0.5}}/99.3\textcolor{black}{\scriptsize{$\pm$1.0}} \\
\cline{1-5}
\multicolumn{2}{c}{Avg. and std. of avg. values} &    77.0\textcolor{black}{\scriptsize{$\pm$15}}/\textbf{79.5}\textcolor{black}{\scriptsize{$\pm$10}} & 92.8\textcolor{black}{\scriptsize{$\pm$5.4}}/\textbf{93.4}\textcolor{black}{\scriptsize{$\pm$3.9}} & 92.3\textcolor{black}{\scriptsize{$\pm$5.9}}/\textbf{92.9}\textcolor{black}{\scriptsize{$\pm$5.2}} \\
    \bottomrule
\end{tabular}
}
\end{center}
\end{table}

\section{\textcolor{black}{Histograms}}\label{sec:histograms}

\begin{figure}[ht]
    \centering
         \begin{subfigure}[b]{1.0\textwidth}
         \centering
         \includegraphics[width=\textwidth]{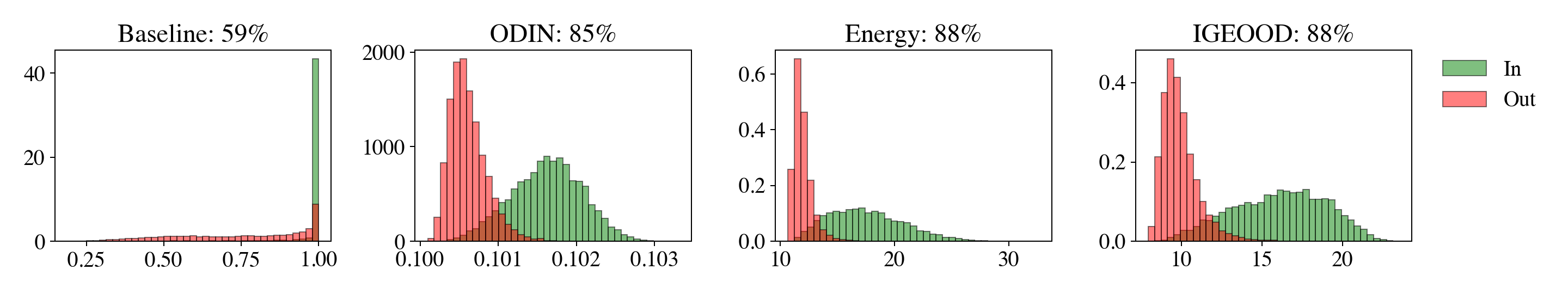}
         \caption{DenseNet on CIFAR-10.}
         \label{fig:h1d1}
     \end{subfigure}
     \begin{subfigure}[b]{1.0\textwidth}
         \centering
         \includegraphics[width=\textwidth]{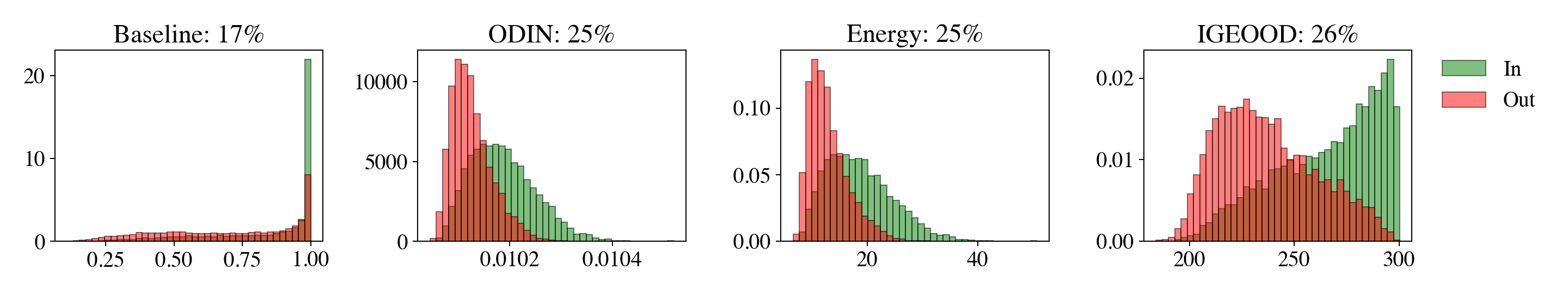}
         \caption{DenseNet on CIFAR-100.}
         \label{fig:h1d2}
     \end{subfigure}
     \begin{subfigure}[b]{1.0\textwidth}
         \centering
         \includegraphics[width=\textwidth]{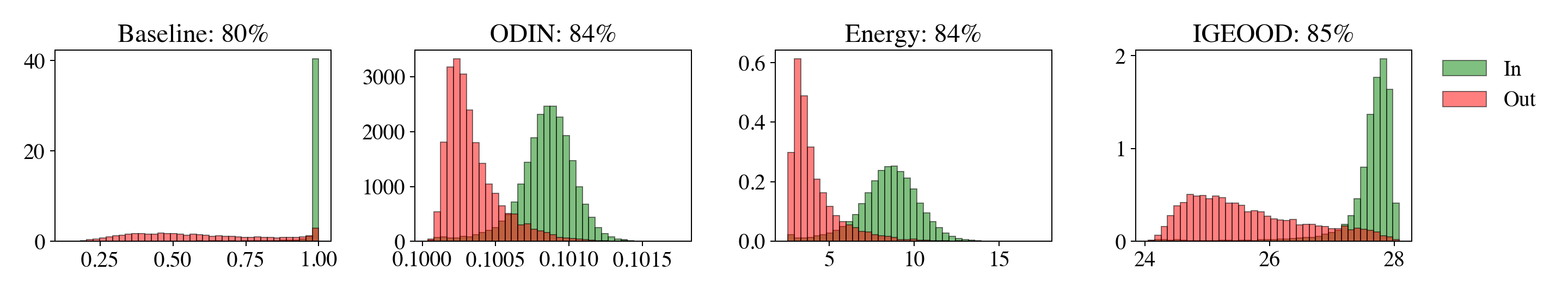}
         \caption{DenseNet on SVHN.}
         \label{fig:h1d3}
     \end{subfigure}
     \begin{subfigure}[b]{1.0\textwidth}
         \centering
         \includegraphics[width=\textwidth]{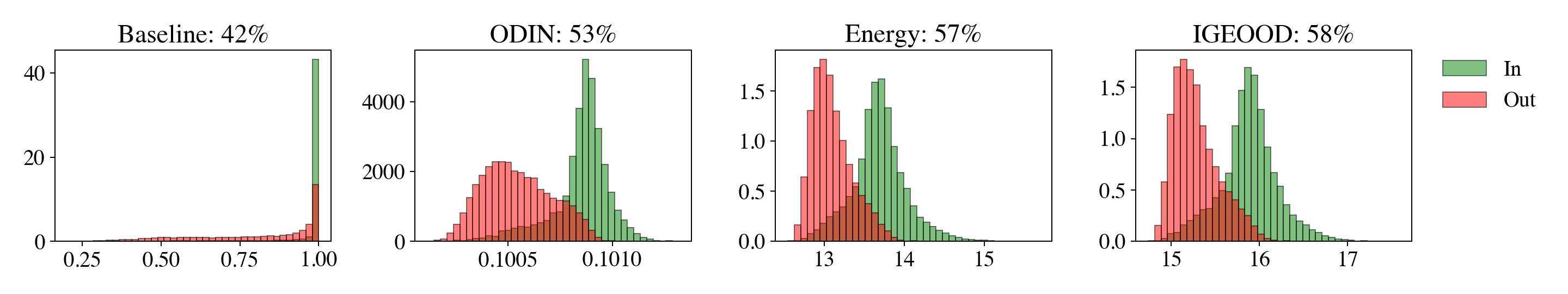}
         \caption{ResNet on CIFAR-10.}
         \label{fig:h1r1}
     \end{subfigure}
     \begin{subfigure}[b]{1.0\textwidth}
         \centering
         \includegraphics[width=\textwidth]{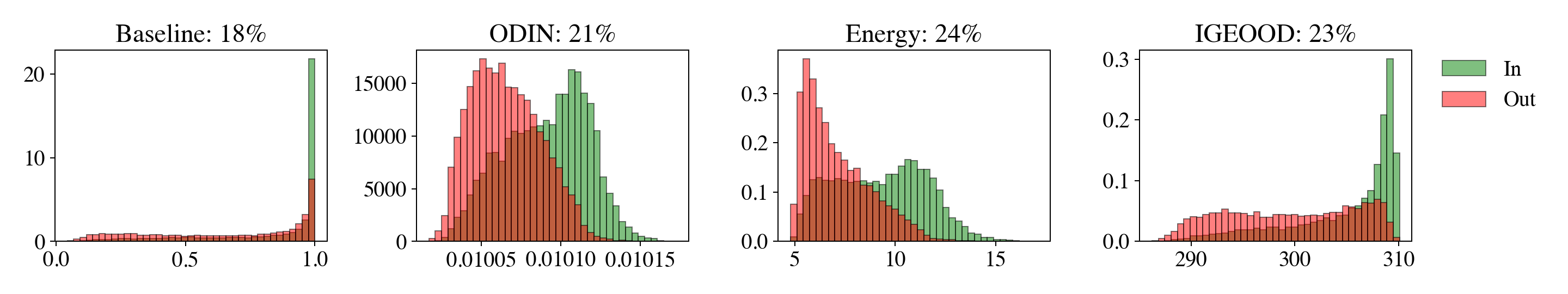}
         \caption{ResNet on CIFAR-100.}
         \label{fig:h1r2}
     \end{subfigure}
     \begin{subfigure}[b]{1.0\textwidth}
         \centering
         \includegraphics[width=\textwidth]{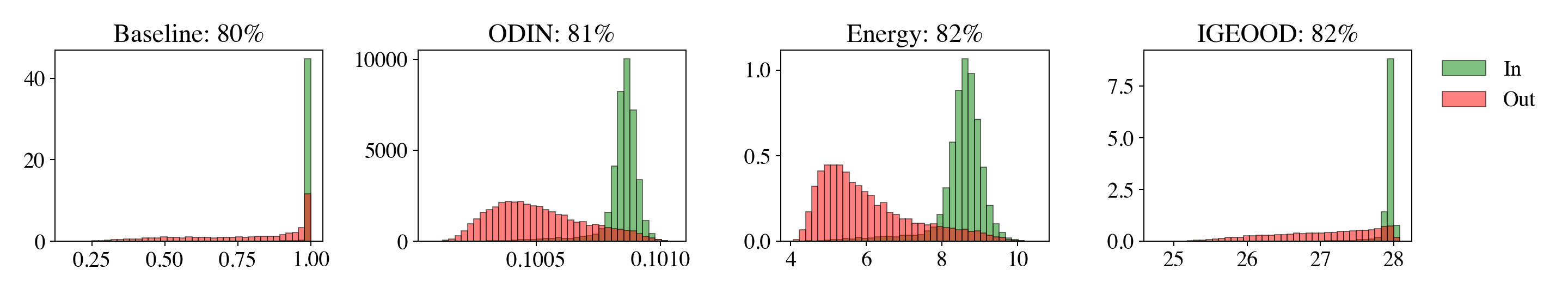}
         \caption{ResNet on SVHN.}
         \label{fig:h1r3}
     \end{subfigure}
         \caption{\textcolor{black}{\textsc{Black-Box} setup. TinyImageNet as OOD dataset.}}
    \label{fig:bb_histograms}
\end{figure}

\begin{figure}[ht]
    \centering
         \begin{subfigure}[b]{0.85\textwidth}
         \centering
         \includegraphics[width=\textwidth]{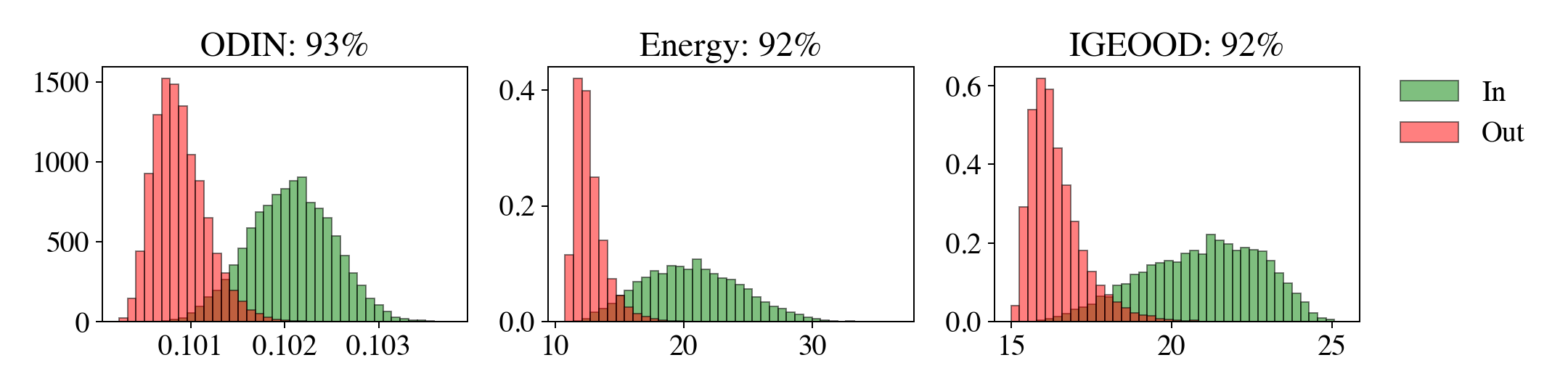}
         \caption{DenseNet on CIFAR-10.}
         \label{fig:h2d1}
     \end{subfigure}
     \begin{subfigure}[b]{0.85\textwidth}
         \centering
         \includegraphics[width=\textwidth]{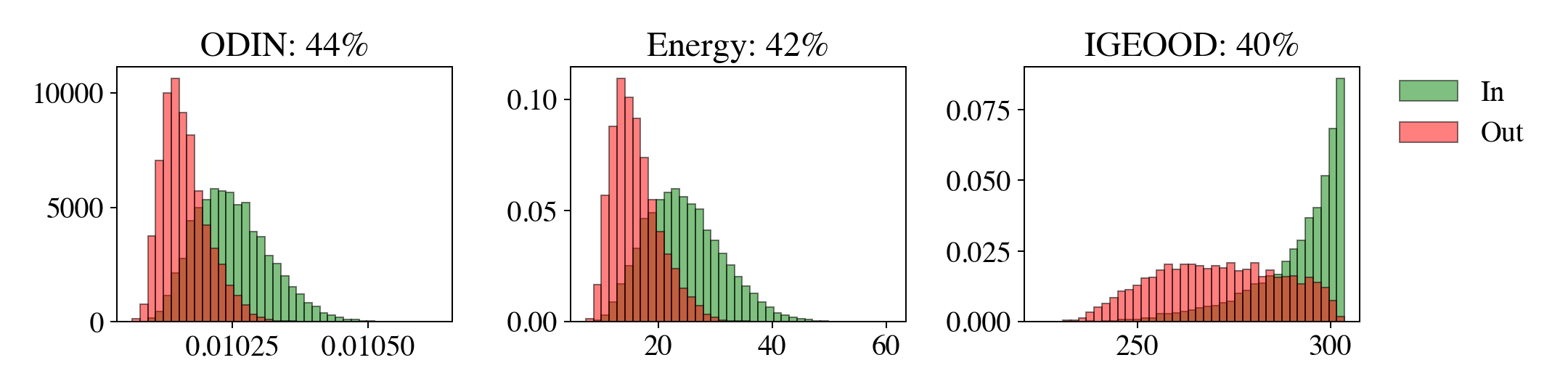}
         \caption{DenseNet on CIFAR-100.}
         \label{fig:h2d2}
     \end{subfigure}
     \begin{subfigure}[b]{0.85\textwidth}
         \centering
         \includegraphics[width=\textwidth]{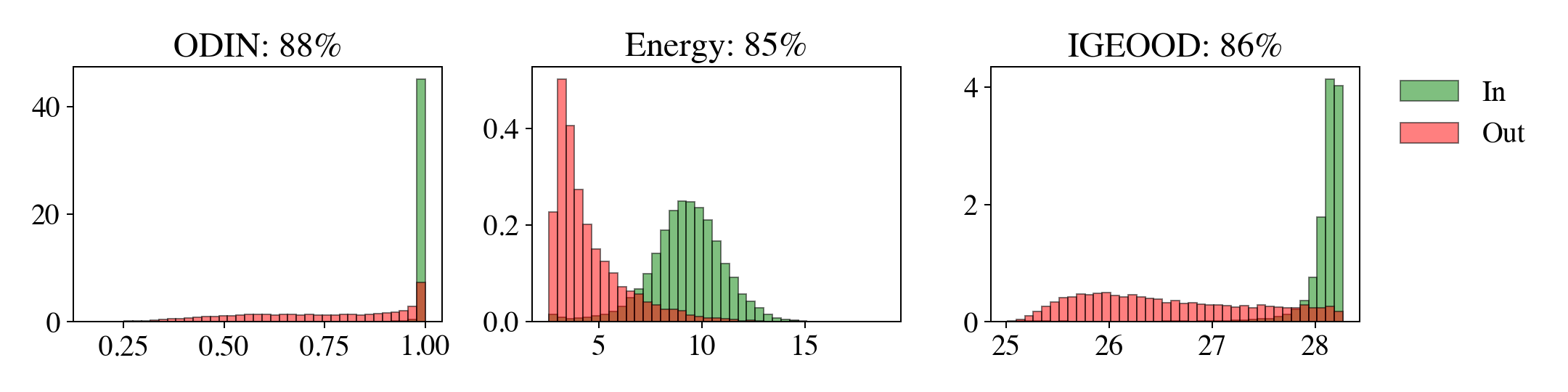}
         \caption{DenseNet on SVHN.}
         \label{fig:h2d3}
     \end{subfigure}
     \begin{subfigure}[b]{0.85\textwidth}
         \centering
         \includegraphics[width=\textwidth]{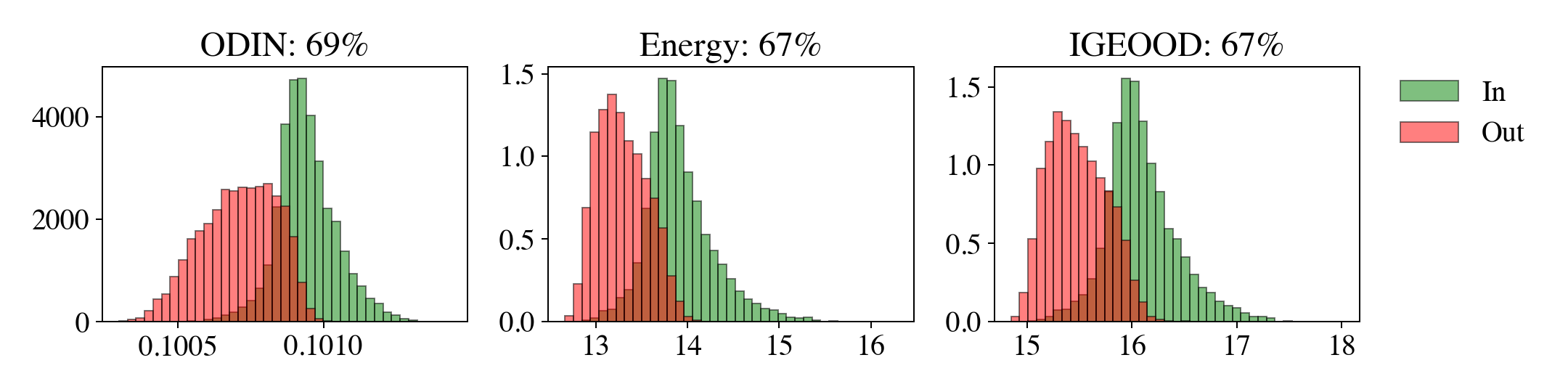}
         \caption{ResNet on CIFAR-10.}
         \label{fig:h2r1}
     \end{subfigure}
     \begin{subfigure}[b]{0.85\textwidth}
         \centering
         \includegraphics[width=\textwidth]{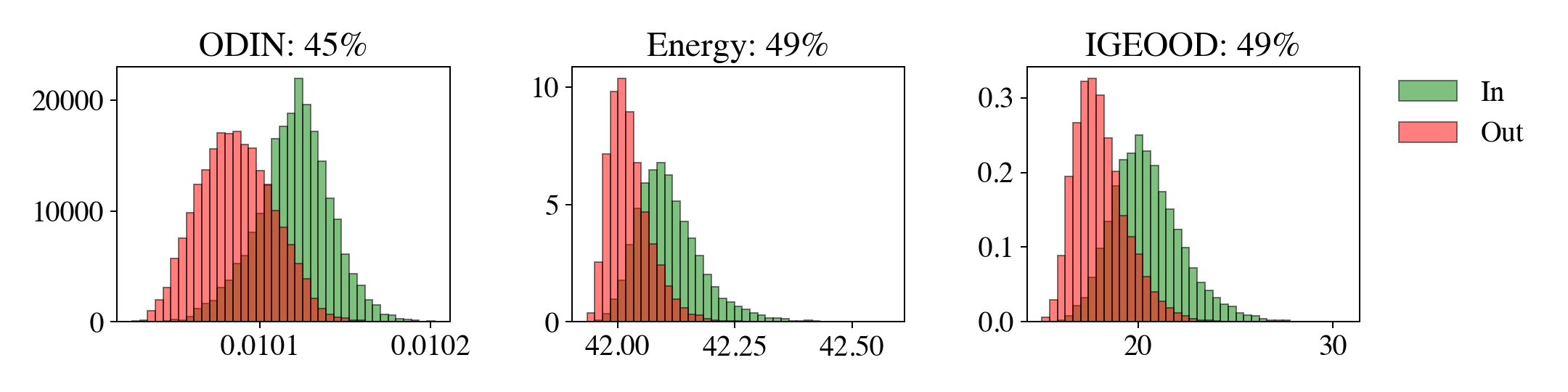}
         \caption{ResNet on CIFAR-100.}
         \label{fig:h2r2}
     \end{subfigure}
     \begin{subfigure}[b]{0.85\textwidth}
         \centering
         \includegraphics[width=\textwidth]{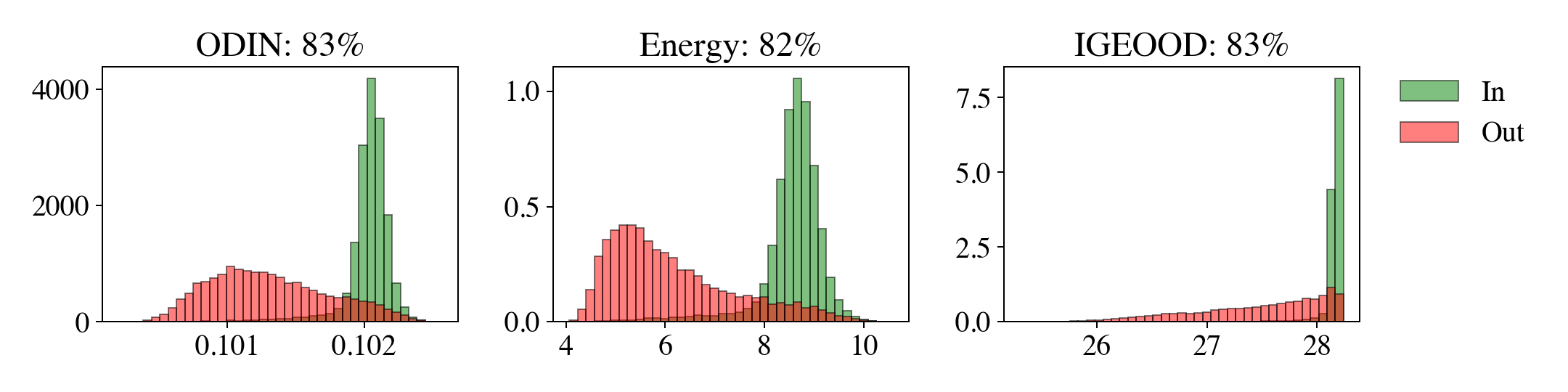}
         \caption{ResNet on SVHN.}
         \label{fig:h2r3}
     \end{subfigure}
         \caption{\textcolor{black}{\textsc{Grey-Box} setup. TinyImageNet as OOD dataset.}}
    \label{fig:gb_histograms}
\end{figure}

\begin{figure}[ht]
    \centering
         \begin{subfigure}[b]{0.85\textwidth}
         \centering
         \includegraphics[width=\textwidth]{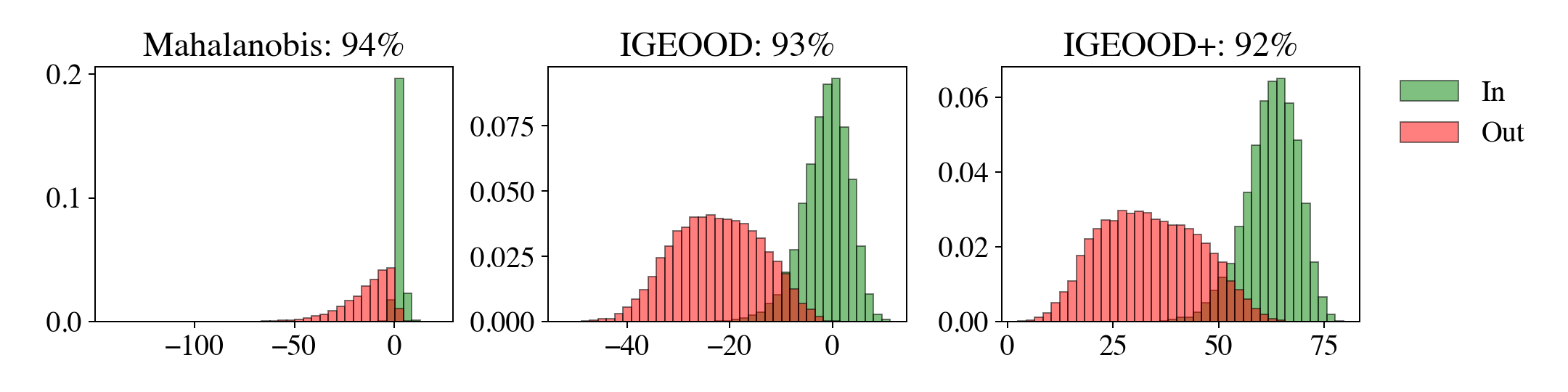}
         \caption{DenseNet on CIFAR-10.}
         \label{fig:h3d1}
     \end{subfigure}
     \begin{subfigure}[b]{0.85\textwidth}
         \centering
         \includegraphics[width=\textwidth]{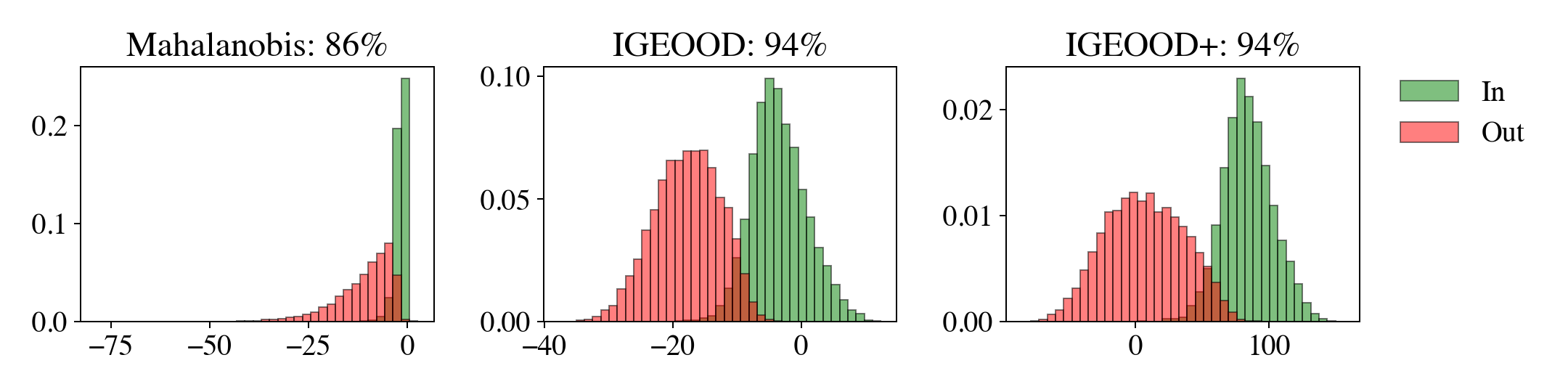}
         \caption{DenseNet on CIFAR-100.}
         \label{fig:h3d2}
     \end{subfigure}
     \begin{subfigure}[b]{0.85\textwidth}
         \centering
         \includegraphics[width=\textwidth]{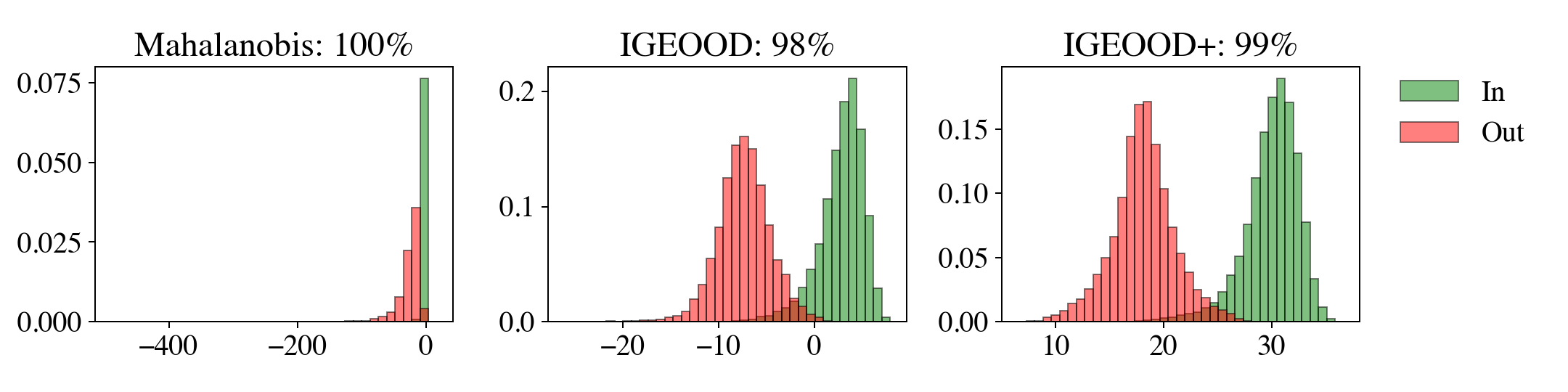}
         \caption{DenseNet on SVHN.}
         \label{fig:h3d3}
     \end{subfigure}
     \begin{subfigure}[b]{0.85\textwidth}
         \centering
         \includegraphics[width=\textwidth]{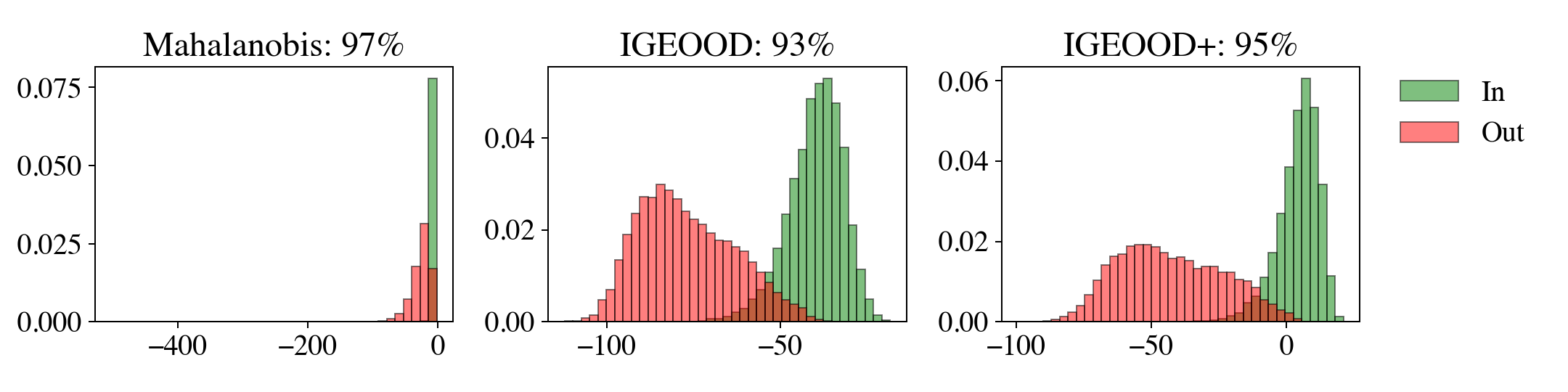}
         \caption{ResNet on CIFAR-10.}
         \label{fig:h3r1}
     \end{subfigure}
     \begin{subfigure}[b]{0.85\textwidth}
         \centering
         \includegraphics[width=\textwidth]{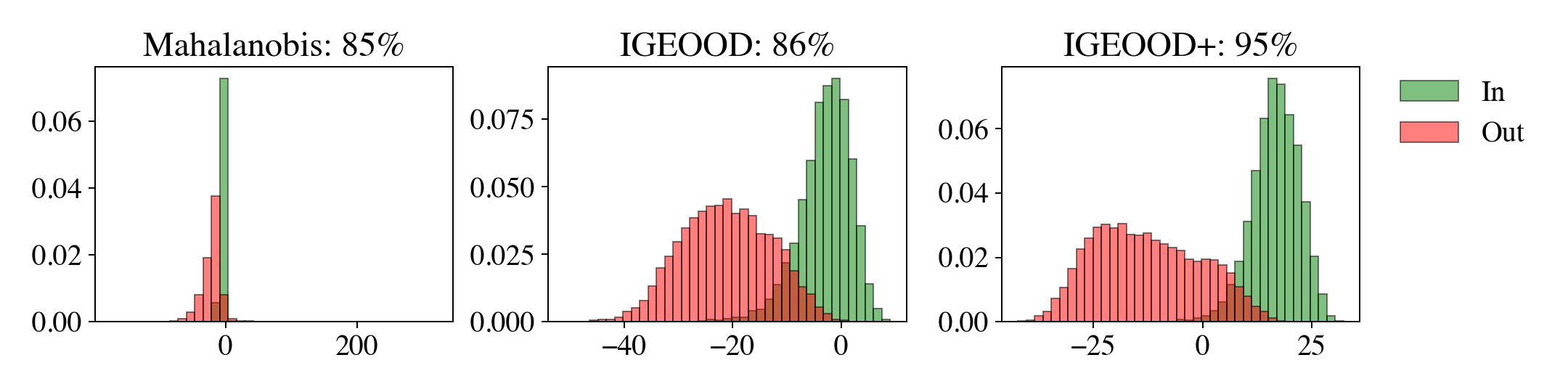}
         \caption{ResNet on CIFAR-100.}
         \label{fig:h3r2}
     \end{subfigure}
     \begin{subfigure}[b]{0.85\textwidth}
         \centering
         \includegraphics[width=\textwidth]{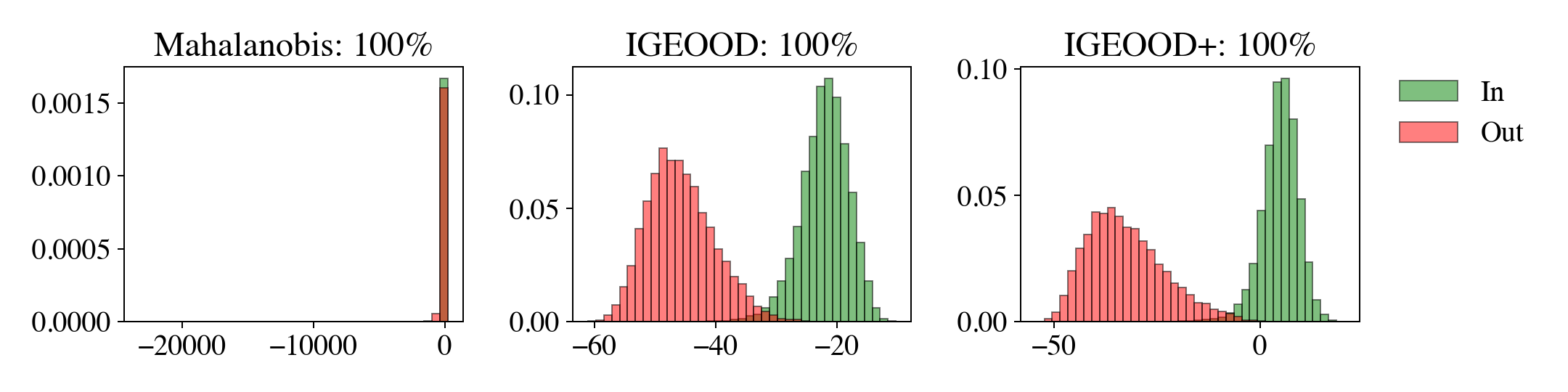}
         \caption{ResNet on SVHN.}
         \label{fig:h3r3}
     \end{subfigure}
         \caption{\textcolor{black}{\textsc{White-Box} setup with adversarial data validation. TinyImageNet as OOD dataset.}}
    \label{fig:wb_adv_histograms}
\end{figure}

\begin{figure}[ht]
    \centering
         \begin{subfigure}[b]{0.85\textwidth}
         \centering
         \includegraphics[width=\textwidth]{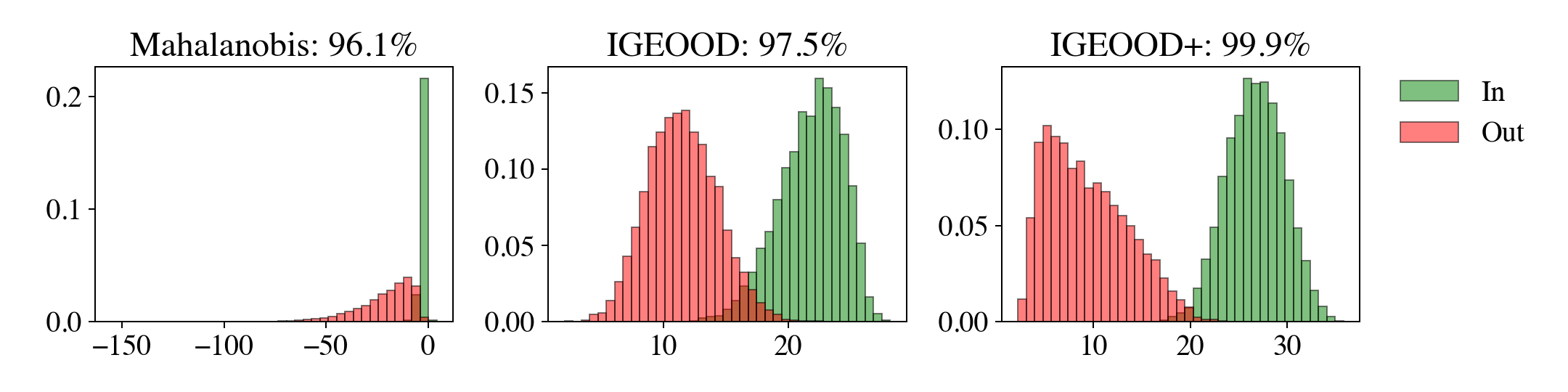}
         \caption{DenseNet on CIFAR-10.}
         \label{fig:h4d1}
     \end{subfigure}
     \begin{subfigure}[b]{0.85\textwidth}
         \centering
         \includegraphics[width=\textwidth]{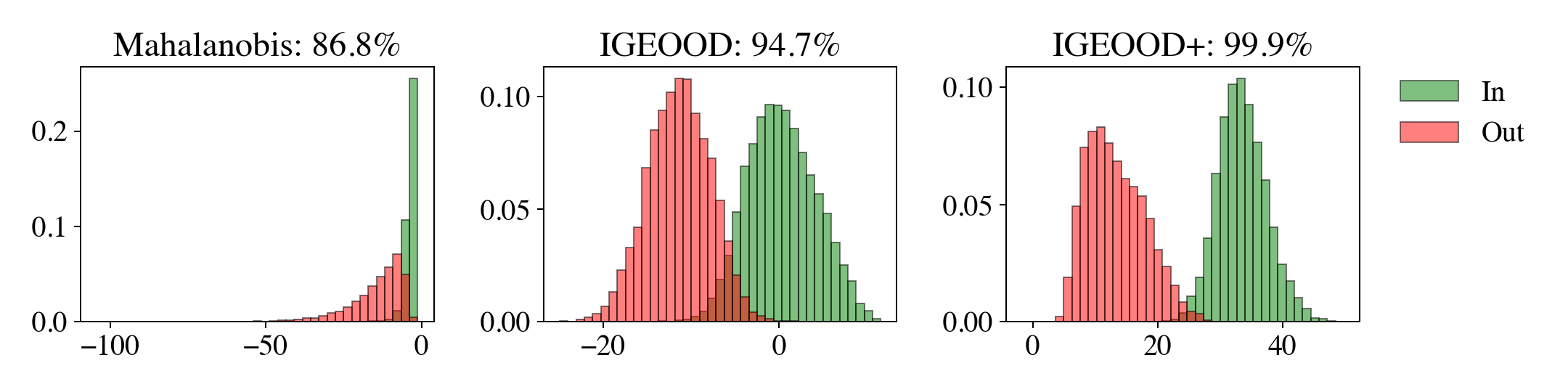}
         \caption{DenseNet on CIFAR-100.}
         \label{fig:h4d2}
     \end{subfigure}
     \begin{subfigure}[b]{0.85\textwidth}
         \centering
         \includegraphics[width=\textwidth]{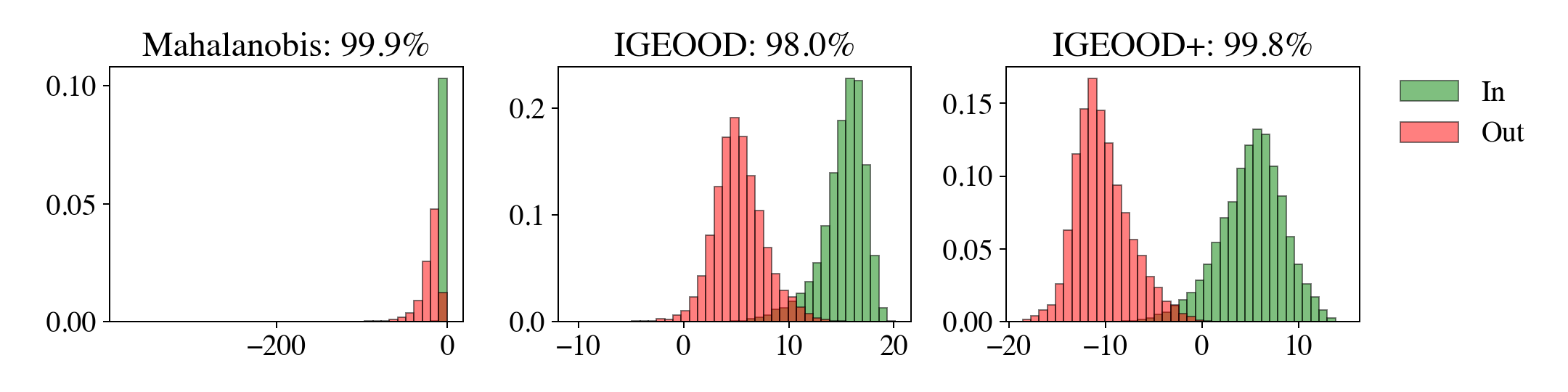}
         \caption{DenseNet on SVHN.}
         \label{fig:h4d3}
     \end{subfigure}
     \begin{subfigure}[b]{0.85\textwidth}
         \centering
         \includegraphics[width=\textwidth]{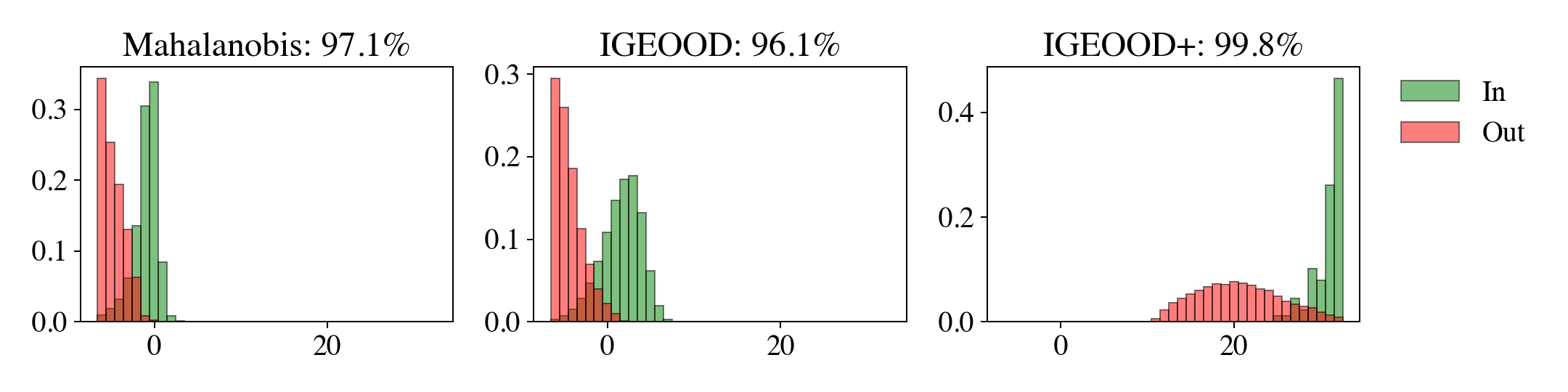}
         \caption{ResNet on CIFAR-10.}
         \label{fig:h4r1}
     \end{subfigure}
     \begin{subfigure}[b]{0.85\textwidth}
         \centering
         \includegraphics[width=\textwidth]{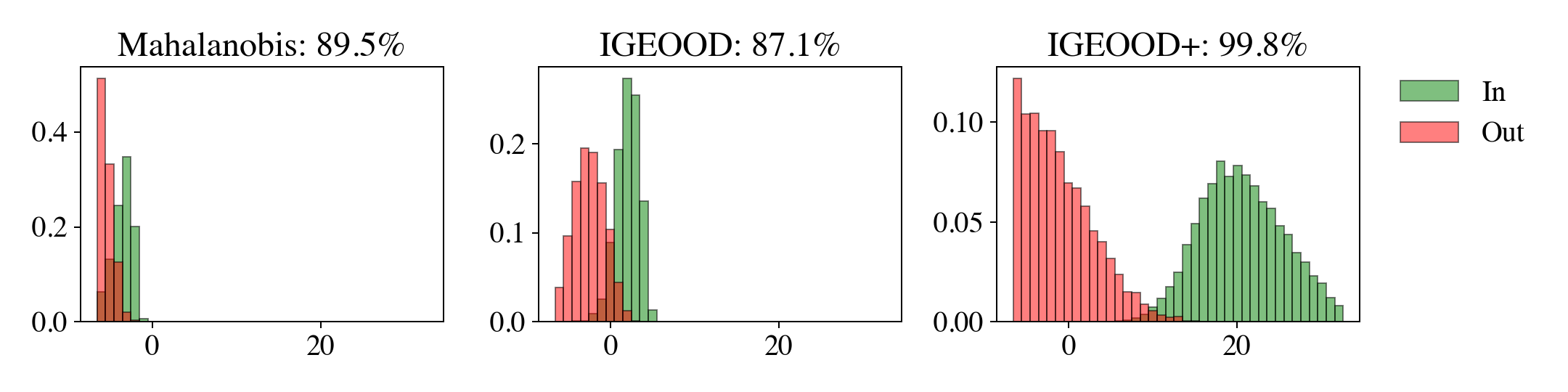}
         \caption{ResNet on CIFAR-100.}
         \label{fig:h4r2}
     \end{subfigure}
     \begin{subfigure}[b]{0.85\textwidth}
         \centering
         \includegraphics[width=\textwidth]{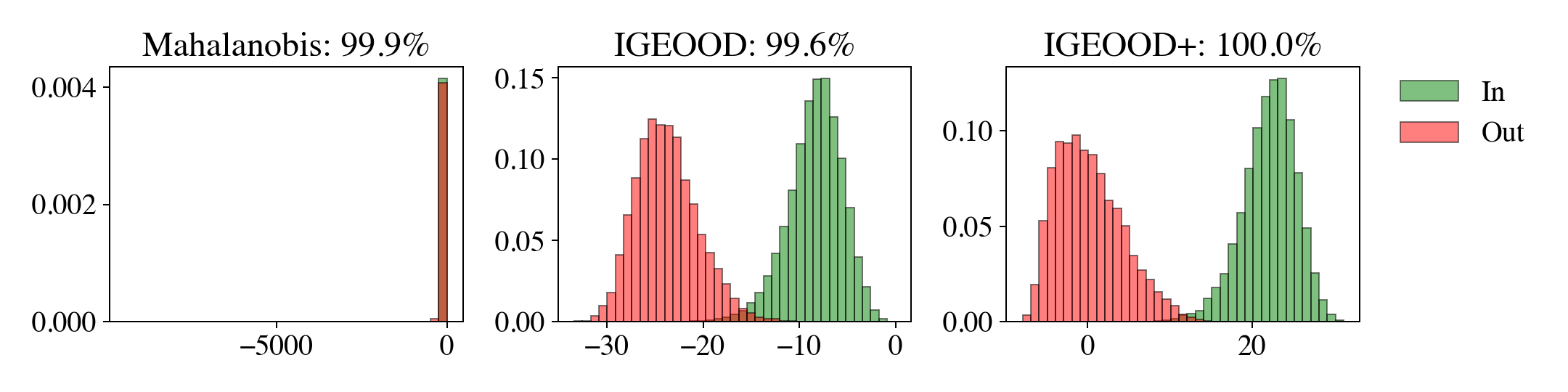}
         \caption{ResNet on SVHN.}
         \label{fig:h4r3}
     \end{subfigure}
         \caption{\textcolor{black}{\textsc{White-Box} setup with validation on OOD data. TinyImageNet as OOD dataset.}}
    \label{fig:wb_histograms}
\end{figure}

\textcolor{black}{Figures \ref{fig:bb_histograms}, \ref{fig:gb_histograms}, \ref{fig:wb_histograms} and \ref{fig:wb_adv_histograms} display histograms for the OOD detection score for \textsc{Igeood} in the \textsc{Block-Box}, \textsc{Grey-Box} and \textsc{White-Box} settings, respectively.}

\end{document}